\documentclass{article}
\pdfoutput=1
\usepackage{microtype}
\usepackage{graphicx}
\usepackage{subcaption}
\usepackage{booktabs} 
\usepackage{color}
\usepackage{dsfont}
\usepackage{lmodern}
\usepackage{float}
\usepackage{amssymb}
\usepackage{amsthm}
\usepackage[dvipsnames]{xcolor}
\usepackage{pifont}
\usepackage{enumitem}
\usepackage[T1]{fontenc}

\usepackage{tabularx}
\usepackage{mathtools}
\usepackage{bbm}
\usepackage[xindy,acronym,toc,smallcaps,nomain]{glossaries}
\makeglossaries

\newacronym{BO}{bo}{Bayesian Optimisation}
\newacronym{GP}{gp}{Gaussian Process}
\newacronym{TuRBO}{turbo}{Trust-region Bayesian Optimisation}
\newacronym{COMBO}{combo}{Combinatorial Bayesian Optimisation}
\newacronym{MAB}{mab}{Multi-Armed Bandit}
\newacronym{CoCaBO}{cocabo}{Categorical and Continuous Bayesian Optimisation}
\newacronym{MVRSM}{mvrsm}{Mixed-Variable ReLU-based Surrogate Modelling}
\newacronym{RF}{rf}{Random Forest}
\newacronym{ASR}{asr}{Attack Success Rate}
\newacronym{CASMOPOLITAN}{casmopolitan}{placeholder}
\newacronym{MaxSAT}{maxsat}{Weighted Maximum Satisfiability}
\newacronym{TR}{tr}{Trust Region}

\glsunset{CASMOPOLITAN}

\usepackage[backref=page, pdflang={en-UK}]{hyperref}
\renewcommand*{\backref}[1]{}
\renewcommand*{\backrefalt}[4]{%
    \ifcase #1 (Not cited)%
    \or        (Cited on page~#2)%
    \else      (Cited on pages~#2)%
    \fi}
    
\usepackage[accepted]{icml2021}

\usepackage{todonotes}

\newtheorem{theorem}{Theorem}[section]

\newtheorem{assumption}{Assumption}[section]

\newtheorem{lemma}{Lemma}[section]

\usepackage{comment}

\begin{document}

\twocolumn[
\icmltitle{Think Global and Act Local: Bayesian Optimisation over High-Dimensional Categorical and Mixed Search Spaces}



\begin{icmlauthorlist}
\icmlauthor{Xingchen Wan}{mlrg}
\icmlauthor{Vu Nguyen}{amzn}
\icmlauthor{Huong Ha}{rmit}
\icmlauthor{Binxin Ru}{mlrg}
\icmlauthor{Cong Lu}{mlrg}
\icmlauthor{Michael A. Osborne}{mlrg}
\end{icmlauthorlist}

\icmlaffiliation{mlrg}{Machine Learning Research Group, University of Oxford, Oxford, UK}
\icmlaffiliation{amzn}{Amazon, Adelaide, Australia}
\icmlaffiliation{rmit}{RMIT University, Melbourne, Australia}

\icmlcorrespondingauthor{Xingchen Wan}{xwan@robots.ox.ac.uk}

\icmlkeywords{Machine Learning, ICML}

\vskip 0.3in
]



\printAffiliationsAndNotice{}  

\global\long\def\se{\hat{\text{se}}}%

\global\long\def\interior{\text{int}}%

\global\long\def\boundary{\text{bd}}%

\global\long\def\new{\text{*}}%

\global\long\def\stir{\text{Stirl}}%

\global\long\def\dist{d}%

\global\long\def\HX{\entro\left(X\right)}%
 
\global\long\def\entropyX{\HX}%

\global\long\def\HY{\entro\left(Y\right)}%
 
\global\long\def\entropyY{\HY}%

\global\long\def\HXY{\entro\left(X,Y\right)}%
 
\global\long\def\entropyXY{\HXY}%

\global\long\def\mutualXY{\mutual\left(X;Y\right)}%
 
\global\long\def\mutinfoXY{\mutualXY}%

\global\long\def\xnew{y}%

\global\long\def\bx{\mathbf{x}}%

\global\long\def\bh{\mathbf{h}}%

\global\long\def\bw{\mathbf{w}}%

\global\long\def\bz{\mathbf{z}}%

\global\long\def\bu{\mathbf{u}}%

\global\long\def\bs{\boldsymbol{s}}%

\global\long\def\bk{\mathbf{k}}%

\global\long\def\bX{\mathbf{X}}%

\global\long\def\tbx{\tilde{\bx}}%

\global\long\def\by{\mathbf{y}}%

\global\long\def\bY{\mathbf{Y}}%

\global\long\def\bZ{\boldsymbol{Z}}%

\global\long\def\bU{\boldsymbol{U}}%

\global\long\def\bv{\boldsymbol{v}}%

\global\long\def\bn{\boldsymbol{n}}%

\global\long\def\bV{\boldsymbol{V}}%

\global\long\def\bK{\boldsymbol{K}}%

\global\long\def\bbeta{\gvt{\beta}}%

\global\long\def\bmu{\gvt{\mu}}%

\global\long\def\btheta{\boldsymbol{\theta}}%

\global\long\def\blambda{\boldsymbol{\lambda}}%

\global\long\def\bgamma{\boldsymbol{\gamma}}%

\global\long\def\bpsi{\boldsymbol{\psi}}%

\global\long\def\bphi{\boldsymbol{\phi}}%

\global\long\def\bpi{\boldsymbol{\pi}}%

\global\long\def\eeta{\boldsymbol{\eta}}%

\global\long\def\bomega{\boldsymbol{\omega}}%

\global\long\def\bepsilon{\boldsymbol{\epsilon}}%

\global\long\def\btau{\boldsymbol{\tau}}%

\global\long\def\bSigma{\gvt{\Sigma}}%

\global\long\def\realset{\mathbb{R}}%

\global\long\def\realn{\realset^{n}}%

\global\long\def\integerset{\mathbb{Z}}%

\global\long\def\natset{\integerset}%

\global\long\def\integer{\integerset}%

\global\long\def\natn{\natset^{n}}%

\global\long\def\rational{\mathbb{Q}}%

\global\long\def\rationaln{\rational^{n}}%

\global\long\def\complexset{\mathbb{C}}%

\global\long\def\comp{\complexset}%

\global\long\def\compl#1{#1^{\text{c}}}%

\global\long\def\and{\cap}%

\global\long\def\compn{\comp^{n}}%

\global\long\def\comb#1#2{\left({#1\atop #2}\right) }%

\global\long\def\nchoosek#1#2{\left({#1\atop #2}\right)}%

\global\long\def\param{\vt w}%

\global\long\def\Param{\Theta}%

\global\long\def\meanparam{\gvt{\mu}}%

\global\long\def\Meanparam{\mathcal{M}}%

\global\long\def\meanmap{\mathbf{m}}%

\global\long\def\logpart{A}%

\global\long\def\simplex{\Delta}%

\global\long\def\simplexn{\simplex^{n}}%

\global\long\def\dirproc{\text{DP}}%

\global\long\def\ggproc{\text{GG}}%

\global\long\def\DP{\text{DP}}%

\global\long\def\ndp{\text{nDP}}%

\global\long\def\hdp{\text{HDP}}%

\global\long\def\gempdf{\text{GEM}}%

\global\long\def\ei{\text{EI}}%

\global\long\def\rfs{\text{RFS}}%

\global\long\def\bernrfs{\text{BernoulliRFS}}%

\global\long\def\poissrfs{\text{PoissonRFS}}%

\global\long\def\grad{\gradient}%
 
\global\long\def\gradient{\nabla}%

\global\long\def\cpr#1#2{\Pr\left(#1\ |\ #2\right)}%

\global\long\def\var{\text{Var}}%

\global\long\def\Var#1{\text{Var}\left[#1\right]}%

\global\long\def\cov{\text{Cov}}%

\global\long\def\Cov#1{\cov\left[ #1 \right]}%

\global\long\def\COV#1#2{\underset{#2}{\cov}\left[ #1 \right]}%

\global\long\def\corr{\text{Corr}}%

\global\long\def\sst{\text{T}}%

\global\long\def\SST{\sst}%

\global\long\def\ess{\mathbb{E}}%

\global\long\def\Ess#1{\ess\left[#1\right]}%

\global\long\def\fisher{\mathcal{F}}%

\global\long\def\bfield{\mathcal{B}}%
 
\global\long\def\borel{\mathcal{B}}%

\global\long\def\bernpdf{\text{Bernoulli}}%

\global\long\def\betapdf{\text{Beta}}%

\global\long\def\dirpdf{\text{Dir}}%

\global\long\def\gammapdf{\text{Gamma}}%

\global\long\def\gaussden#1#2{\text{Normal}\left(#1, #2 \right) }%

\global\long\def\gauss{\mathbf{N}}%

\global\long\def\gausspdf#1#2#3{\text{Normal}\left( #1 \lcabra{#2, #3}\right) }%

\global\long\def\multpdf{\text{Mult}}%

\global\long\def\poiss{\text{Pois}}%

\global\long\def\poissonpdf{\text{Poisson}}%

\global\long\def\pgpdf{\text{PG}}%

\global\long\def\wshpdf{\text{Wish}}%

\global\long\def\iwshpdf{\text{InvWish}}%

\global\long\def\nwpdf{\text{NW}}%

\global\long\def\niwpdf{\text{NIW}}%

\global\long\def\studentpdf{\text{Student}}%

\global\long\def\unipdf{\text{Uni}}%

\global\long\def\transp#1{\transpose{#1}}%
 
\global\long\def\transpose#1{#1^{\mathsf{T}}}%

\global\long\def\mgt{\succ}%

\global\long\def\mge{\succeq}%

\global\long\def\idenmat{\mathbf{I}}%

\global\long\def\trace{\mathrm{tr}}%

\global\long\def\argmax#1{\underset{_{#1}}{\text{argmax}} }%

\global\long\def\argmin#1{\underset{_{#1}}{\text{argmin}\ } }%

\global\long\def\diag{\text{diag}}%

\global\long\def\norm{}%

\global\long\def\spn{\text{span}}%

\global\long\def\vtspace{\mathcal{V}}%

\global\long\def\field{\mathcal{F}}%
 
\global\long\def\ffield{\mathcal{F}}%

\global\long\def\inner#1#2{\left\langle #1,#2\right\rangle }%
 
\global\long\def\iprod#1#2{\inner{#1}{#2}}%

\global\long\def\dprod#1#2{#1 \cdot#2}%

\global\long\def\norm#1{\left\Vert #1\right\Vert }%

\global\long\def\entro{\mathbb{H}}%

\global\long\def\entropy{\mathbb{H}}%

\global\long\def\Entro#1{\entro\left[#1\right]}%

\global\long\def\Entropy#1{\Entro{#1}}%

\global\long\def\mutinfo{\mathbb{I}}%

\global\long\def\relH{\mathit{D}}%

\global\long\def\reldiv#1#2{\relH\left(#1||#2\right)}%

\global\long\def\KL{KL}%

\global\long\def\KLdiv#1#2{\KL\left(#1\parallel#2\right)}%
 
\global\long\def\KLdivergence#1#2{\KL\left(#1\ \parallel\ #2\right)}%

\global\long\def\crossH{\mathcal{C}}%
 
\global\long\def\crossentropy{\mathcal{C}}%

\global\long\def\crossHxy#1#2{\crossentropy\left(#1\parallel#2\right)}%

\global\long\def\breg{\text{BD}}%

\global\long\def\lcabra#1{\left|#1\right.}%

\global\long\def\lbra#1{\lcabra{#1}}%

\global\long\def\rcabra#1{\left.#1\right|}%

\global\long\def\rbra#1{\rcabra{#1}}%

\begin{abstract}


High-dimensional black-box optimisation remains an important yet notoriously challenging problem. Despite the success of Bayesian optimisation methods on continuous domains, domains that are categorical, or that mix continuous and categorical variables, remain challenging. We propose a novel solution -- we combine local optimisation with a tailored kernel design, effectively handling high-dimensional categorical and mixed search spaces, whilst retaining sample efficiency. We further derive convergence guarantee for the proposed approach. Finally, we demonstrate empirically that our method outperforms the current baselines on a variety of synthetic and real-world tasks in terms of performance, computational costs, or both.

\end{abstract}

\section{Introduction}




\gls{BO} \cite{Jones_1998Efficient,Brochu_2010Tutorial,Shahriari_2016Taking}, which features expressive surrogate model(s) and sample efficiency, has found many applications in black-box optimisation, particularly when each evaluation is expensive. Such applications include but not limited to selection of chemical compounds \cite{Hernandez_2017Parallel}, reinforcement learning \cite{pb2}, hyperparameter optimisation of machine learning algorithms \cite{Snoek_2012Practical}, and neural architecture search \cite{kandasamy2018neural, nguyen2020optimal, ru2020neural}

Despite its impressive performance, various challenges still remain for \gls{BO}. The popular surrogate choice,  vanilla \gls{GP} models, is limited to problems of modest dimensionality defined in a continuous space. 
However, real-world optimisation problems are often neither low-dimensional nor continuous: many large-scale practical problems exhibit complex interactions among high-dimensional input variables, and are often \textit{categorical} in nature or involve a mixture of both continuous and categorical input variables. An example of the former is the maximum satisfiability problem, whose exact solution is \textsc{np}-hard \cite{creignou2001complexity}, and an example for the latter is the hyperparameter tuning for a deep neural network: the optimisation scope comprise both continuous hyperparameters, e.g., learning rate and momentum, and categorical ones, e.g., optimiser type \{\textsc{sgd}, Adam, ...\} and learning rate scheduler type \{step decay, cosine annealing\}.

These problems are challenging for a number of reasons: 
first, categorical variables do not have a natural ordering similar to continuous ones for which \gls{GP}s are well-suited. 
Second, the search space grows exponentially with the dimension and the mixed spaces are usually high-dimensional, making the objective function highly multimodal, often heterogeneous, and thus difficult to be modelled by a good, global surrogate \cite{Rana_ICML2017High,eriksson2019scalable}. Partially due to these difficulties, only very few prior works \cite{Hutter_2011Sequential,gopakumar2018algorithmic_NIPS,nguyen2020bayesian,cocabo} have focused on developing \gls{BO} strategies for such problems, and, to the best of our knowledge, achieving promising performance, easy applicability for high-dimensional inputs and reasonable computing costs simultaneously is still an open question.
 
 To tackle these challenging yet important problems, we propose a novel yet conceptually simple method. It not only fully preserves the merits of \gls{GP}-based \gls{BO} approaches, such as expressiveness and sample efficiency, but also demonstrates state-of-the-art performance in high-dimensional optimisation problems, involving categorical or mixed search spaces. Specifically, we make the following contributions:
 
\begin{itemize}[leftmargin=0.04in, noitemsep, topsep=0.05pt]
    \item Propose a new \gls{GP}-based \gls{BO} approach which designs tailored \gls{GP} kernels and harnesses the concept of local trust region to effectively handle high-dimensional optimisation over categorical and mixed search spaces.
    \item Derive convergence analysis to show that our proposed method converges to the global maximum of the objective function in both categorical and mixed space settings, under some assumptions.
    \item Empirically show that our method achieves superior performance, better sample efficiency, or both, over the existing approaches for a wide variety of tasks. The code implementation of our method is available at \url{https://github.com/xingchenwan/Casmopolitan}.
\end{itemize}

\section{Related work}


\paragraph{BO for high-dimensional problems}
A popular class of high-dimensional \gls{BO} methods \cite{kandasamy2015high, rolland2018high, wang2017batched, wang2018batched, mutny2019efficient} decompose the search space into multiple overlapping or disjoint low-dimensional subspaces and use an additive surrogate (e.g. additive \gls{GP}s). However, accurately inferring the decomposition is often very expensive. Another group of \gls{BO} methods \cite{binois2015warped, wang2016bayesian, binois2020choice} assume the objective function is mainly influenced by a small subset of effective dimensions and aims to learn such low-dimensional effective embedding \cite{wang2016bayesian, nayebi2019framework, letham2020re}. However, its effectiveness is conditional on the extent the assumption holds.
A recent state-of-the-art approach is \gls{TuRBO} \cite{eriksson2019scalable}, which constrains \gls{BO} on local \gls{TR} centered around the best inputs so far. This circumvents the aforementioned issues such as the need for finding an accurate global surrogate and over-exploration due to large regions of high posterior variance. However, its convergence properties are not analysed, and it only works in continuous spaces.


\paragraph{BO for categorical search spaces}
The basic approach is to one-hot transform the categorical variables into continuous \cite{Rasmussen_2006gaussian, gpyopt2016,Snoek_2012Practical}. While simple in implementation, the drawbacks are equally obvious: first, for a $d_h$-dimensional problems with $\{n_1, ..., n_{d_h}\}$ choices per input, the one-hot-transformed problem has $\sum_{i=1}^{d_h}n_i$ dimensions, further aggravating the curse of dimensionality. Second, categorical spaces differ fundamentally with the continuous in, for e.g., differentiability and continuity, with function values only defined in finite locations. These lead to difficulties in using gradient-based methods in acquisition function optimisation of the transformed problems.

To ameliorate these drawbacks, \textsc{bocs} \cite{baptista2018bayesian} first tailors \gls{BO} in categorical spaces: it uses a sparse monomial representation up to the second order and Bayesian linear regression as the surrogate, and is primarily used for boolean optimisation. Inevitably, its expressiveness is constrained by the quadratic model, while scaling beyond the second order and/or to high dimensionality is usually intractable due to the exponentially-increasing number of parameters that need be learnt explicitly. \gls{COMBO} \cite{oh2019combinatorial} is a state-of-the-art method that instead uses a \gls{GP} surrogate (which is capable of learning interactions of an arbitrary order), and is capable of dealing with multi-categorical problems via a combinatorial graph over all possible joint assignments of the variables and a diffusion graph kernel to model the interactions.
Nonetheless, both methods deal with categorical optimisation only, which is an important problem in its own right, but does not extend to our setting of mixed-variable problems. They also suffer from poor scalability (e.g. to avoid overfitting \gls{COMBO} approximately marginalises the posterior via Monte Carlo sampling instead of cheaper optimisation, and it needs to pre-compute the combinatorial graph beforehand). Other methods, such as \textsc{comex} and its inspired works \cite{dadkhahi2020combinatorial, dadkhahi2021fourier} take a non-Bayesian black-box optimisation approach to improve computing efficiency, but they are typically less sample-efficient with respect to the number of function queries and are less suitable for problems where querying the objective functions is expensive. Finally, several recent works aim to improve \gls{BO} on combinatorial structures by improving the effectiveness \cite{Deshwal_Belakaria_Doppa_Fern_2020} or reducing the expenses \cite{swersky2020amortized} of the \emph{acquisition function}; these are largely orthogonal to our method, and we defer a thorough investigation on whether there are additional benefits by combining with these methods to a future work.

\begin{figure*}[t]
    \centering
    \includegraphics[trim=0cm 0cm 0cm 0cm, clip, width=1.0\linewidth]{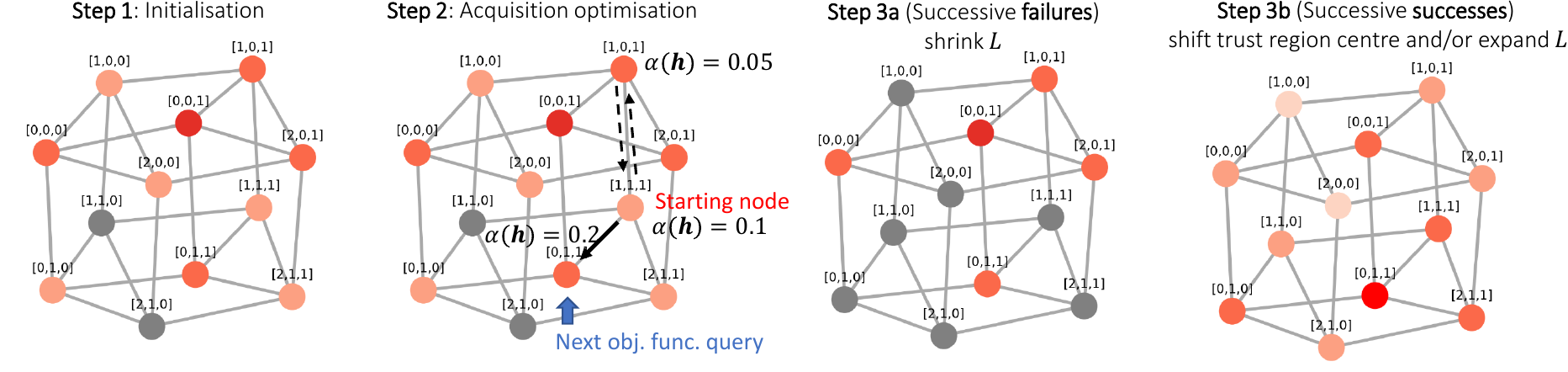}
    \caption{Illustration of \gls{CASMOPOLITAN} in categorical space. Suppose we optimise over a 3-dimensional problem with $\{3, 2, 2\}$ choices for each input respectively. Initially (\textbf{Step 1}), the best location so far $\mathbf{h}^*_T = \arg \max_{\mathbf{h}} \{y_j\}_{j=1}^T$ (marked in \textcolor{red}{red}) is $[0, 0, 1]$ with \gls{TR} radius $L=2$ (the \textcolor{orange}{orange} nodes, with different shades denoting their Hamming distances to $\mathbf{h}^*_T$. The \textcolor{gray}{gray} nodes are outside the current \gls{TR}). In optimisation of the acquisition function (\textbf{Step 2}), we conduct local search within the \gls{TR}, moving to a neighbour only if it has a higher acquisition function value $\alpha(\cdot)$ and is still within the \gls{TR}. In case of successive failures (\textbf{Step 3a}) in increasing $\mathbf{h}^*_T$, we shrink the \gls{TR} down to length $L^h_{\min}$, below which we restart the optimisation,  or in case of successive successes (\textbf{Step 3b}), we shift the \gls{TR} centre to the new $\mathbf{h}^*_T$ and/or expand \gls{TR} up to length $L^h_{\max}$. Note that the combinatorial graph is shown here for illustration; it does not need to be computed explicitly or otherwise.
    }  
    \label{fig:catmethod}
\end{figure*}

\paragraph{BO for mixed input types}
\gls{BO} in mixed categorical-continuous search spaces is still rather under-explored, despite attempts in modelling less complicated spaces, such as mixed continuous-integer problems \cite{daxberger2019mixed,garrido2020dealing}. In our specific setting, \gls{CoCaBO} \cite{cocabo} first explicitly handles multiple categorical and continuous variables: it alternates between selecting the categorical inputs with a \gls{MAB} and the continuous inputs with \gls{GP}-\gls{BO}, and uses a tailored kernel to connect the two. However, \gls{CoCaBO} requires optimising a \gls{MAB} over a non-stationary reward (since the values of continuous variables improves over \gls{BO} iterations and hence so does the function value). Furthermore, \gls{MAB} requires pulling each arm at least once, and hence it is difficult to scale \gls{CoCaBO} to high-dimensional problems, where the total number of possible arm combinations explode exponentially. Lastly, while the two sub-components are provably convergent, \gls{CoCaBO} as a whole is not. Related works along this direction also include \citet{gopakumar2018algorithmic_NIPS} and \citet{nguyen2020bayesian}, but the continuous inputs are constrained to be \textit{specific} to the categorical choice, and being \gls{MAB}-based, it also suffers from aforementioned limitations. Separately, \citet{bliek2020black} recently propose \gls{MVRSM}, which the authors claim to be suitable for mixed-variable, high-dimensional problems. However, in trading for efficiency, the expressiveness is limited by the ReLU formulation and we compare against it in Sec. \ref{sec:experiments}.


 In addition to these more recent works explicitly handling the mixed spaces, earlier attempts such as \textsc{smac} with \gls{RF} \cite{Breiman_2001Random} surrogates \cite{Hutter_2011Sequential} are also compatible. However, the predictive distribution of the \gls{RF} used to select new evaluation is less accurate due to reliance on randomness from bootstrap samples and the randomly chosen subset of variables to be tested at each node to split the data. Moreover, \gls{RF}s easily suffer from overfitting and require careful hyperparameter choice.


\section{CASMOPOLITAN: BO for Categorical and Mixed Search Spaces}


\paragraph{Problem Statement} 
We consider the problem of optimising an expensive black-box function, defined over a categorical domain or one with mixed continuous and categorical inputs. Formally, we consider a function in the mixed domain for generality: $f: [\mathcal{H}, \mathcal{X}] \rightarrow \mathbb{R}$ where $\mathcal{H}$ and $\mathcal{X} \subset \mathbb{R}^{d_x}$ denote the categorical and continuous search spaces, respectively (for problems over categorical domains, we simply have $f: \mathcal{H} \rightarrow \mathbb{R}$ and the goal is to find $\mathbf{h}^*  = \arg \max f(\mathbf{h})$).
 We further denote $\mathbf{z} = [\mathbf{h}, \mathbf{x} ]$ to be an input in the mixed space where $\mathbf{h}$ and $\mathbf{x}$ are the categorical and continuous parts, $d_h$ to be the number of categorical variables, i.e. $\mathbf{h}=[h_1, h_2, ...,h_{d_h}]$, and the number of possible, distinct value that the $j$-th categorical variable may take to be $n_j$. Given $f$, at time $t$ we observe the noisy perturbation of the form $y_t = f(\mathbf{z}_t) + \epsilon_t$ where $\epsilon_t \sim \mathcal{N}(0,\sigma^2)$ and $\sigma^2$ is a noise variance which can be learned by maximizing the log-marginal likelihood \cite{Rasmussen_2006gaussian}. We sequentially select inputs $\bz_t  \text{ } \forall t=1,...,T $ (or simply $\mathbf{h}_t$ if the problem is purely categorical) to query $f$ with the goal of finding the maximiser the objective $\bz^* = \arg \max f(\bz)$ with the fewest numbers of iterations. 
We further include a primer on \gls{GP} and \gls{BO} in App. \ref{app:primer_on_gp}.

\begin{algorithm}[tb]
\begin{footnotesize}
	    \caption{\gls{CASMOPOLITAN}.} 
	    \label{alg:LBO_categorical}
	\begin{algorithmic}[1]
		\STATE {\bfseries Input:}  \#init (the number of random initialing points at initialisation or restarts), \#iter $T$, initial \gls{TR} size for categorical $L^h_0 \in \mathbb{Z}^+$, and continuous variables $L^x_0 \in \mathbb{R}^+$.
		\STATE {\bfseries Output:} The best recommendation $\mathbf{z}_T$
\STATE \texttt{restart} = \texttt{True} // \textit{Set restart to True initially}
        \FOR{$t=1, \dots, T$}
        \IF {\texttt{restart}}
        \STATE Reset \gls{TR} $L^h = L^h_0$ and $L^x = L^x_0$ and reset \gls{GP}. Randomly select \#init points in the search space as $\mathbf{z}_t$ \emph{(if at initialisation)}, or set the \gls{TR} center as the point determined by Eq. (3) and randomly select \#init points within the newly constructed \gls{TR} as $\mathbf{z}_t$ \emph{(if at subsequent restarts)}.
        \ELSE
        	\STATE Construct a \gls{TR} $\mathrm{TR_{\mathbf{h}}(\mathbf{h}^*_t)}$  around the categorical dimensions of the best point  $\mathbf{h}^*_t$ using Eq. (\ref{eq:trust_region_categorical}). 
        	\STATE Construct a hyper-rectangular \gls{TR} of length $L^x$, $\mathrm{TR_{\mathbf{x}}(\mathbf{x}^*_t)}$ for the continuous variables.
        	\STATE Select next query pt(s) \textit{within the \gls{TR}s} $\mathbf{z}_t = \mathrm{argmax}_\mathbf{z} \alpha(\mathbf{z}) \text{ } \mathrm{s.t.} \text{ } \mathbf{x} \in \mathrm{TR}_{x}(\mathbf{x^*_t)}, \mathbf{h} \in \mathrm{TR}_{h}(\mathbf{h^*_t)}$.
         \ENDIF
        \STATE Query at $\mathbf{z}_{t}$ to obtain $y_{t}$; fit/update the surrogate $\mathcal{D}_{t} \leftarrow \mathcal{D}_{t-1} \cup (\mathbf{z}_t , y_{t})$ and optimise \gls{GP} hyperparameters.
        \STATE Update the \gls{TR}s and decide whether to \texttt{restart}.
    	\ENDFOR
	\end{algorithmic}
\end{footnotesize}
\end{algorithm}

\begin{figure*}[t]
    \centering
    \includegraphics[trim=0cm 0cm 0cm 0cm, clip, width=1.0\linewidth]{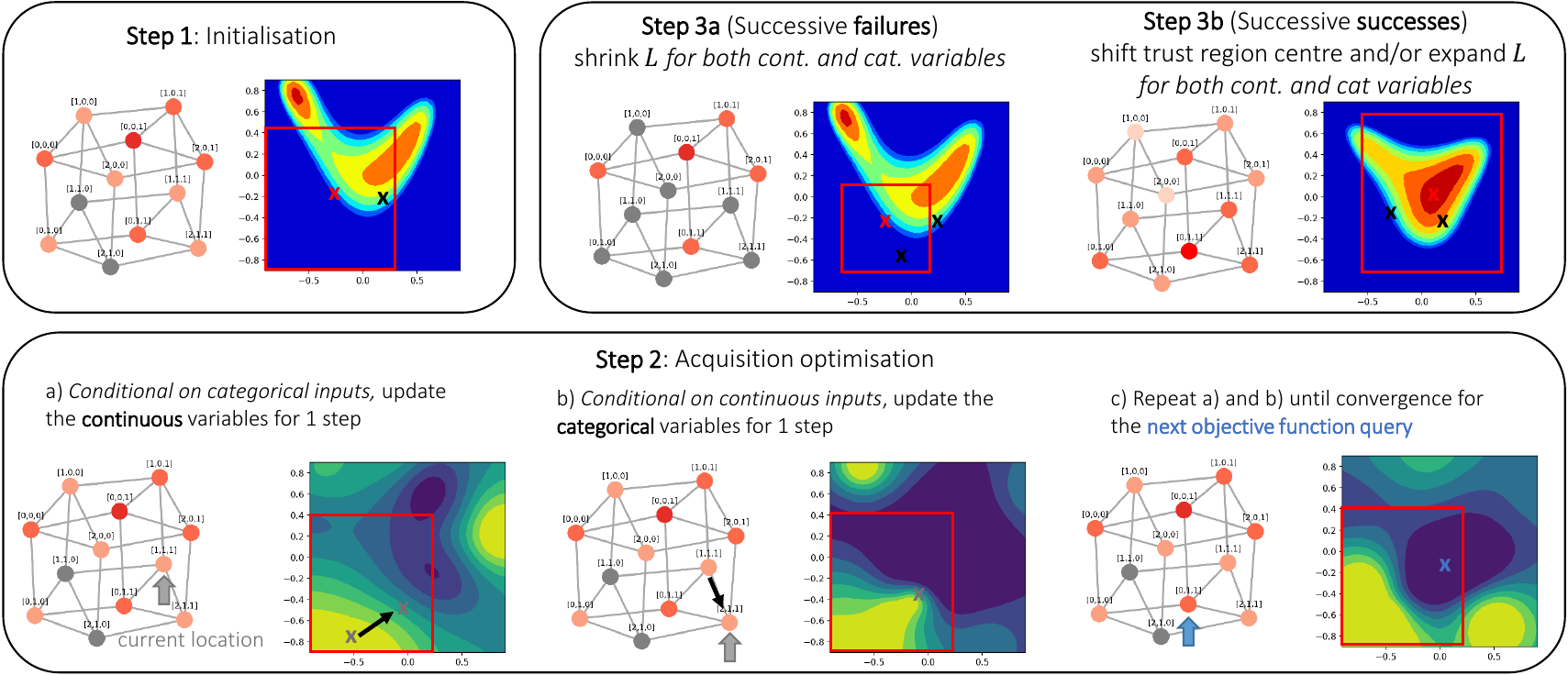}
    \caption{Illustration of \gls{CASMOPOLITAN} in mixed space. Note that in \textbf{Steps 1} \& \textbf{3} we show the \textit{\gls{GP} posterior} on $\mathcal{X}$ conditioned on the incumbent $\mathbf{h^*_T}$, and in \textbf{Step 2} we show the \textit{acquisition function} on $\mathcal{X}$ conditioned on $\mathbf{h}$ at various optimisation steps. Suppose we optimise over a 5-dimensional mixed problem with the categorical dimensions identical to that in Fig. \ref{fig:catmethod} and 2 additional continuous dimensions. Initially (\textbf{Step 1}), the best location so far $\mathbf{z}^*_T = \arg \max_{\mathbf{z}} \{y_j)\}_{z=1}^T = [\mathbf{h^*_T}, \mathbf{x^*_T}]$ (with the continuous TR and $\mathbf{x}^*_T$ in \textcolor{red}{red box and cross}).  In optimisation of acquisition function (\textbf{Step 2}), we interleave the local search on $\mathcal{H}$ described in Sec. \ref{subsec:combinatorialmethod} with gradient-based optimisation on $\mathcal{X}$ until convergence. In \textbf{Steps 3a}/\textbf{3b}, we adjust both the continuous and categorical \gls{TR}s correspondingly and restart if/when either shrinks below its minimum length.
    }  
    \label{fig:mixedmethod}
\end{figure*}

\subsection{Categorical Search Space}
\label{subsec:combinatorialmethod}

Our first contribution is to propose a conceptually-simple yet effective \gls{BO} strategy that preserves all of the advantages of \gls{GP} modelling, but is specifically designed for the categorical search space (later extended to the mixed space in Sec. \ref{subsec:mixedmethod}). We present an illustration in Fig. \ref{fig:catmethod} and the pseudocode in Algorithm \ref{alg:LBO_categorical}. We name our algorithm \gls{CASMOPOLITAN} (\underline{CA}tegorical \underline{S}paces, or \underline{M}ixed, \underline{OP}timisati\underline{O}n with \underline{L}ocal-trust-reg\underline{I}ons \& \underline{TA}ilored \underline{N}on-parametric),
  and we highlight the key design features in this section.
\paragraph{Kernel design} In Line 12 of Algorithm \ref{alg:LBO_categorical}, we impose \gls{GP} on the categorical variables with a kernel defined directly on them (note that it does not increase the dimensions like one-hot transform). Specifically, we modify the overlap (or Hamming) kernel $k(\mathbf{h}, \mathbf{h}') = \frac{\sigma}{d_h}\sum_{i=1}^{d_h} \delta(h_i, h_i'),
$ in \citet{cocabo} and \citet{kondor2002diffusion}:
\begin{equation}
\label{eq:catkernel}
k_h(\mathbf{h}, \mathbf{h}') = \exp\Big(\frac{1}{d_h}\sum_{i=1}^{d_h} \ell_i \delta(h_i, h_i')\Big),
\end{equation}
where $\{\ell_i\}_i^{d_h}$ are the lengthscale(s)\footnote{The lengthscales will be different for each dimension if we enable automatic relevance determination (\textsc{ard}).}, and $\delta(\cdot, \cdot)$ is the Kronecker delta function. The modification affords additional expressiveness in modelling more complicated functions: for e.g., the kernel in Eq. (\ref{eq:catkernel}) can discern the dimensions to which the objective function value is more sensitive via learning different lengthscales but the original categorical overlap kernel treats all dimensions equally. We empirically validate the performance gain of the exponentiated kernel in Sec. \ref{subsec:ablation}, and we prove this kernel is positive semi-definite (p.s.d) in App. \ref{sec:proof-psd}.

\paragraph{Trust region}
One key difficulty in applying \gls{GP}-\gls{BO} in high-dimensional problems is that the surrogate, by default, attempts to model the entire function landscape and over-explores. Optimisation over the categorical search space also suffer this problem. To effectively scale up the dimensions, we adapt the \gls{TR} approach from \citet{eriksson2019scalable} in categorical search space (Line 8 in Algorithm \ref{alg:LBO_categorical}). However, the challenge is that the Euclidean distance-based \gls{TR} is no longer applicable; instead, we define \gls{TR}s in terms of \textit{Hamming distance}, i.e. a \gls{TR} of radius $L^h$ from the best location, $\mathbf{h}^*$, observed at iteration $T$ includes all points that are up to $L^h$ variables different from $\mathbf{h}^*$: 
\begin{equation}
\centering
\mathrm{TR}_{h}(\mathbf{h}^*)_{L^h} = \Big\{\mathbf{h} \mid \sum_{i=1}^{d_h} \delta(h_i, h^*_i) \leq L^h\Big\}.    \label{eq:trust_region_categorical}
\end{equation}
The \gls{TR} radius is adjusted dynamically during optimisation, expanding on successive successes (if best function value $f^*_T$ improves) and shrinking otherwise. Since Hamming distance is integer-valued bounded in $[0, d_h]$, we also set these two values as the minimum and maximum \gls{TR} lengths $L^h_{\min}$ and $L^h_{\max}$.


\gls{TR}s in local optimisation are typically biased toward the starting points. Therefore, most local optimisation approaches rely on a restarting strategy to attain good performance \cite{shylo2011restart, kim2018adaptive}. In our case, we restart the optimisation when the \gls{TR} length $L^h$ reaches the smallest possible value (Line 13 in Algorithm~\ref{alg:LBO_categorical}).

Rather than restarting randomly as in \citet{eriksson2019scalable}, we propose to restart our method using \gls{GP}-\textsc{ucb} principle \cite{Srinivas_2010Gaussian}, which as we will show in Section \ref{subsec:theory} is crucial for theoretical guarantee. Specifically, we introduce an auxiliary global \gls{GP} model to achieve this. Suppose we are restarting the $i$-th time, we first fit the global \gls{GP} model on a subset of data $D^*_{i-1}= \lbrace \mathbf{h}^*_j, y^*_j \rbrace_{j=1}^{i-1}$, where $\mathbf{h}^*_j$ is the local maxima found after the $j$-th restart. Alternative, a random data point, if the found local maxima after the $j$-th restart is same as one of previous restart. Let us also denote $\mu_{gl}(\mathbf{h}; D^*_{i-1})$ and $\sigma^2_{gl}(\mathbf{h}; D^*_{i-1})$ as the posterior mean and variance of the global \gls{GP} learned from $D^*_{i-1}$. Then, at the $i$-th restart, we select the following location $\mathbf{h}^{(0)}_i$ as the initial centre of the new \gls{TR}:
\begin{align}
    \mathbf{h}^{(0)}_i= \arg \max_{\mathbf{h} \in \mathcal{H}} \mu_{gl}(\mathbf{h}; D^*_{i-1}) + \sqrt{\beta_i} \sigma_{gl}(\mathbf{h}; D^*_{i-1}),
    \label{eq:trucb}
\end{align}
where $\beta_i$ is the trade-off parameter. 
As formally shown in Sec. \ref{subsec:theory}, this strategy is optimal in deciding the next \gls{TR} by balancing exploration against exploitation \cite{Srinivas_2010Gaussian}. Finally, while the use of \textsc{ucb}-restart is primarily theoretically driven, we show that it could offer empirical benefits over random restarts, and the readers are referred to App. \ref{appendix:additionalexperiments} for details.

\paragraph{Optimisation of the acquisition function}
Since we preserve the discrete nature of the variables in our method, we cannot optimise the acquisition function via gradient-based methods. Instead, we use the simple strategy of local search within the \gls{TR}s defined previously: at each \gls{BO} iteration, we randomly sample an initial configuration $\mathbf{h}_0 \in \mathrm{TR}_{h}(\mathbf{h^*})$. We then randomly select a neighbour point of Hamming distance $1$ to $\mathbf{h}_0$, evaluate its acquisition function $\alpha(\cdot)$, and move from $\mathbf{h}_0$ if the neighbour has a higher acquisition function value and is still within the \gls{TR}. We repeat this process until a pre-set budget of queries is exhausted and dispatch the best configurations for objective function evaluation (Line 10 in Algorithm \ref{alg:LBO_categorical}).

\subsection{Extension to Mixed Search Spaces}
\label{subsec:mixedmethod}
In addition to the purely categorical problems, our method naturally generalises to mixed, and potentially high-dimensional, categorical-continuous spaces, a setting frequently encountered in real life but hitherto under-explored in \gls{BO} literature. To handle such an input $\mathbf{z} = [ \mathbf{h}, \mathbf{x}]$ where $\mathbf{x}$ is the continuous inputs, we first modify the \gls{GP} kernel to the one proposed in \citet{cocabo}:
\begin{align}
\label{eq:mixkernel}
    k(\mathbf{z}, \mathbf{z'}) & = \lambda \Big(k_x(\mathbf{x}, \mathbf{x}')k_h(\mathbf{h}, \mathbf{h}')\Big) \nonumber \\     & + (1 - \lambda) \Big(k_h(\mathbf{h}, \mathbf{h}') + k_x(\mathbf{x}, \mathbf{x}')\Big),
\end{align}
where $\lambda \in [0, 1]$ is a trade-off parameter, $k_h$ is defined in Eq. (\ref{eq:catkernel}) and $k_x$ is a kernel over continuous variables (we use the Mat\'{e}rn 5/2 kernel). While we use the same kernel as \citet{cocabo}, we emphasise and formally show in Sec. \ref{subsec:theory} that, unlike \gls{CoCaBO}, \gls{CASMOPOLITAN} retains convergence guarantee even in the mixed space. 

This formulation therefore allows us to use tailored kernels that are most appropriate for the different input types while still flexibly capturing the possible additive and multiplicative interactions between them. For the continuous inputs, we use a standard \gls{TuRBO} surrogate \cite{eriksson2019scalable} by maintaining, and adjusting where necessary, separate standard hyper-rectangular \gls{TR}(s) for them $\mathrm{TR}_{x}(\mathbf{x^*})_L = \big\{\mathbf{x}|$ $\mathbf{x \in \mathcal{X}}$ and within the box centered around $\mathbf{x}^* \big\}$. We include an illustration in Fig. \ref{fig:mixedmethod}. We restart the continuous \gls{TR} $\mathrm{TR}_x$ in similar manner as described in Eq. (\ref{eq:trucb}), if and when the either $\mathrm{TR}_h$ or $\mathrm{TR}_x$ length $L$ reaches the smallest possible value. 

\paragraph{Interleaved acquisition optimisation} In \citet{cocabo}, the categorical $\mathbf{h}$ and continuous $\bx$  of the proposed points $\bz=[\mathbf{h}, \bx]$  are optimised separately similar to a \textit{single} \textsc{em}-style iteration: the categorical parts are first proposed by the multi-armed bandit; \textit{conditioned on these}, the continuous parts are then suggested by optimising the acquisition function. In our approach, since both the categorical and continuous inputs are handled by a single, unified \gls{GP}, we may propose points and optimise acquisition functions more naturally and effectively: at each optimisation step (Line 10 of Algorithm \ref{alg:LBO_categorical}), we simply do one step of local search defined in Sec. \ref{subsec:combinatorialmethod} on the categorical variables,  followed by one step of gradient-based optimisation of the acquisition function on the continuous variables. However, instead of doing this alternation once, we repeat until convergence or when a maximum number of steps is reached.

\paragraph{Other types of discrete input} While we mainly focus on categorical-continuous problems, our method can  be easily generalised to more complex settings by virtue of its highly flexible sub-components. For instance, we often encounter combinatorial variables with ordinal relations: for these, we treat them as categorical, but instead of using Kronecker delta function in Eq. (\ref{eq:catkernel}) we encode the problem-specific distances. We defer a full investigation to a future work, but we include some preliminary studies in App. \ref{subsec:additionalresults}.  

\subsection{Theoretical Analysis}
\label{subsec:theory}

We first provide upper bounds on the maximum information gains of our proposed categorical kernel in Eq. (\ref{eq:catkernel}) and mixed kernel in Eq. (\ref{eq:mixkernel}) (Theorem \ref{thr:kernelmig}). We then prove that after a restart, under Assumptions \ref{assu:f-bounded} and \ref{assu:gp-approx}, \gls{CASMOPOLITAN} converges to a local maxima after a finite number of iterations or converges to the global maximum (Theorem \ref{thr:local-converge}). Finally, we prove that with our UCB-restart strategy, under Assumptions \ref{assu:f-bounded}, \ref{assu:gp-approx} and some assumptions described in \citet{Srinivas_2010Gaussian}, \gls{CASMOPOLITAN} converges to the global maximum with a sublinear rate over the number of restarts in both categorical (Theorem \ref{thr:gconverge-cat}) and mixed space settings (Theorem \ref{thr:gconverge-mix}). We refer readers to App. \ref{appendix:proofs} for the detailed proofs.

\begin{theorem} \label{thr:kernelmig}
Let us define $\gamma(T;k;V) := \max_{A \subseteq V, \vert A \vert \leq T} \dfrac{1}{2} \log \vert I + \sigma^{-2} \lbrack k(\mathbf{v}, \mathbf{v}') \rbrack_{\mathbf{v}, \mathbf{v}' \in A} \vert$ as the maximum  information gain achieved by sampling $T$ points in a \gls{GP} defined over a set $V$ with a kernel  $k$. Let us define $\tilde{N} := \prod_{j=1}^{d_h} n_j$, then we have,
\begin{enumerate} [topsep=0pt, noitemsep]
\item
For the categorical kernel $k_h$, $\gamma(T;k_h;\mathcal{H}) = \mathcal{O}(\tilde{N} \log T)$;
\item
For the mixed kernel $k$, $\gamma(T;k;[\mathcal{H}, \mathcal{X}]) \leq \mathcal{O} \big( (\lambda \tilde{N} + 1 - \lambda)\gamma(T;k_x;\mathcal{X}) + (\tilde{N}+2-2\lambda)\log T \big) $.
\end{enumerate}
\end{theorem}

Using Theorem \ref{thr:kernelmig}, the maximum information gain of the mixed kernel $k$ can be upper bounded for some common continuous kernels $k_x$. For instance, when $k_x$ is the Mat\'{e}rn kernel, the maximum information gain $\gamma(T;k;[\mathcal{H}, \mathcal{X}])$ of the mixed kernel is upper bounded by $\mathcal{O} \big( (\lambda \tilde{N} + 1 - \lambda) T^{d_x(d_x+1)/(2v + d_x(d_x+1))}(\log T) + (\tilde{N}+2-2\lambda)\log T \big)$ as $\gamma(T;k_{Mt};\mathcal{X}) = \mathcal{O}(T^{d_x(d_x+1)/(2v + d_x(d_x+1))}(\log T))$ \cite{Srinivas_2010Gaussian}. Similar bounds can be established when  $k_x$ is the squared exponential or the linear kernel.

To analyse the convergence property of \gls{CASMOPOLITAN}, similar to any \gls{TR}-based algorithm \cite{yuan2000review}, we assume that (i) $f$ is bounded in $[\mathcal{H}, \mathcal{X}]$ (Assumption \ref{assu:f-bounded}), and (ii), given a small enough region, the surrogate model (i.e. \gls{GP}) accurately approximates $f$ with any data point belonging to this region (Assumption \ref{assu:gp-approx}). We note that Assumption \ref{assu:f-bounded} is common as it is generally assumed in \gls{BO} that $f$ is Lipschitz continuous \cite{Brochu_2010Tutorial}, thus $f$ is bounded given the search space is bounded. Assumption \ref{assu:gp-approx} considers the minimum \gls{TR} lengths $L^x_{\min}, L^h_{\min}$ are set to be small enough so that \gls{GP} approximates $f$ accurately in \gls{TR}s specified in Assumption \ref{assu:gp-approx}. We note that in practice, this assumption is only possible asymptotically, i.e. when the number of observed data in these \gls{TR}s goes to infinity. In our implementation (see App. \ref{appendix:details}), these \gls{TR}s are always set to be very small so that Assumption \ref{assu:gp-approx} can be close to true.


\begin{assumption} \label{assu:f-bounded}
The objective function $f(\mathbf{z})$ is bounded in $[\mathcal{H}, \mathcal{X}]$, i.e. $\exists F_l, F_u \in \mathbb{R} : \forall \mathbf{z} \in [\mathcal{H}, \mathcal{X}]$, $F_l \leq  f(\mathbf{z}) \leq F_u$.
\end{assumption}

\begin{assumption} \label{assu:gp-approx} Let us denote $L^h_{\min}$, $L^x_{\min}$ and $L_0^h, L_0^x$ be the minimum and initial \gls{TR} lengths for the categorical and continuous variables, respectively. Let us also denote $\alpha_s$ as the shrinking rate of the \gls{TR}s. In the categorical setting, for any \gls{TR} with length $\leq \lceil (L^h_{\min}+1)/\alpha_s \rceil -1$,\footnote{The operator $\lceil . \rceil$ denotes the ceiling function.} the corresponding local \gls{GP} approximates $f$ accurately. That is, the \gls{GP} posterior mean approximates $f$ accurately whilst the \gls{GP} posterior variance is negligible within this \gls{TR}. In the mixed space setting, the local \gls{GP} approximates $f$ accurately within any \gls{TR} with length $L^x \leq \max \big(L^x_{\min}/\alpha_s, L_0^x (\lceil (L^h_{\min}+1)/\alpha_s \rceil -1)/L_0^h \big)$ and $L^h \leq \max \big(\lceil (L^h_{\min}+1)/\alpha_s \rceil - 1, \lceil L_0^h L^x_{\min}/ (\alpha_s L_0^x) \rceil \big)$.
\end{assumption}

\begin{theorem} \label{thr:local-converge}
Given Assumptions \ref{assu:f-bounded} \& \ref{assu:gp-approx}, after a restart, \gls{CASMOPOLITAN} converges to a local maxima after a finite number of iterations or converges to the global maximum.
\end{theorem}

Finally, we define the cumulative regret after $I$ restarts, $R_{I}$, to be $\sum_{j=1}^I (f(\mathbf{z}^*)-f(\mathbf{z}^*_{j}))$ with $\mathbf{z}^*_j$ being the local maxima found at the $j$-th restart and $\mathbf{z}^*$ being the global maximum of $f$. We then provide the regret bounds of \gls{CASMOPOLITAN} in both categorical (Theorem \ref{thr:gconverge-cat}) and mixed space setting (Theorem \ref{thr:gconverge-mix}). With these regret bounds, it can be seen that \gls{CASMOPOLITAN} converges to the global maximum with a sublinear rate over the number of restarts (i.e. $R_{I} / I  \xrightarrow{I \rightarrow \infty} 0 $) in both categorical and mixed space settings.

\begin{theorem} \label{thr:gconverge-cat}
 Let us consider the categorical setting, $f: \mathcal{H} \rightarrow \mathbb{R}$. Let $\zeta \in (0, 1)$ and $\beta_i=2\log(\vert \mathcal{H} \vert i^2 \pi^2 / 6 \zeta)$ at the $i$-th restart. Suppose the objective function $f$ satisfies that: there exists a class of functions which pass through all the local maxima of $f$,\footnote{This means for every function $g$ belonging to this class of functions, $g(\mathbf{h}_j^*) = f(\mathbf{h}_j^*)$ where $\mathbf{h}^*_j$ is a local maxima of $f$.} share the same global maximum with $f$, and is sampled from the auxiliary global \gls{GP} $GP(0,k_h)$. Then given Assumptions \ref{assu:f-bounded} \& \ref{assu:gp-approx}, \gls{CASMOPOLITAN} obtains a regret bound of $\mathcal{O}^*\big(\sqrt{I \gamma(I; k_h, \mathcal{H}) \log \vert \mathcal{H} \vert}\big)$ w.h.p. Formally,
 \begin{equation} \nonumber
 \text{Pr} \Big\lbrace R_I \leq  \sqrt{C_1 I \beta_I \gamma(I; k_h, \mathcal{H}) } \quad \forall I \geq 1 \Big \rbrace \geq 1 - \zeta,
 \end{equation}
 with $C_1=8/\log(1+\sigma^{-2})$, $\gamma(I; k_h, \mathcal{H}) = \mathcal{O} (\tilde{N} \log(\tilde{N}) \log (I)$) and $\tilde{N} = \prod_{j=1}^{d_h} n_j$.
\end{theorem}

\begin{theorem} \label{thr:gconverge-mix}
Let us consider the mixed space setting, $f: [\mathcal{H}, \mathcal{X}]  \rightarrow \mathbb{R}$. Let $\zeta \in (0, 1)$. Suppose the objective function $f$ satisfies that: there exists a class of functions $g$ which pass through all the local maximas of $f$, share the same global maximum with $f$ and lies in the RKHS $\mathcal{G}_k([\mathcal{H}, \mathcal{X}])$ corresponding to the kernel $k$ of the auxiliary global \gls{GP} model.  Suppose that the noise $\epsilon_i$ has zero mean conditioned on the history and is bounded by $\sigma$ almost surely. Assume $\Vert g \Vert^2_k \leq B$, and let $\beta_i = 2B + 300\gamma_i \log(i/\zeta)^3$, then given Assumptions \ref{assu:f-bounded} \& \ref{assu:gp-approx}, \gls{CASMOPOLITAN} obtains a regret bound of $\mathcal{O}^*\big(\sqrt{I \gamma(I; k, \lbrack \mathcal{H}, \mathcal{X} \rbrack) \beta_I}\big)$ w.h.p. Specifically,
 \begin{equation} \nonumber
 \text{Pr} \Big\lbrace R_I \leq  \sqrt{C_1 I \beta_I \gamma(I;k;[\mathcal{H}, \mathcal{X}])} \quad \forall I \geq 1 \Big \rbrace \geq 1 - \zeta,
 \end{equation}
 with $C_1=8/\log(1+\sigma^{-2})$, $\gamma(I;k;[\mathcal{H}, \mathcal{X}]) = \mathcal{O} \big( (\lambda \tilde{N} + 1 - \lambda)\gamma(T;k_x;\mathcal{X}) + (\tilde{N}+2-2\lambda)\log T \big)$ and $\tilde{N} = \prod_{j=1}^{d_h} n_j$.
\end{theorem}

\paragraph{Discussion} We show in Theorem \ref{thr:local-converge} that our \gls{TR}-based algorithm with \gls{BO} converges to a local maxima or global maximum after a restart. We note that similar convergence can be found in the original \gls{TR}-based algorithms using gradient-descent \cite{yuan2000review}. However, our proof technique is very different from \citet{yuan2000review}. In addition, in Theorems \ref{thr:gconverge-cat} \& \ref{thr:gconverge-mix}, the fact that \gls{CASMOPOLITAN} converges to the global maximum with a sublinear rate over the number of restarts - not over the number of iterations as in \citet{Srinivas_2010Gaussian} - can be considered as the price paid for a more relaxed assumption. In particular, \citet{Srinivas_2010Gaussian} assume that it is possible to model the objective function $f$ with a \gls{GP} with kernel $k$ on the whole search space. On the other hand, we relax this assumption in Theorems \ref{thr:gconverge-cat} \& \ref{thr:gconverge-mix} by assuming that there is a class of functions, which pass through the local maxima and share the same global maximum with $f$, that we can model with a \gls{GP} with kernel $k$. Further details on this class of functions can be found in Apps. \ref{sec:proof-catconverge} \& \ref{sec:proof-mixconverge}.

Despite the aforementioned strengths, there are some limitations with our theoretical analysis. First, the maximum information gains $\gamma(T;k_h;\mathcal{H})$ and $\gamma(T;k;[\mathcal{H}, \mathcal{X}])$ derived in Theorem \ref{thr:kernelmig} increase exponentially with the dimension of the categorical input ($d_h$). Thus, these terms can be  large when the categorical dimension is high. As we are solving a noisy NP-hard combinatorial problem, it might not be possible to get away these exponential terms without a strict assumption. Second, as briefly discussed above, Assumption \ref{assu:gp-approx} is true asymptotically, resulting Theorems \ref{thr:local-converge}, \ref{thr:gconverge-cat} and \ref{thr:gconverge-mix} to hold asymptotically. One way to eliminate this assumption is to instead prove \gls{CASMOPOLITAN} achieves $\epsilon$~-~$accuracy$, that is, \gls{CASMOPOLITAN} can find a point whose function value is within $\epsilon$ of the objective function global maximum, where $\epsilon$ is a small positive value depending on the minimum \gls{TR} lengths $L^x_{\min}, L^h_{\min}$. We consider these directions for future work.

\section{Experiments}
\label{sec:experiments}

\subsection{Categorical Problems}
\label{subsec:combinatorialresults}


\begin{figure*}[t]
    \centering
     \begin{subfigure}{0.23\linewidth}
    \includegraphics[trim=0cm 0cm 0cm  0cm, clip, width=1.0\linewidth]{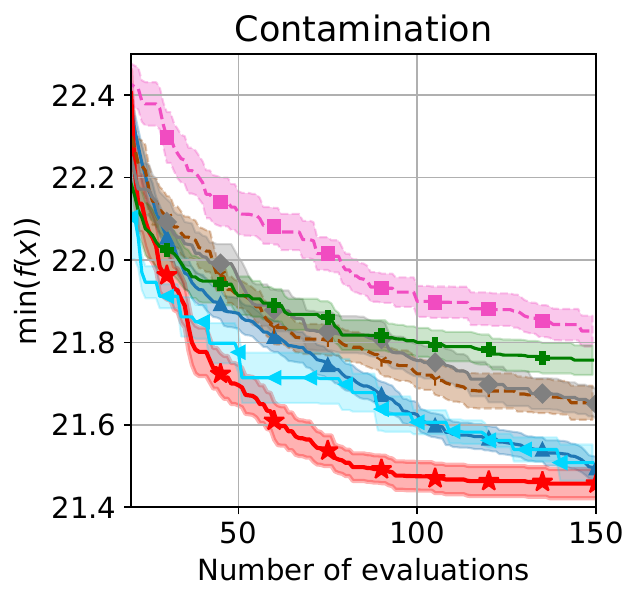}
    \end{subfigure}
    \begin{subfigure}{0.23\linewidth}
        \includegraphics[trim=0cm 0cm 0cm  0cm, clip, width=1.0\linewidth]{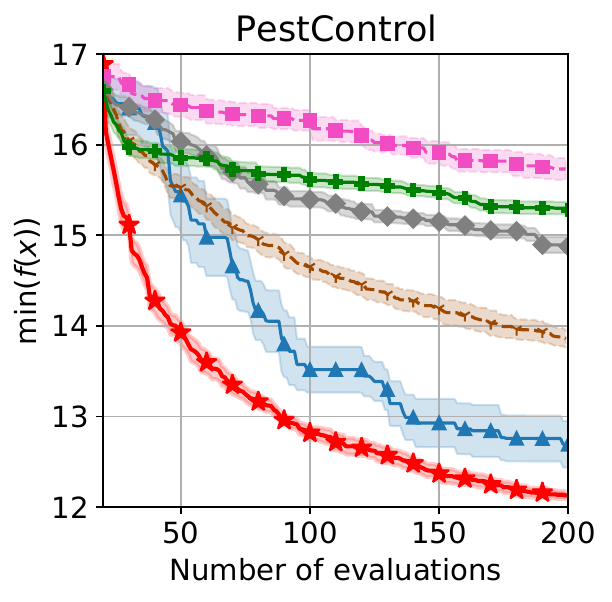}
    \end{subfigure}
         \begin{subfigure}{0.23\linewidth}
    \includegraphics[trim=0cm 0cm 0cm  0cm, clip, width=1.0\linewidth]{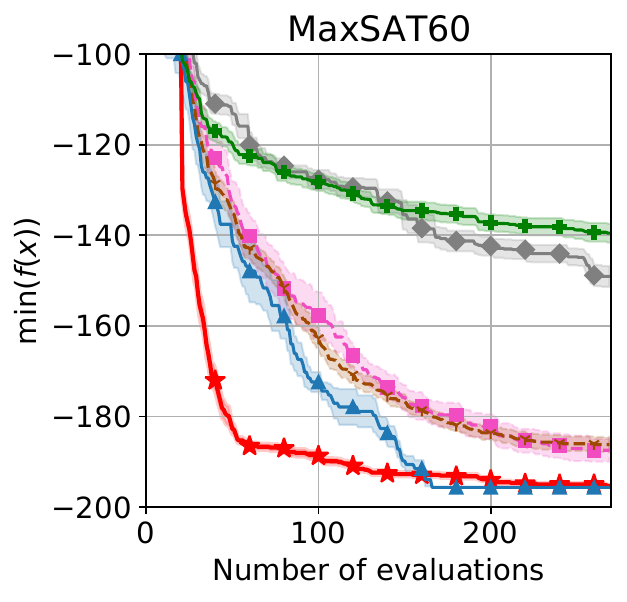}
    \end{subfigure}
        \begin{subfigure}{0.8\linewidth}
        \includegraphics[trim=0cm 0.5cm 0cm  6.8cm, clip, width=1.0\linewidth]{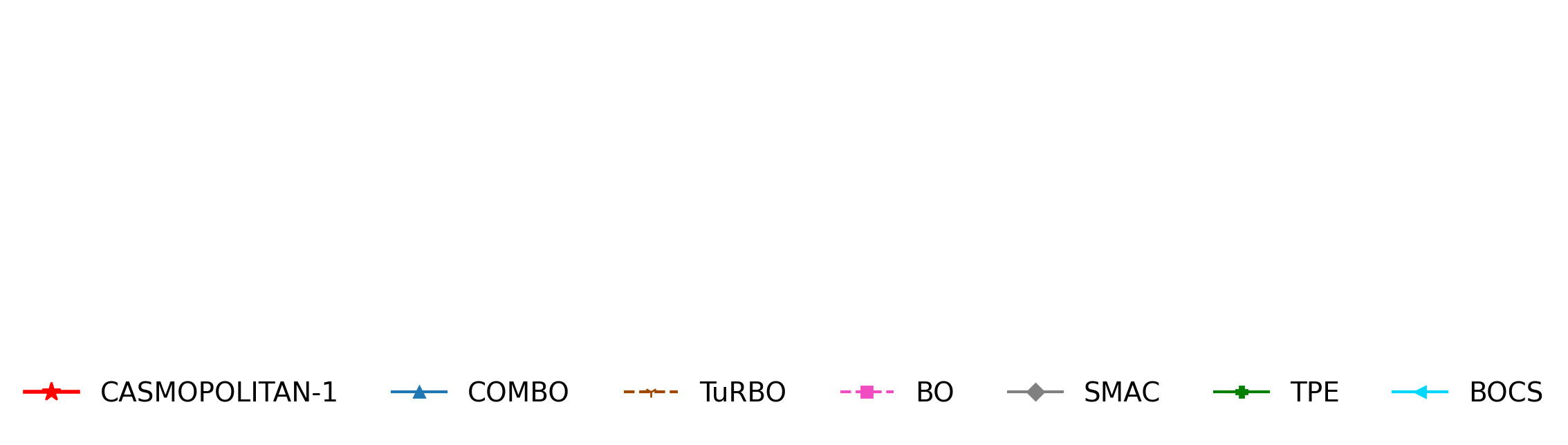}   
        \end{subfigure}
    \caption{Results on various categorical optimisation problems. Lines and shaded area denote mean $\pm$ 1 standard error. 
    }    
    \label{fig:categoricalproblems}
\end{figure*}


\begin{figure*}[t]
    \centering
     \begin{subfigure}{0.24\linewidth}
    \includegraphics[trim=0cm 0cm 0cm  0cm, clip, width=1.0\linewidth]{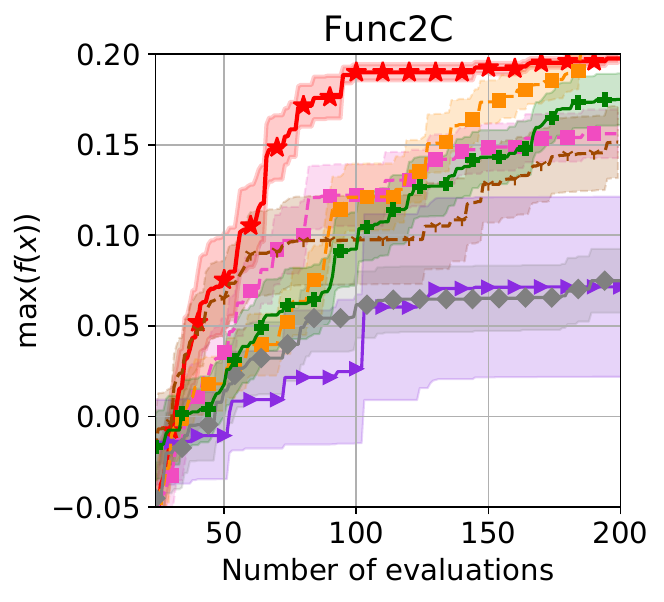}
    \end{subfigure}
    \begin{subfigure}{0.23\linewidth}
        \includegraphics[trim=0cm 0cm 0cm  0cm, clip, width=1.0\linewidth]{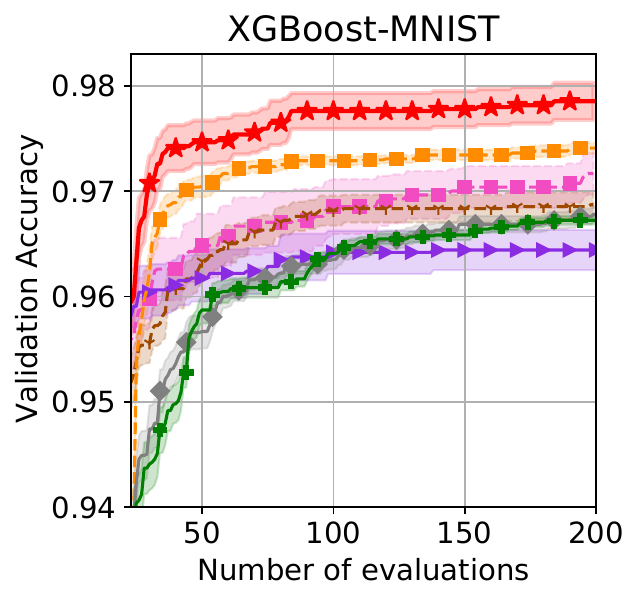}
    \end{subfigure}
         \begin{subfigure}{0.23\linewidth}
    \includegraphics[trim=0cm 0cm 0cm  0cm, clip, width=1.0\linewidth]{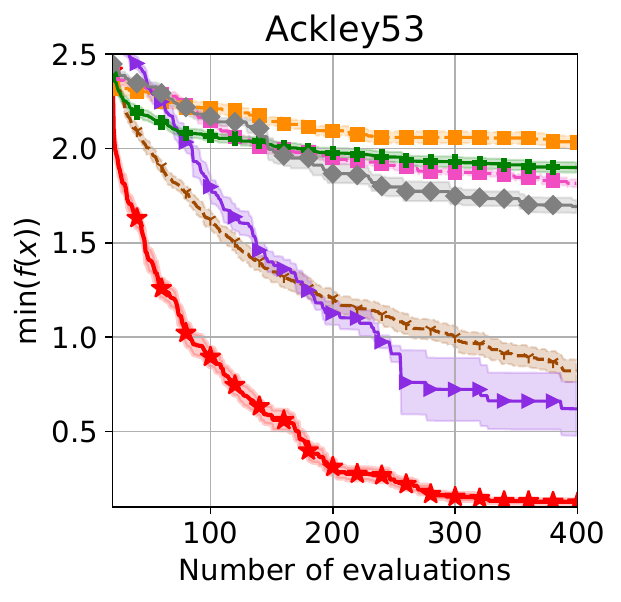}
    \end{subfigure}
             \begin{subfigure}{0.22\linewidth}
    \includegraphics[trim=0cm 0cm 0cm  0cm, clip, width=1.0\linewidth]{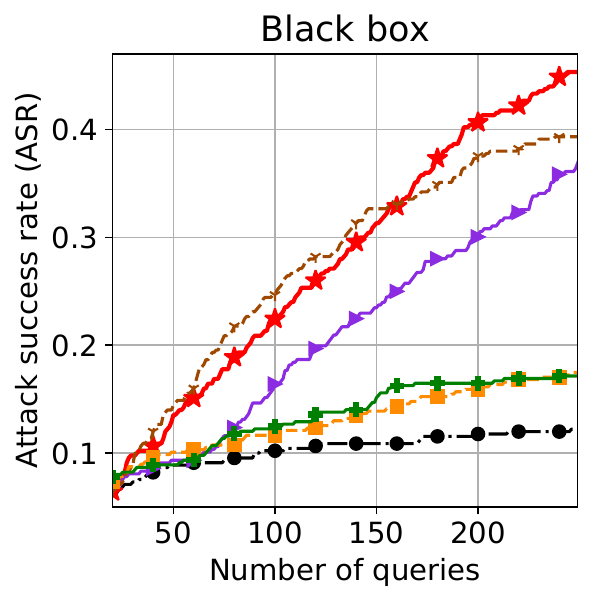}
    \end{subfigure}
        \begin{subfigure}{0.9\linewidth}
        \includegraphics[trim=0cm 0.5cm 0cm  6.8cm, clip, width=1.0\linewidth]{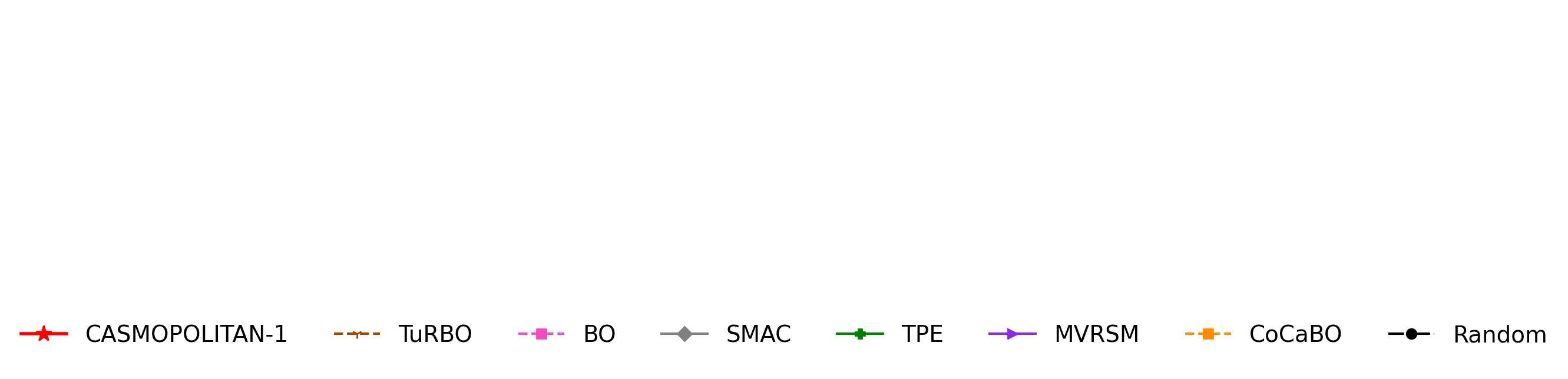}   
        \end{subfigure}
    \caption{Results on various mixed optimisation problems. Lines and shaded area denote mean $\pm$ 1 standard error (except for Black-box where we show the \gls{ASR}  against number of queries). Additional experiment results in App. \ref{appendix:additionalexperiments}.
    }    
    \label{fig:mixedproblems}
\end{figure*}

We first evaluate our proposed method on a number of optimisation problems in the categorical search space against a number of competitive baselines, including \textsc{tpe} \cite{Bergstra_2011Algorithms}, \textsc{smac} \cite{Hutter_2011Sequential}, \textsc{bocs} \cite{baptista2018bayesian}\footnote{\textsc{bocs} is only run in Contamination, as it by default does not support multi-categorical optimisation and on \gls{MaxSAT}, a single trial takes more than $100$ hours, rendering comparison infeasible within our computing constraints.} and \gls{COMBO} \cite{oh2019combinatorial} which claims the state-of-the-art performance amongst comparable algorithms. We also include two additional baselines: \texttt{BO}, which performs the na\"ive \gls{BO} approach after converting the categorical variables into one-hot representations, and \texttt{TuRBO}, which is identical to \texttt{BO} except that we additionally incorporate the \gls{TR} approach in \citet{eriksson2019scalable}. We experiment on following real-life problems (for detailed implementation and descriptions for the setup of these problems and those in Sec. \ref{subsec:mixedresults}, see App. \ref{appendix:details}). 
\begin{itemize}[leftmargin=0.04in, noitemsep, topsep=0.05pt]
    \item Contamination control over $25$ binary variables  ($3.35\times10^7$ configurations). This problem and the Pest control problem below simulate the dynamics of real-life problems whose evaluations are extremely expensive \cite{hu2010contamination}.
    \item Pest control over $25$ variables, with $5$ possible options for each ($2.98\times10^{17}$ configurations) \cite{oh2019combinatorial}. 
    \item Weighted maximum satisfiability (\gls{MaxSAT}) problem over $60$ binary variables ($1.15\times10^{18}$ configurations).
\end{itemize}

In all experiments in this section and Sec. \ref{subsec:mixedresults}, we report the sequential version (denoted as \texttt{\gls{CASMOPOLITAN}-1} as batch size $b=1$) of our method as all baselines we consider are also sequential. We investigate the parallel version of varying batch sizes of our method in Sec. \ref{subsec:parallel}.

The results are shown in Fig. \ref{fig:categoricalproblems}: our method achieves the best convergence speed and sample efficiency in general, and in terms of the performance at termination, our method again outperforms the rest except in Contamination and \gls{MaxSAT} where it performs on par with \gls{COMBO}. However, it is worth noting that in terms of wall-clock speed, our method is $2-3$ times faster than \gls{COMBO} in the problems considered (See App. \ref{appendix:additionalexperiments}). 

\subsection{Mixed Problems}
\label{subsec:mixedresults}

We then consider the optimisation problems involving a mix of continuous and categorical input variables. In these experiments, in addition to \textsc{smac}, \textsc{tpe}, \gls{BO} and \gls{TuRBO} described in Section \ref{subsec:combinatorialresults}, we also include a number of recent advancements in this setup including \gls{CoCaBO} \cite{cocabo} and \gls{MVRSM} \cite{bliek2020black}. Additionally, we  run a small comparison against several other high-dimensional \gls{BO} methods such as \textsc{alebo} \cite{letham2020re} and \textsc{rembo} \cite{wang2016bayesian}, and the readers are referred to details in App. \ref{appendix:additionalexperiments}. Note that we do not compare against \textsc{bocs} and \gls{COMBO} since they are suitable for purely categorical spaces only. Under this setup, we consider the following synthetic and real-life problems of increasing dimensionality and complexity:
\begin{itemize}[leftmargin=0.04in, noitemsep, topsep=0.05pt]
    \item Func2C with $d_h=2$ and $d_x = 2$, and Func3C with $d_h = 3$ and $d_x = 3$, respectively \cite{cocabo}.
    \item Hyperparameter tuning of the XGBoost model \cite{chen2016xgboost} on the \textsc{mnist}  dataset \cite{lecun1998mnist}, with $d_x = 5$ and $d_h = 3$ with $2$ choices for each.
    \item 53-dimensional Ackley function (Ackley-53) \cite{bliek2020black} with $d_h = 50$ where $\mathbf{h} \in \{0, 1\}^{50}$ and $d_x = 3$ where $\mathbf{x} \in [-1, 1]^3$.
    \item Black-box adversarial attack on a \textsc{cnn} trained on \textsc{cifar-10} inspired by \citet{ru2019bayesopt}, but with adapted \emph{sparse} setups where we perturb a small number of pixels only. The task is an optimisation problem with $d_h=43$ ($42$ pixel locations being attacked with $n_{1:42}=14$ choices each and the image upsampling technique which has $n_{43}=3$ choices) and $d_x=42$ for continuous perturbation added to each pixel under attack. We perform a total of $450$ targetted attack instances and limit the maximum budget to be $250$ queries for each attack to simulate a highly constrained attack setup. 
    
\end{itemize}


We report the results on the objective function values in Fig. \ref{fig:mixedproblems} except for the black-box attack, where we instead report the attack success rate \gls{ASR} against the number of queries following \citet{ru2019bayesopt} (Additional attack results are shown in App. \ref{appendix:additionalexperiments}). In this problem we also compare against random search, as it has been shown to be a strong baseline both in adversarial attack \cite{croce2020sparse} and high-dimensional black-box optimisation \cite{Rana_ICML2017High} literature. Overall, it is evident that \gls{CASMOPOLITAN} performs the best, but it is also interesting to observe that in lower dimensions (the first 2 problems), \gls{CoCaBO} featuring tailored categorical kernels performs well, while \gls{MVRSM} and categorical variable-agnostic \gls{TuRBO}, both focusing on high dimensions, under-perform. However, in high-dimensional problems (last 2 problems), the relative performance switches completely, suggesting that the focus on dimensionality now outweighs the importance of treating different input types differently. Nonetheless, with \emph{both} tailored kernels and focus on scaling to high dimensions, \gls{CASMOPOLITAN} consistently out-performs by a comfortable margin, further demonstrating its versatility.

\subsection{Parallel Setting}
\label{subsec:parallel}

\begin{figure}[h]
    \centering
         \begin{subfigure}{0.43\linewidth}
    \includegraphics[trim=0cm 0cm 0cm  0cm, clip, width=1.0\linewidth]{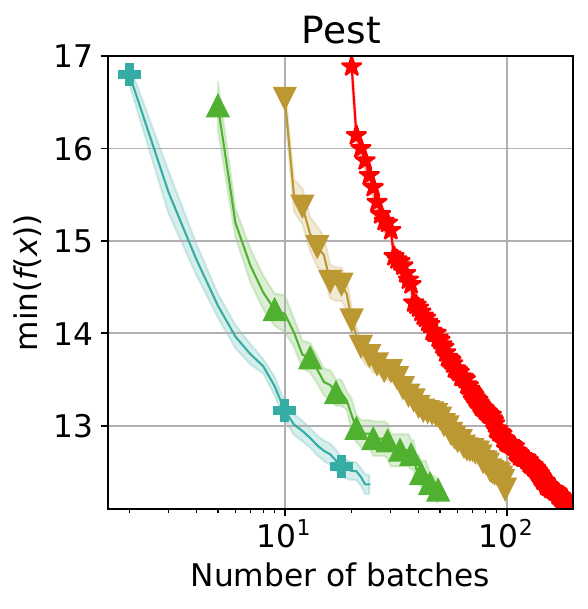}
    \end{subfigure}
         \begin{subfigure}{0.43\linewidth}
    \includegraphics[trim=0cm 0cm 0cm  0cm, clip, width=1.0\linewidth]{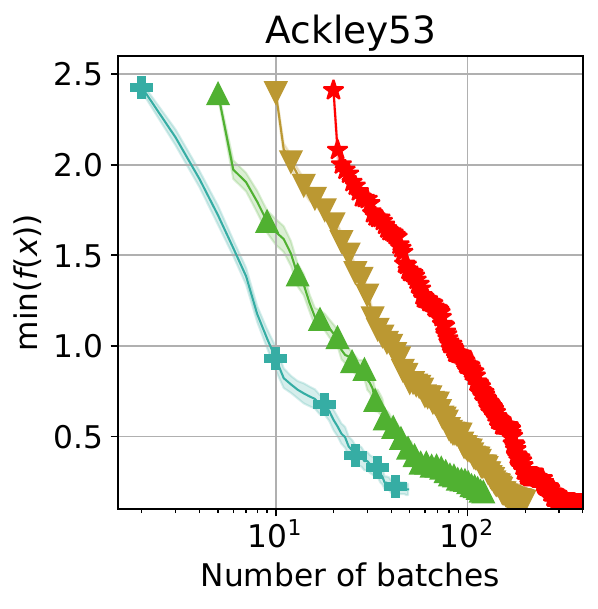}
    \end{subfigure}
    
    \begin{subfigure}{0.7\linewidth}
    \includegraphics[trim=0cm 0.5cm 0cm  5.5cm, clip, width=1.0\linewidth]{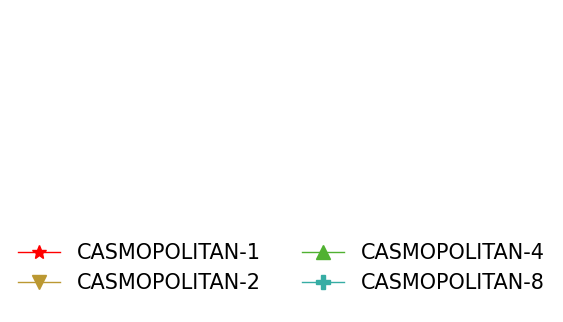}   
    \end{subfigure}
    \caption{Parallel \gls{CASMOPOLITAN} on representative categorical and mixed problems by \emph{number of batches} excluding the initially randomly-sampled batches. Note the x-axis is in log-scale for better presentation. We show the comparison by \emph{number of function queries} in App. \ref{appendix:additionalexperiments}. 
    }    
    \label{fig:parallel}
\end{figure}

We would often like to exploit parallelism in computing where we dispatch different queries to the black-box objective function for independent evaluations. This setting necessitates the development of \textit{batch} methods to propose a batch of $b$ points for simultaneous evaluation at each \gls{BO} iteration. However, this often involves trade-off between  wall-clock time efficiency against performance, because surrogates in batch methods are updated only once per $b$ objective function evaluations. Here we investigate the performance of \gls{CASMOPOLITAN} under different batch sizes where $b = 1$ (sequential setting) $2, 4$ \& $8$ in Pest Control and Ackley-53 problems previously considered; where $b > 1$, we use the Kriging believer strategy \cite{ginsbourger2010kriging} during acquisition optimisation to deliver $b$ proposals simultaneously. In both experiments, we keep the budget of the objective function queries to be identical to that in Sec. \ref{subsec:combinatorialresults} \& \ref{subsec:mixedresults} but scale the number of batches accordingly, and the results are shown in Fig. \ref{fig:parallel}: it is evident that larger batch sizes, while leading to almost linear reduction in wall-clock time, do not lead to significant performance deterioration, except some minor under-performance at the end which seems to scale with $b$. However, in both problems, \gls{CASMOPOLITAN} even with the largest batch size investigated still outperforms \textit{sequential} baselines in Figs. \ref{fig:categoricalproblems} \& \ref{fig:mixedproblems}.

\subsection{Ablation Studies}
\label{subsec:ablation}

\begin{figure}[h]
    \centering
     \begin{subfigure}{0.48\linewidth}
    \includegraphics[trim=0cm 0cm 0cm  0cm, clip, width=1.0\linewidth]{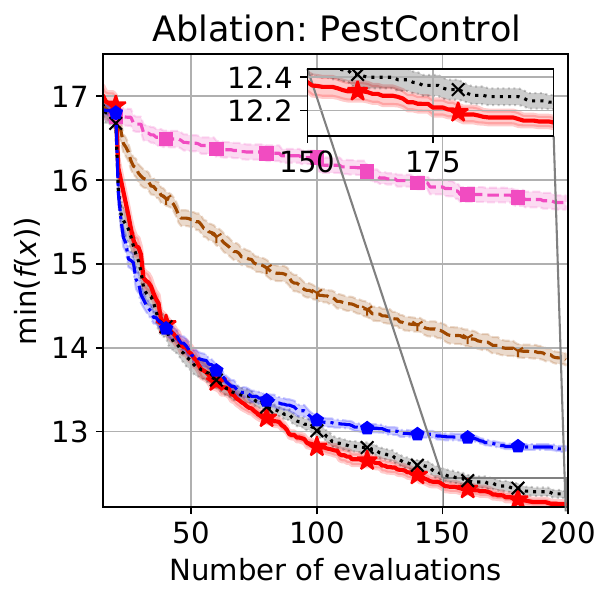}
    \end{subfigure}
    \begin{subfigure}{0.48\linewidth}
        \includegraphics[trim=0cm 0cm 0cm  0cm, clip, width=1.0\linewidth]{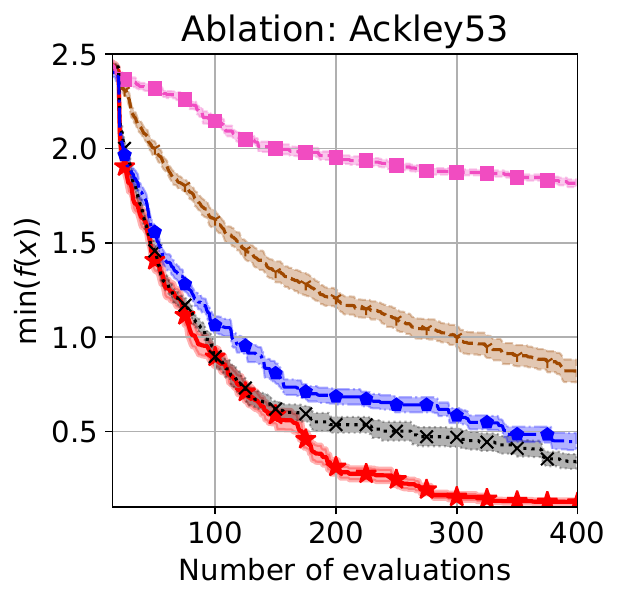}
    \end{subfigure}
    
    \centering
     \begin{subfigure}{0.48\linewidth}
    \includegraphics[trim=0cm 0cm 0cm  0cm, clip, width=1.0\linewidth]{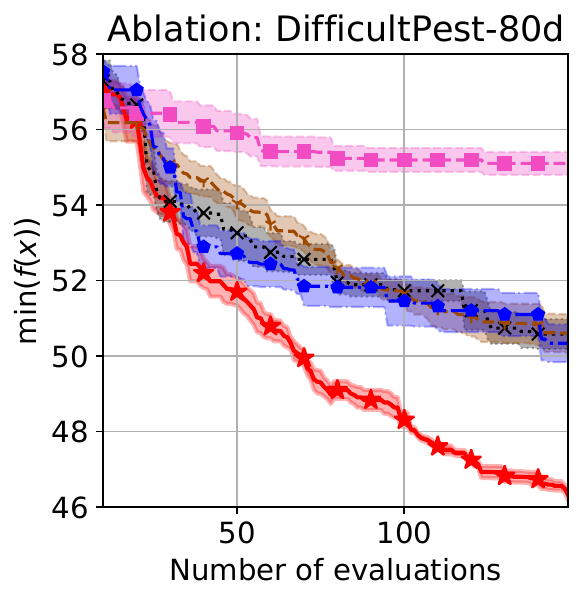}
    \end{subfigure}
    \begin{subfigure}{0.48\linewidth}
        \includegraphics[trim=0cm 0cm 0cm  0cm, clip, width=1.0\linewidth]{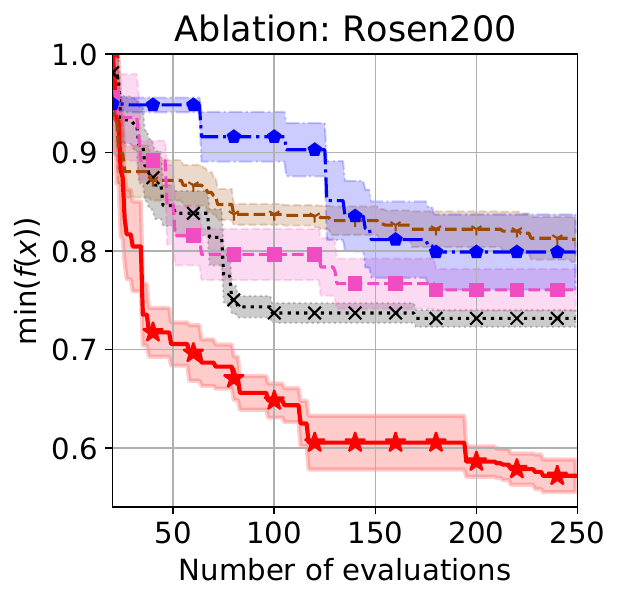}
    \end{subfigure}
    
        \begin{subfigure}{\linewidth}
    \includegraphics[trim=0cm 0.5cm 0cm  6.5cm, clip, width=1.0\linewidth]{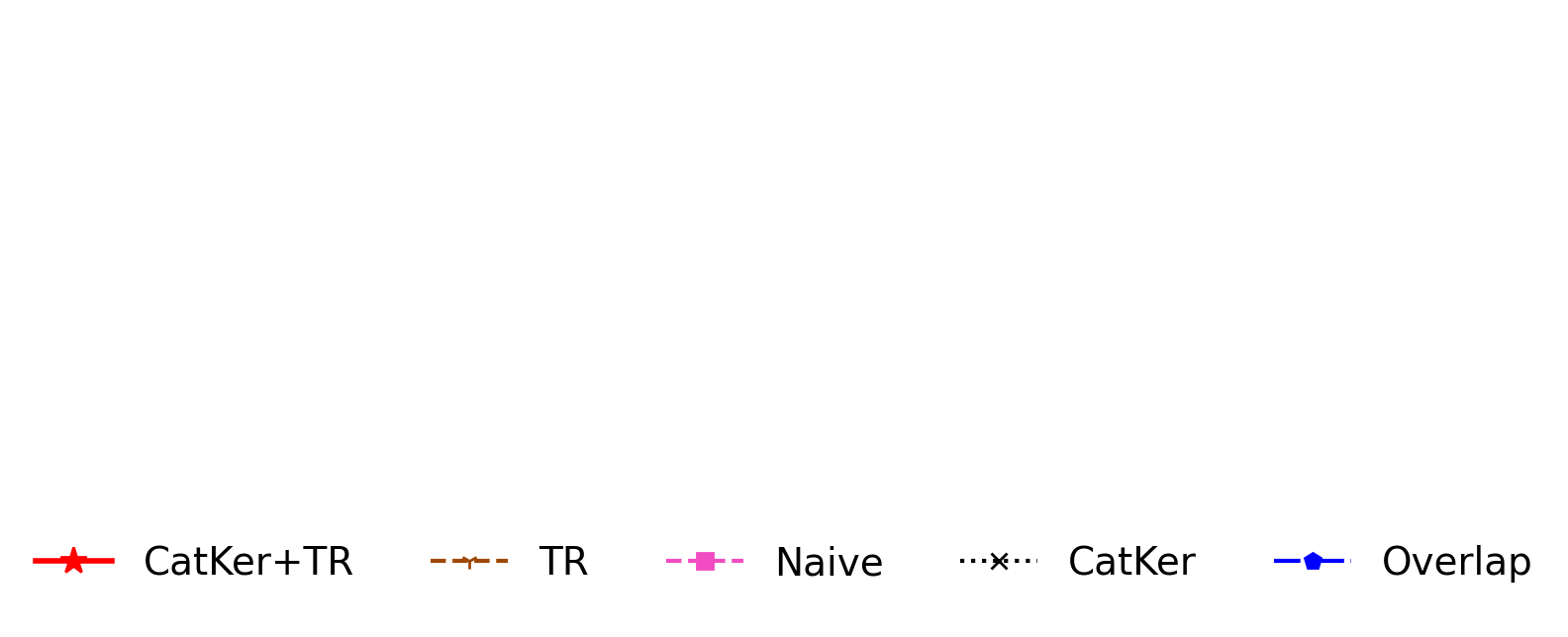}   
    \end{subfigure}

    \caption{Ablation studies of our method in categorical (left) and mixed (right) optimisation problems. First row: Pest (left), Ackley-53 (right); Second row: DifficultPest (left), Rosenbrock-200 (right).
    }

    \label{fig:ablation}
\end{figure}

Our method introduces a number of modifications over the na\"ive \gls{BO} approach.
To understand the benefits of these, we conduct ablation studies in both the categorical and mixed problems. Specifically, we include the following setups.
\begin{itemize}[leftmargin=0.04in, noitemsep, topsep=0.05pt]
    \item  The na\"ive \gls{BO} approach with global \gls{GP} surrogate and one-hot transformation on the categorical variables (\texttt{Naive});
    \item  One-hot transformed \gls{BO}, but with \emph{local} \gls{TR}s i.e. \textsc{turbo} (\texttt{TR});
    \item \gls{GP} with \emph{global} surrogates, but with the categorical overlapping kernel in \citet{cocabo} where applicable (\texttt{Overlap});
    \item \gls{BO} with \emph{global} \gls{GP} surrogate, but with the kernel defined in Eq. (\ref{eq:catkernel}) or (\ref{eq:mixkernel}), where appropriate (\texttt{CatKer});
    \item Our approach that incorporates both local modelling and the kernel in Eq. (\ref{eq:catkernel}) or (\ref{eq:mixkernel}) (\texttt{CatKer+TR}).
\end{itemize}
We firstly include Pest Control and the Ackley-53 problems as representative problems for the categorical and mixed setups for the ablation studies. To further understand the relative importance of the various features of \gls{CASMOPOLITAN} especially as the dimensionality of the problems changes, we also include two even higher-dimensional problems, namely 1) Pest control with number of stages expanded to $80$, which we term DifficultPest (the number of possible configurations is more than $8.27 \times 10^{55}$), and 2) 200-d Rosenbrock with 100 binary dimensions and 100 continuous dimensions (detailed in App. \ref{appendix:details}). 

We show the results in Fig. \ref{fig:ablation}: in most problems, the usage of the categorical kernel leads to improvements over baselines, with kernels used in our method generally outperforming the overlap kernel. Unsurprisingly, the additional benefits of local optimisation and the use of trust regions increase with increasing dimensionality and complexity of the problems, with largest benefits coming from the two high-dimensional problems of the second row. Nonetheless, it is worth noting that even in the relatively modestly-dimensioned Pest Control problem where the difference between  \texttt{CatKer+TR} and \texttt{CatKer} seems small, the outperformance is still statistically significant (Two-sample Student's t-test yields $p = 0.043 < 0.05$ at the final iteration). Finally, our method, similar to \gls{TuRBO}, introduces a number of additional hyperparameters related to the \gls{TR}; we examine the sensitivity of performance towards these extra hyperparameters in App. \ref{appendix:additionalexperiments}.


\section{Conclusion and Future Work}
We propose \gls{CASMOPOLITAN}, a novel \gls{GP}-\gls{BO} approach using ideas of tailored kernels and trust regions to tackle the challenging high-dimensional optimisation problem over categorical and mixed search spaces. We both analyse our method theoretically and empirically demonstrate its effectiveness over a wide range of problems. Possible future directions may extend our model to even more diverse search spaces, such as problems on graphs, trees, and/or in conditional spaces.

\section*{Acknowledgements}
The authors would like to thank the Oxford-Man Institute of Quantitative Finance for providing computing resources in this project. The authors also thank the anonymous ICML reviewers and the area chair for the constructive feedback which helped to improve the paper. 

\bibliographystyle{icml2021.bst}
\bibliography{refs,vunguyen}

\begin{thebibliography}{61}
\providecommand{\natexlab}[1]{#1}
\providecommand{\url}[1]{\texttt{#1}}
\expandafter\ifx\csname urlstyle\endcsname\relax
  \providecommand{\doi}[1]{doi: #1}\else
  \providecommand{\doi}{doi: \begingroup \urlstyle{rm}\Url}\fi

\bibitem[Alzantot et~al.(2019)Alzantot, Sharma, Chakraborty, Zhang, Hsieh, and
  Srivastava]{alzantot2019genattack}
Moustafa Alzantot, Yash Sharma, Supriyo Chakraborty, Huan Zhang, Cho-Jui Hsieh,
  and Mani~B Srivastava.
\newblock Genattack: Practical black-box attacks with gradient-free
  optimization.
\newblock In \emph{Proceedings of the Genetic and Evolutionary Computation
  Conference}, pages 1111--1119, 2019.

\bibitem[Baptista and Poloczek(2018)]{baptista2018bayesian}
Ricardo Baptista and Matthias Poloczek.
\newblock Bayesian optimization of combinatorial structures.
\newblock In \emph{International Conference on Machine Learning}, pages
  462--471. PMLR, 2018.

\bibitem[Bergstra et~al.(2011)Bergstra, Bardenet, Bengio, and
  K{\'{e}}gl]{Bergstra_2011Algorithms}
James Bergstra, R{\'{e}}mi Bardenet, Yoshua Bengio, and Bal{\'{a}}zs
  K{\'{e}}gl.
\newblock Algorithms for hyper-parameter optimization.
\newblock In \emph{Advances in Neural Information Processing Systems}, pages
  2546--2554, 2011.

\bibitem[Berkenkamp et~al.(2019)Berkenkamp, Schoellig, and
  Krause]{Berkenkamp2019}
Felix Berkenkamp, Angela~P. Schoellig, and Andreas Krause.
\newblock No-regret bayesian optimization with unknown hyperparameters.
\newblock \emph{Journal of Machine Learning Research}, 20\penalty0
  (50):\penalty0 1--24, 2019.

\bibitem[Bibby(1974)]{bibby1974axiomatisations}
John Bibby.
\newblock Axiomatisations of the average and a further generalisation of
  monotonic sequences.
\newblock \emph{Glasgow Mathematical Journal}, 15\penalty0 (1):\penalty0
  63--65, 1974.

\bibitem[Binois et~al.(2015)Binois, Ginsbourger, and
  Roustant]{binois2015warped}
Micka{\"e}l Binois, David Ginsbourger, and Olivier Roustant.
\newblock A warped kernel improving robustness in {B}ayesian optimization via
  random embeddings.
\newblock In \emph{International Conference on Learning and Intelligent
  Optimization}, pages 281--286. Springer, 2015.

\bibitem[Binois et~al.(2020)Binois, Ginsbourger, and
  Roustant]{binois2020choice}
Micka{\"e}l Binois, David Ginsbourger, and Olivier Roustant.
\newblock On the choice of the low-dimensional domain for global optimization
  via random embeddings.
\newblock \emph{Journal of global optimization}, 76\penalty0 (1):\penalty0
  69--90, 2020.

\bibitem[Bliek et~al.(2020)Bliek, Verwer, and de~Weerdt]{bliek2020black}
Laurens Bliek, Sicco Verwer, and Mathijs de~Weerdt.
\newblock Black-box mixed-variable optimisation using a surrogate model that
  satisfies integer constraints.
\newblock \emph{arXiv preprint arXiv:2006.04508}, 2020.

\bibitem[Breiman(2001)]{Breiman_2001Random}
Leo Breiman.
\newblock Random forests.
\newblock \emph{Machine learning}, 45\penalty0 (1):\penalty0 5--32, 2001.

\bibitem[Brochu et~al.(2010)Brochu, Cora, and De~Freitas]{Brochu_2010Tutorial}
Eric Brochu, Vlad~M Cora, and Nando De~Freitas.
\newblock A tutorial on {B}ayesian optimization of expensive cost functions,
  with application to active user modeling and hierarchical reinforcement
  learning.
\newblock \emph{arXiv preprint arXiv:1012.2599}, 2010.

\bibitem[Chen and Guestrin(2016)]{chen2016xgboost}
Tianqi Chen and Carlos Guestrin.
\newblock Xgboost: A scalable tree boosting system.
\newblock In \emph{Proceedings of the 22nd ACM SigKDD International Conference
  on Knowledge Discovery and Data Mining}, pages 785--794. ACM, 2016.

\bibitem[Creignou et~al.(2001)Creignou, Khanna, and
  Sudan]{creignou2001complexity}
Nadia Creignou, Sanjeev Khanna, and Madhu Sudan.
\newblock \emph{Complexity classifications of boolean constraint satisfaction
  problems}.
\newblock SIAM, 2001.

\bibitem[Croce et~al.(2020)Croce, Andriushchenko, Singh, Flammarion, and
  Hein]{croce2020sparse}
Francesco Croce, Maksym Andriushchenko, Naman~D Singh, Nicolas Flammarion, and
  Matthias Hein.
\newblock Sparse-rs: a versatile framework for query-efficient sparse black-box
  adversarial attacks.
\newblock \emph{arXiv preprint arXiv:2006.12834}, 2020.

\bibitem[Dadkhahi et~al.(2020)Dadkhahi, Shanmugam, Rios, Das, Hoffman,
  Loeffler, and Sankaranarayanan]{dadkhahi2020combinatorial}
Hamid Dadkhahi, Karthikeyan Shanmugam, Jesus Rios, Payel Das, Samuel~C Hoffman,
  Troy~David Loeffler, and Subramanian Sankaranarayanan.
\newblock Combinatorial black-box optimization with expert advice.
\newblock In \emph{Proceedings of the 26th ACM SIGKDD International Conference
  on Knowledge Discovery \& Data Mining}, pages 1918--1927, 2020.

\bibitem[Dadkhahi et~al.(2021)Dadkhahi, Rios, Shanmugam, and
  Das]{dadkhahi2021fourier}
Hamid Dadkhahi, Jesus Rios, Karthikeyan Shanmugam, and Payel Das.
\newblock Fourier representations for black-box optimization over categorical
  variables.
\newblock 2021.

\bibitem[Daxberger et~al.(2019)Daxberger, Makarova, Turchetta, and
  Krause]{daxberger2019mixed}
Erik Daxberger, Anastasia Makarova, Matteo Turchetta, and Andreas Krause.
\newblock Mixed-variable {B}ayesian optimization.
\newblock \emph{arXiv preprint arXiv:1907.01329}, 2019.

\bibitem[Deshwal et~al.(2020)Deshwal, Belakaria, Doppa, and
  Fern]{Deshwal_Belakaria_Doppa_Fern_2020}
Aryan Deshwal, Syrine Belakaria, Janardhan~Rao Doppa, and Alan Fern.
\newblock Optimizing discrete spaces via expensive evaluations: A learning to
  search framework.
\newblock \emph{Proceedings of the AAAI Conference on Artificial Intelligence},
  34\penalty0 (04):\penalty0 3773--3780, Apr. 2020.

\bibitem[Eriksson et~al.(2019)Eriksson, Pearce, Gardner, Turner, and
  Poloczek]{eriksson2019scalable}
David Eriksson, Michael Pearce, Jacob Gardner, Ryan~D Turner, and Matthias
  Poloczek.
\newblock Scalable global optimization via local {B}ayesian optimization.
\newblock In \emph{Advances in Neural Information Processing Systems}, pages
  5496--5507, 2019.

\bibitem[Frazier(2018)]{frazier2018tutorial}
Peter~I Frazier.
\newblock A tutorial on {B}ayesian optimization.
\newblock \emph{arXiv preprint arXiv:1807.02811}, 2018.

\bibitem[Gardner et~al.(2018)Gardner, Pleiss, Bindel, Weinberger, and
  Wilson]{gardner2018gpytorch}
Jacob~R Gardner, Geoff Pleiss, David Bindel, Kilian~Q Weinberger, and
  Andrew~Gordon Wilson.
\newblock Gpytorch: Blackbox matrix-matrix gaussian process inference with gpu
  acceleration.
\newblock \emph{arXiv preprint arXiv:1809.11165}, 2018.

\bibitem[Garrido-Merch{\'a}n and
  Hern{\'a}ndez-Lobato(2020)]{garrido2020dealing}
Eduardo~C Garrido-Merch{\'a}n and Daniel Hern{\'a}ndez-Lobato.
\newblock Dealing with categorical and integer-valued variables in {B}ayesian
  optimization with {G}aussian processes.
\newblock \emph{Neurocomputing}, 380:\penalty0 20--35, 2020.

\bibitem[Ginsbourger et~al.(2010)Ginsbourger, Le~Riche, and
  Carraro]{ginsbourger2010kriging}
David Ginsbourger, Rodolphe Le~Riche, and Laurent Carraro.
\newblock Kriging is well-suited to parallelize optimization.
\newblock In \emph{Computational intelligence in expensive optimization
  problems}, pages 131--162. Springer, 2010.

\bibitem[Gopakumar et~al.(2018)Gopakumar, Gupta, Rana, Nguyen, and
  Venkatesh]{gopakumar2018algorithmic_NIPS}
Shivapratap Gopakumar, Sunil Gupta, Santu Rana, Vu~Nguyen, and Svetha
  Venkatesh.
\newblock Algorithmic assurance: An active approach to algorithmic testing
  using {B}ayesian optimisation.
\newblock In \emph{Advances in Neural Information Processing Systems
  (NeurIPS)}, pages 5465--5473, 2018.

\bibitem[GPyOpt(2016)]{gpyopt2016}
GPyOpt.
\newblock {GPyOpt}: A {B}ayesian optimization framework in python.
\newblock \url{http://github.com/SheffieldML/GPyOpt}, 2016.

\bibitem[Hern{\'a}ndez-Lobato et~al.(2017)Hern{\'a}ndez-Lobato, Requeima,
  Pyzer-Knapp, and Aspuru-Guzik]{Hernandez_2017Parallel}
Jos{\'e}~Miguel Hern{\'a}ndez-Lobato, James Requeima, Edward~O Pyzer-Knapp, and
  Al{\'a}n Aspuru-Guzik.
\newblock Parallel and distributed {T}hompson sampling for large-scale
  accelerated exploration of chemical space.
\newblock \emph{In International Conference on Machine Learning}, pages
  1470--1479, 2017.

\bibitem[Hu et~al.(2010)Hu, Hu, Xu, Wang, and Cao]{hu2010contamination}
Yingjie Hu, JianQiang Hu, Yifan Xu, Fengchun Wang, and Rong~Zeng Cao.
\newblock Contamination control in food supply chain.
\newblock In \emph{Proceedings of the 2010 Winter Simulation Conference}, pages
  2678--2681. IEEE, 2010.

\bibitem[Hutter et~al.(2011)Hutter, Hoos, and
  Leyton-Brown]{Hutter_2011Sequential}
Frank Hutter, Holger~H Hoos, and Kevin Leyton-Brown.
\newblock Sequential model-based optimization for general algorithm
  configuration.
\newblock In \emph{Learning and Intelligent Optimization}, pages 507--523.
  Springer, 2011.

\bibitem[Jones et~al.(1998)Jones, Schonlau, and Welch]{Jones_1998Efficient}
Donald~R Jones, Matthias Schonlau, and William~J Welch.
\newblock Efficient global optimization of expensive black-box functions.
\newblock \emph{Journal of Global optimization}, 13\penalty0 (4):\penalty0
  455--492, 1998.

\bibitem[Kandasamy et~al.(2015)Kandasamy, Schneider, and
  P{\'o}czos]{kandasamy2015high}
Kirthevasan Kandasamy, Jeff Schneider, and Barnab{\'a}s P{\'o}czos.
\newblock High dimensional {B}ayesian optimisation and bandits via additive
  models.
\newblock In \emph{International Conference on Machine Learning}, pages
  295--304, 2015.

\bibitem[Kandasamy et~al.(2018)Kandasamy, Neiswanger, Schneider, Poczos, and
  Xing]{kandasamy2018neural}
Kirthevasan Kandasamy, Willie Neiswanger, Jeff Schneider, Barnabas Poczos, and
  Eric~P Xing.
\newblock Neural architecture search with bayesian optimisation and optimal
  transport.
\newblock In \emph{Advances in Neural Information Processing Systems}, pages
  2016--2025, 2018.

\bibitem[Kim and Fessler(2018)]{kim2018adaptive}
Donghwan Kim and Jeffrey~A Fessler.
\newblock Adaptive restart of the optimized gradient method for convex
  optimization.
\newblock \emph{Journal of Optimization Theory and Applications}, 178\penalty0
  (1):\penalty0 240--263, 2018.

\bibitem[Kingma and Ba(2015)]{kingma2014adam}
Diederik~P Kingma and Jimmy Ba.
\newblock Adam: A method for stochastic optimization.
\newblock \emph{International Conference on Learning Representations}, 2015.

\bibitem[Kondor and Lafferty(2002)]{kondor2002diffusion}
Risi Kondor and John~D. Lafferty.
\newblock Diffusion kernels on graphs and other discrete input spaces.
\newblock In \emph{International Conference on Machine Learning}, pages
  315--322, 2002.

\bibitem[Krause and Ong(2011)]{Krause_2011Contextual}
Andreas Krause and Cheng~S Ong.
\newblock Contextual {G}aussian process bandit optimization.
\newblock In \emph{Advances in Neural Information Processing Systems}, pages
  2447--2455, 2011.

\bibitem[LeCun(1998)]{lecun1998mnist}
Yann LeCun.
\newblock The mnist database of handwritten digits.
\newblock \emph{http://yann. lecun. com/exdb/mnist/}, 1998.

\bibitem[Letham et~al.(2020)Letham, Calandra, Rai, and Bakshy]{letham2020re}
Ben Letham, Roberto Calandra, Akshara Rai, and Eytan Bakshy.
\newblock Re-examining linear embeddings for high-dimensional bayesian
  optimization.
\newblock \emph{Advances in Neural Information Processing Systems}, 33, 2020.

\bibitem[Mazya and Shaposhnikova(1999)]{Vladimir1999Hadamard}
Vladimir Mazya and Tatyana Shaposhnikova.
\newblock \emph{Jacques Hadamard: A Universal Mathematician}.
\newblock 1st edition, 1999.

\bibitem[Mutn{\`y} and Krause(2019)]{mutny2019efficient}
Mojm{\'\i}r Mutn{\`y} and Andreas Krause.
\newblock Efficient high dimensional {B}ayesian optimization with additivity
  and quadrature fourier features.
\newblock \emph{Advances in Neural Information Processing Systems 31}, pages
  9005--9016, 2019.

\bibitem[Nayebi et~al.(2019)Nayebi, Munteanu, and
  Poloczek]{nayebi2019framework}
Amin Nayebi, Alexander Munteanu, and Matthias Poloczek.
\newblock A framework for {B}ayesian optimization in embedded subspaces.
\newblock In \emph{International Conference on Machine Learning}, pages
  4752--4761. PMLR, 2019.

\bibitem[Nguyen et~al.(2020)Nguyen, Gupta, Rana, Shilton, and
  Venkatesh]{nguyen2020bayesian}
Dang Nguyen, Sunil Gupta, Santu Rana, Alistair Shilton, and Svetha Venkatesh.
\newblock {B}ayesian optimization for categorical and category-specific
  continuous inputs.
\newblock In \emph{Proceedings of the AAAI Conference on Artificial
  Intelligence}, volume~34, pages 5256--5263, 2020.

\bibitem[Nguyen et~al.(2021)Nguyen, Le, Yamada, and Osborne]{nguyen2020optimal}
Vu~Nguyen, Tam Le, Makoto Yamada, and Michael~A Osborne.
\newblock Optimal transport kernels for sequential and parallel neural
  architecture search.
\newblock In \emph{International Conference on Machine Learning}, 2021.

\bibitem[Oh et~al.(2019)Oh, Tomczak, Gavves, and Welling]{oh2019combinatorial}
Changyong Oh, Jakub Tomczak, Efstratios Gavves, and Max Welling.
\newblock Combinatorial {B}ayesian optimization using the graph cartesian
  product.
\newblock In \emph{Advances in Neural Information Processing Systems}, pages
  2914--2924, 2019.

\bibitem[Parker-Holder et~al.(2020)Parker-Holder, Nguyen, and Roberts]{pb2}
Jack Parker-Holder, Vu~Nguyen, and Stephen~J Roberts.
\newblock Provably efficient online hyperparameter optimization with
  population-based bandits.
\newblock \emph{Advances in Neural Information Processing Systems}, 33, 2020.

\bibitem[Rana et~al.(2017)Rana, Li, Gupta, Nguyen, and
  Venkatesh]{Rana_ICML2017High}
Santu Rana, Cheng Li, Sunil Gupta, Vu~Nguyen, and Svetha Venkatesh.
\newblock High dimensional {B}ayesian optimization with elastic gaussian
  process.
\newblock In \emph{Proceedings of the 34th International Conference on Machine
  Learning (ICML)}, pages 2883--2891, 2017.

\bibitem[Rasmussen(2006)]{Rasmussen_2006gaussian}
Carl~Edward Rasmussen.
\newblock Gaussian processes for machine learning.
\newblock 2006.

\bibitem[Rolland et~al.(2018)Rolland, Scarlett, Bogunovic, and
  Cevher]{rolland2018high}
Paul Rolland, Jonathan Scarlett, Ilija Bogunovic, and Volkan Cevher.
\newblock High-dimensional {B}ayesian optimization via additive models with
  overlapping groups.
\newblock In \emph{International conference on artificial intelligence and
  statistics}, pages 298--307. PMLR, 2018.

\bibitem[Ru et~al.(2020{\natexlab{a}})Ru, Alvi, Nguyen, Osborne, and
  Roberts]{cocabo}
Binxin Ru, Ahsan Alvi, Vu~Nguyen, Michael~A Osborne, and Stephen Roberts.
\newblock {B}ayesian optimisation over multiple continuous and categorical
  inputs.
\newblock In \emph{International Conference on Machine Learning}, pages
  8276--8285. PMLR, 2020{\natexlab{a}}.

\bibitem[Ru et~al.(2020{\natexlab{b}})Ru, Cobb, Blaas, and Gal]{ru2019bayesopt}
Binxin Ru, Adam Cobb, Arno Blaas, and Yarin Gal.
\newblock Bayesopt adversarial attack.
\newblock In \emph{International Conference on Learning Representations},
  2020{\natexlab{b}}.

\bibitem[Ru et~al.(2021)Ru, Wan, Dong, and Osborne]{ru2020neural}
Binxin Ru, Xingchen Wan, Xiaowen Dong, and Michael Osborne.
\newblock Interpretable neural architecture search via {B}ayesian optimisation
  with weisfeiler-lehman kernels.
\newblock \emph{International Conference on Learning Representations}, 2021.

\bibitem[Shahriari et~al.(2016)Shahriari, Swersky, Wang, Adams, and
  de~Freitas]{Shahriari_2016Taking}
Bobak Shahriari, Kevin Swersky, Ziyu Wang, Ryan~P Adams, and Nando de~Freitas.
\newblock Taking the human out of the loop: A review of {B}ayesian
  optimization.
\newblock \emph{Proceedings of the IEEE}, 104\penalty0 (1):\penalty0 148--175,
  2016.

\bibitem[Shylo et~al.(2011)Shylo, Middelkoop, and Pardalos]{shylo2011restart}
Oleg~V Shylo, Timothy Middelkoop, and Panos~M Pardalos.
\newblock Restart strategies in optimization: parallel and serial cases.
\newblock \emph{Parallel Computing}, 37\penalty0 (1):\penalty0 60--68, 2011.

\bibitem[Snoek et~al.(2012)Snoek, Larochelle, and Adams]{Snoek_2012Practical}
Jasper Snoek, Hugo Larochelle, and Ryan~P Adams.
\newblock Practical {B}ayesian optimization of machine learning algorithms.
\newblock In \emph{Advances in Neural Information Processing Systems}, pages
  2951--2959, 2012.

\bibitem[Srinivas et~al.(2010)Srinivas, Krause, Kakade, and
  Seeger]{Srinivas_2010Gaussian}
Niranjan Srinivas, Andreas Krause, Sham Kakade, and Matthias Seeger.
\newblock Gaussian process optimization in the bandit setting: No regret and
  experimental design.
\newblock In \emph{Proceedings of the 27th International Conference on Machine
  Learning}, pages 1015--1022, 2010.

\bibitem[Swersky et~al.(2020)Swersky, Rubanova, Dohan, and
  Murphy]{swersky2020amortized}
Kevin Swersky, Yulia Rubanova, David Dohan, and Kevin Murphy.
\newblock Amortized bayesian optimization over discrete spaces.
\newblock In \emph{Conference on Uncertainty in Artificial Intelligence}, pages
  769--778. PMLR, 2020.

\bibitem[Sylvester(1851)]{sylvester1851det}
James~Joseph Sylvester.
\newblock Xxxvii. on the relation between the minor determinants of linearly
  equivalent quadratic functions.
\newblock \emph{The London, Edinburgh, and Dublin Philosophical Magazine and
  Journal of Science}, 1\penalty0 (4):\penalty0 295--305, 1851.
\newblock \doi{10.1080/14786445108646735}.

\bibitem[Tu et~al.(2019)Tu, Ting, Chen, Liu, Zhang, Yi, Hsieh, and
  Cheng]{tu2019autozoom}
Chun-Chen Tu, Paishun Ting, Pin-Yu Chen, Sijia Liu, Huan Zhang, Jinfeng Yi,
  Cho-Jui Hsieh, and Shin-Ming Cheng.
\newblock Autozoom: Autoencoder-based zeroth order optimization method for
  attacking black-box neural networks.
\newblock In \emph{Proceedings of the AAAI Conference on Artificial
  Intelligence}, volume~33, pages 742--749, 2019.

\bibitem[Wang et~al.(2017)Wang, Li, Jegelka, and Kohli]{wang2017batched}
Zi~Wang, Chengtao Li, Stefanie Jegelka, and Pushmeet Kohli.
\newblock Batched high-dimensional {B}ayesian optimization via structural
  kernel learning.
\newblock In \emph{International Conference on Machine Learning}, pages
  3656--3664. PMLR, 2017.

\bibitem[Wang et~al.(2018)Wang, Gehring, Kohli, and Jegelka]{wang2018batched}
Zi~Wang, Clement Gehring, Pushmeet Kohli, and Stefanie Jegelka.
\newblock Batched large-scale {B}ayesian optimization in high-dimensional
  spaces.
\newblock In \emph{International Conference on Artificial Intelligence and
  Statistics}, pages 745--754. PMLR, 2018.

\bibitem[Wang et~al.(2013)Wang, Zoghi, Hutter, Matheson, Freitas,
  et~al.]{Wang_2013Bayesian}
Ziyu Wang, Masrour Zoghi, Frank Hutter, David Matheson, N~Freitas, et~al.
\newblock Bayesian optimization in high dimensions via random embeddings.
\newblock AAAI Press/International Joint Conferences on Artificial
  Intelligence, 2013.

\bibitem[Wang et~al.(2016)Wang, Hutter, Zoghi, Matheson, and
  de~Feitas]{wang2016bayesian}
Ziyu Wang, Frank Hutter, Masrour Zoghi, David Matheson, and Nando de~Feitas.
\newblock {B}ayesian optimization in a billion dimensions via random
  embeddings.
\newblock \emph{Journal of Artificial Intelligence Research}, 55:\penalty0
  361--387, 2016.

\bibitem[Yuan(2000)]{yuan2000review}
Ya-xiang Yuan.
\newblock A review of trust region algorithms for optimization.
\newblock In \emph{Iciam}, volume~99, pages 271--282. Citeseer, 2000.

\end{thebibliography}

\newpage
\appendix
\clearpage
\section*{Appendix}
\section{Primer on GP and BO}
\label{app:primer_on_gp}

\paragraph{Gaussian processes}
We consider a \gls{GP} surrogate model for a black-box function $f$ which takes an input  $\bz =[\bh,\bx]$ and returns an output $y=f(\bz) + \epsilon$ where $\epsilon \sim \mathcal{N}(0,\sigma^2)$. Here, the input includes a continuous variable $\bx$ and a categorical variable $\bh$. A \gls{GP} defines a probability distribution over functions $f$ under the assumption that any finite subset $\lbrace (\bz_i, f(\bz_i) \rbrace$ follows a normal distribution \cite{Rasmussen_2006gaussian}. Formally, a \gls{GP} is denoted as $f(\bz)\sim \text{GP}\left(m\left(\bz\right),k\left(\bz,\bz'\right)\right)$, where $m\left(\bz\right)$ and $k\left(\bz,\bz'\right)$ are called the mean and covariance functions respectively, i.e. $m(\bz)=\mathbb{E}\left[f\left(\bz\right)\right]$ and $k(\bz,\bz')=\mathbb{E}\left[(f\left(\bz\right)-m\left(\bz\right))(f\left(\bz'\right)-m\left(\bz'\right))^{T}\right]$. The covariance function (kernel) $k(\bz,\bz')$ can be thought of as a similarity measure relating $f(\bz)$ and $f(\bz')$. There have been various proposed kernels which encode different prior beliefs about the function $f(\bz)$, typically in the continuous space. Popular choices include the Square Exponential kernel, the Mat\'{e}rn kernel \cite{Rasmussen_2006gaussian}.

Assume the zero mean prior $m(\bz)=0$, to predict $f_{*}=f\left(\bz_{*}\right)$ at a new data point $\bz_{*}$, we have,
\begin{align}
\left[\begin{array}{c}
\boldsymbol{f}\\
f_{*}
\end{array}\right] & \sim\mathcal{N}\left(0,\left[\begin{array}{cc}
\bK & \bk_{*}^{T}\\
\bk_{*} & k_{**}
\end{array}\right]\right),\label{eq:p(f|f*)}
\end{align}
 where $k_{**}=k\left(\bz_{*},\bz_{*}\right)$, $\bk_{*}=[k\left(\bz_{*},\bz_{i}\right)]_{\forall i\le N}$
and $\bK=\left[k\left(\bz_{i},\bz_{j}\right)\right]_{\forall i,j\le N}$. Combining Eq. (\ref{eq:p(f|f*)}) with the fact that $p\left(f_{*}\mid\boldsymbol{f}\right)$ follows a univariate Gaussian distribution $\mathcal{N}\left(\mu\left(\bz_{*}\right),\sigma^{2}\left(\bz_{*}\right)\right)$, the \gls{GP} posterior mean and variance can be computed as,
\begin{align*}
\mu\left(\bz_{*}\right)= & \mathbf{k}_{*} \left[ \mathbf{K} + \sigma^2 \idenmat \right]^{-1}\mathbf{y},\\
\sigma^{2}\left(\bz_{*}\right)= & k_{**}-\mathbf{k}_{*} \left[ \mathbf{K} + \sigma^2 \idenmat \right]^{-1} \mathbf{k}_{*}^{T}.
\end{align*}
As \gls{GP}s give full uncertainty information with any prediction, they provide a flexible nonparametric prior for Bayesian optimisation. We refer the interested readers to \citet{Rasmussen_2006gaussian} for further details on \gls{GP}s.

\paragraph{Bayesian optimisation}

Bayesian optimisation is a powerful sequential approach to find the global optimum of an expensive black-box function $f(\mathbf{z})$ without making use of derivatives. First, a surrogate model is learned from all the current observed data $\mathcal{D}_t= \lbrace \mathbf{z}_i, y_i \rbrace_{i=1}^t$ to approximate the behavior of $f(\mathbf{z})$. Second, an acquisition function is derived from the surrogate model to select new data points that mostly inform about the global optimum. The process is conducted iteratively until the evaluation budget is depleted, and the global optimum is estimated based on all the sampled data. In-depth discussions about Bayesian optimisation beyond this brief overview can be found in recent surveys \cite{Brochu_2010Tutorial,Shahriari_2016Taking,frazier2018tutorial}.


\section{Additional Experimental Results}
\label{appendix:additionalexperiments}

\subsection{Running Time Comparison}
\label{subsec:runtime}

In this section, we provide comparison of \gls{CASMOPOLITAN} against some baselines in terms of wall-clock running time on a number of problems considered. However, since we conduct our experiments on a shared server, inevitably there are fluctuations in wall clock time depending on the server load, leading to (perhaps rather significant) amount of uncertainty over the computing time reported here and thus, the figures here are for ballpark reference only. 
\gls{CASMOPOLITAN} scales $\mathcal{O}(N^3)$, where $N$ here refers to the number of training data (note that in \gls{CASMOPOLITAN}, this is not necessarily the total number of observations, but only the number of training samples of the \gls{GP} surrogate of the current restart), which is the time complexity of any \gls{GP}-\gls{BO} method where the computational bottleneck is the inversion of the covariance matrix \cite{Shahriari_2016Taking}. Practically, due the implementation in Gpytorch which utilises Black-box Matrix-matrix (\textsc{bbmm}) multiplication which reduces the cost of exact \gls{GP} inference to $\mathcal{O}(N^2)$ \cite{gardner2018gpytorch}. Overall, the computing cost of \gls{CASMOPOLITAN} is generally comparable to \gls{TuRBO}. On the other hand, previous methods generally scale worse. For example, in addition to the inherent $\mathcal{O}(N^3)$ complexity (or $\mathcal{O}(N^2)$ if \textsc{bbmm} is similarly exploited), \gls{COMBO} additionally incurs the cost in the graph Fourier transform of $\mathcal{O}(\sum_{i=1}^{d_h} n_h^3)$ using the notations of our paper (Proposition 2.3.1 in \citet{oh2019combinatorial}). Furthermore, it also uses slice sampling for the approximate marginalisation of the posterior predictive distribution, which is arguably more expensive than simple optimisation of the log marginal-likelihood. A single iteration of \textsc{bocs} incurs complexity of $\mathcal{O}(N^2{d_h}^2)$ \cite{baptista2018bayesian}, suggesting that the runtime of \textsc{BOCS} quadratically also with respect to the \emph{dimensionality} of the problem. Furthermore, it is worth noting that the quadratic dependence of $d_h$ stems from the second order \emph{approximation} of their sparse Bayesian linear regression model. This term will become much more expensive if a higher order approximation is used, e.g. it becomes $d_h^m$ if an $m$-th order approximation is used. 

For the categorical problems, \gls{COMBO} achieves comparable performance in terms of the function value at termination in 2 out of 3 problems in Fig. \ref{fig:categoricalproblems} and thus the analysis of computing cost against it is of our prime interest (other methods are either not competitive in terms of performance (e.g. \textsc{tpe}), or are \emph{much} more expensive and/or more constrained in applicability (e.g. \textsc{bocs}). The comparison against \gls{COMBO} (and \textsc{bocs} where applicable) is shown in Table \ref{tab:categoricaltimes}, where it is evident that our method offers around $2$-$3$ times speedup compared to \gls{COMBO}, whereas \textsc{bocs} is orders-of-magnitude more expensive.

\begin{table}[h]
    \centering
        \caption{Wall-clock time comparison of a single trial (mean $\pm$ standard deviation across 20 trials for \gls{COMBO} and \gls{CASMOPOLITAN}) on categorical problems on a shared Intel Xeon server. }
        \begin{footnotesize}
    \begin{tabular}{llll}
    \toprule
    Problem & Ours & \gls{COMBO} & \textsc{bocs}\\
    \midrule
    Pest & $\mathbf{1130}_{\pm80}$s & $3330_{\pm50}$s & $\sim 1$d\\
    Contamination & $\mathbf{2350}_{\pm180}$s & $6630_{\pm400}$s & n.s.\\
    \textsc{maxsat-60} & $\mathbf{12100}_{\pm3000}$s & $34300_{\pm2000}$s & o.o.t \\
    \bottomrule 
    \multicolumn{4}{l}{n.s: setup not supported} \\ 
    \multicolumn{4}{l}{o.o.t: run out-of-time ($>100$ hours) and did not finish.} \\
    \end{tabular}
    \label{tab:categoricaltimes}
    \end{footnotesize}
\end{table}

For the mixed problems, we analyse the black-box attack problem. On average, the time taken to attack an image (successful or not) is around $45$ mins with \gls{MVRSM}, which uses ReLU surrogate instead of \gls{GP}, whereas \gls{TuRBO} and our \gls{CASMOPOLITAN} run roughly $1.6\times$ and $2.0\times$ more expensive -- \gls{TuRBO} run faster likely due to its more frequent restarts. However, in realistic setups suitable for \gls{BO} where either the objective function evaluation time completely eclipses the algorithm running time (e.g. tuning of large-scale machine learning system) or where \emph{sample efficiency}, as contrasted \emph{wall-clock efficiency}, is otherwise more valued (e.g. the black-box attack setup discussed here), the larger cost of \gls{CASMOPOLITAN} is likely justifies by its better performance. Overall, we believe that \gls{CASMOPOLITAN} offers a sound balance between good performance and reasonable computing cost.

\subsection{Additional Problems}
\label{subsec:otherexperiments}

\begin{figure}[h]
    \centering
     \begin{subfigure}{0.5\linewidth}
    \includegraphics[trim=0cm 0cm 0cm  0cm, clip, width=1.0\linewidth]{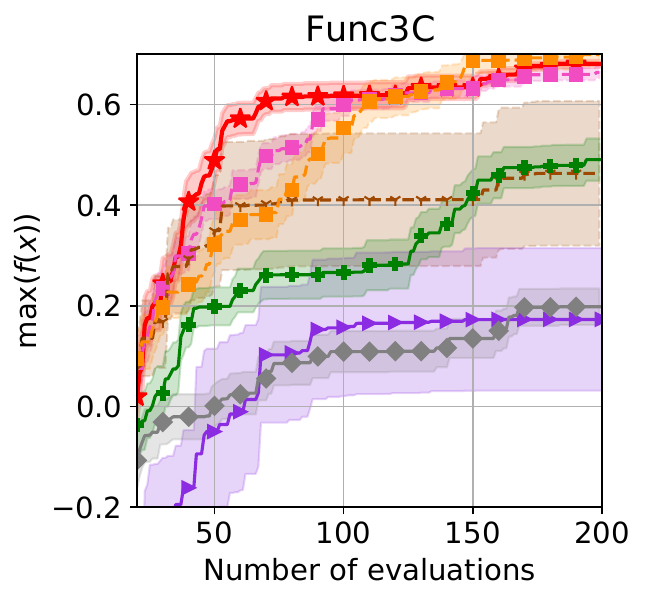}
    \end{subfigure}
        \caption{Results on Func3C}  
    \label{fig:func3C}
\end{figure}

\paragraph{Func3C} The results are shown in Fig. \ref{fig:func3C}. The results are broadly comparable to that of Func2C in Fig. \ref{fig:mixedproblems}(a), although in this case \gls{CoCaBO} and vanilla-\gls{BO} perform more strongly near the end.

\begin{figure}[h]
    \centering
     \begin{subfigure}{0.5\linewidth}
    \includegraphics[trim=0cm 0cm 0cm  0cm, clip, width=1.0\linewidth]{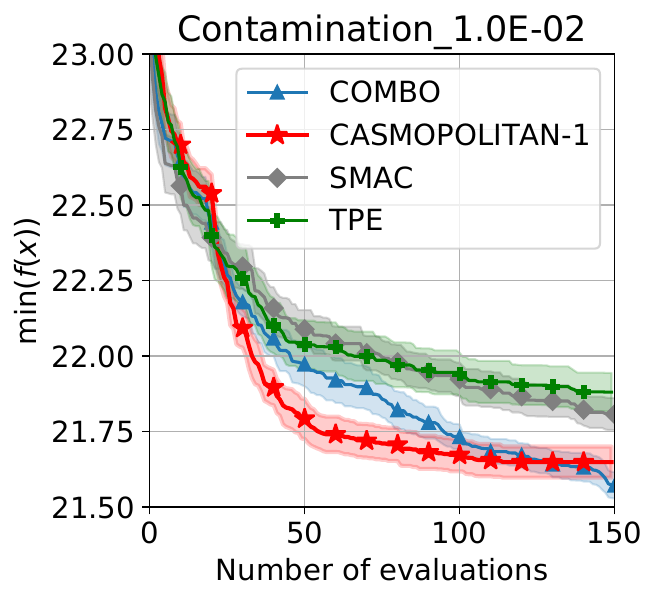}
    \end{subfigure}
        \caption{Results on Contamination problem with noise variance of $0.01$.}  
    \label{fig:noisy_contamination}
\end{figure}

\paragraph{Noisy Contamination} We conduct a further experiment on the Contamination problem but with an additional noise variance of $1 \times 10^{-2}$, and the results are shown in Fig. \ref{fig:noisy_contamination}. In this case, we again see \gls{CASMOPOLITAN} and \gls{COMBO} outperforming the rest and \gls{CASMOPOLITAN} again enjoys a faster convergence than the other methods. In this particular case, \gls{COMBO} outperforms \gls{CASMOPOLITAN}, albeit marginally, at the end.

\paragraph{Ordinal problems} Sometimes we encounter ordinal problems, which are discrete variables that are similar to the categorical problems, but unlike categorical, there exists some kind of ordering between the different values that a variable can take. For example, in deep learning we often choose batch size as powers of $2$, where possible batch sizes are $\{64, 128, 256, 512\}$. While current methods and popular packages (e.g. \gls{CoCaBO} and the Bayesmark\footnote{\url{https://github.com/uber/bayesmark}} package) often treat these as ordinary categorical variables by ignoring such ordinal structure, this practice might not be optimal. In this section, we describe an exemplary adaptation of \gls{CASMOPOLITAN} in the ordinal setting that recognises and leverage such relations, and conduct a preliminary experiment to validate it as a demonstration of the versatility of our approach.

In our tailored kernel for the categorical variables (Eq. (\ref{eq:catkernel})), we use Kronecker delta function which only has two possible outcomes: $0$ if the two values are different or $1$ if the two values are the same. This is appropriate in the categorical setting because there exists no ordering amongst different choices a variable may take (e.g. consider choosing from \{\textsc{sgd}, Adam, \textsc{rmsprop}\}: to \textsc{sgd}, Adam can be considered ``as different as''  \textsc{rmsprop}. However, in ordinal-structured problems such as the batch size example above, the choice of $128$ is certainly ``more similar'' to $64$ than $256$. To recognise this, we modify Eq. (\ref{eq:catkernel}), reproduced below for convenience:
\begin{equation*}
    k_h (\mathbf{h}, \mathbf{h}') = \exp \Big( \frac{1}{d_n} \sum_{i=1}^{d_h} \ell_i \delta(h_i, h'_i) \Big).
\end{equation*}
For ordinal variables, we use the \emph{ordinal kernel} $k_o$ by replacing the Kronecker delta function with an appropriate distance metric. One possible formulation is:
\begin{equation}
    k_o (\mathbf{h}, \mathbf{h}') = \exp \Big( \frac{1}{d_n} \sum_{i=1}^{d_h} \ell_i \big(1 - \frac{|h_i - h'_i|}{|h_i - h'_i|_{\max}}\big) \Big),
\label{eq:ordinalkernel}
\end{equation}
where $|h_i - h'_i|$ is the distance that is dependent on the problem-specific metric and $|h_i - h'_i|_{\max}$ is the maximum possible distance (in the context of the batch size problem, this is $512 - 64$). Note that when no ordinal structure exists, $|h_i - h'_i|$ is either $0$ or $|h_i - h'_i|_{\max}$ and Eq. (\ref{eq:ordinalkernel}) reduces to the categorical kernel in Eq. (\ref{eq:catkernel}).

\begin{figure}[h]
    \centering
     \begin{subfigure}{0.54\linewidth}
    \includegraphics[trim=0cm 0cm 0cm  0cm, clip, width=1.0\linewidth]{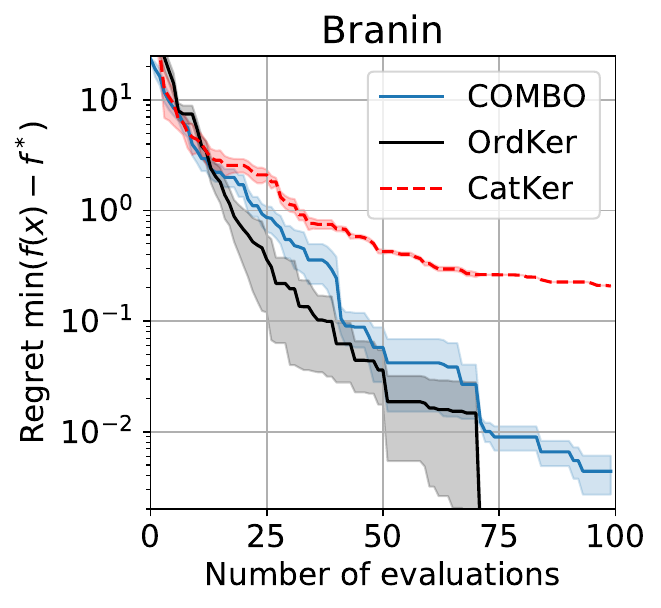}
    \end{subfigure}
        \caption{Results on Discretised Branin. Lines and shades denote mean and standard error across 20 trials. Note that since the optimum is known in this case ($f^* = 0.404$), in the y-axis we show the regret in log-scale.}  
    \label{fig:branin}
\end{figure}

We further include a preliminary empirical validation on the $2$D discretised Branin problem introduced in \citet{oh2019combinatorial}, where each dimension of the Branin function in $[-1, 1]^2$ is discretised into $51$ equally spaced points -- as such, the problem has $2$ ordinal dimensions with $51$ choices for each. Note that this is a rather extreme example due to the large number of choices relative to the number of variables, and the fact that the function landscape resembles much more to a continuous problem instead of a typical ordinal one, but we include it for the sake of illustration. We also do not use trust region for this example due to the low dimensionality and the fact that the point of this experiment is to compare categorical and ordinal kernels. We show the results in Fig. \ref{fig:branin}, where we also include the results for \gls{COMBO} which is the strongest baseline shown to outperform other methods such as \textsc{smac} and \textsc{tpe} in \citet{oh2019combinatorial}. It is worth emphasising that \gls{COMBO} also explicitly accounts for the ordinal relations, so the comparison of it against \gls{CASMOPOLITAN} with ordinal kernel is fair.

\begin{figure}[h]
    \centering
     \begin{subfigure}{0.44\linewidth}
    \includegraphics[trim=0cm 0cm 0cm  0cm, clip, width=1.0\linewidth]{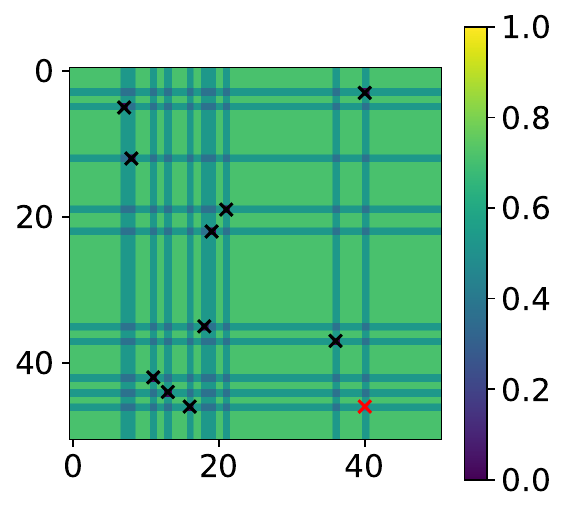}
    \caption{Categorical kernel $k_h$}
    \end{subfigure}
     \begin{subfigure}{0.44\linewidth}
    \includegraphics[trim=0cm 0cm 0cm  0cm, clip, width=1.0\linewidth]{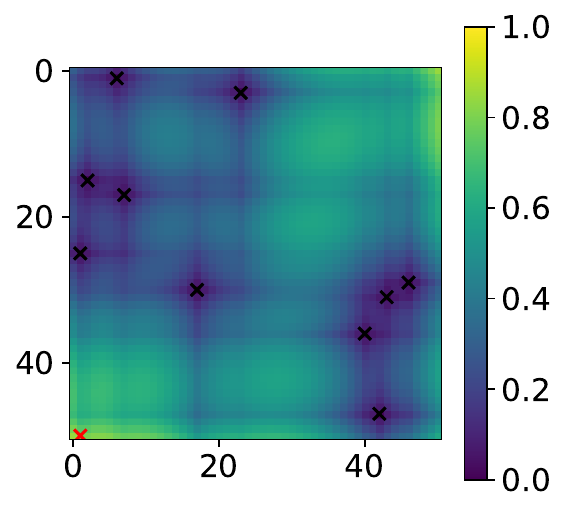}
    \caption{Ordinal kernel $k_o$}
    \end{subfigure}
    \caption{Posterior variance of GP with categorical and ordinal kernels after 10 random initial points on the discretised Branin problem. Black markers are the GP observations; \textcolor{red}{red markers} are the proposed locations for the next evaluations.
    }    
    \label{fig:braninposteriorvar}
\end{figure}

It is clear that \gls{CASMOPOLITAN} with ordinal kernel (\texttt{OrdKer}) outperforms both the ordinal-agnostic \gls{CASMOPOLITAN} (\texttt{CatKer}) and the ordinal-aware \gls{COMBO} in both convergence speed and final performance (\texttt{OrdKer} converges to $f^*$ every single trial). To show why it is the case, we plot the GP posterior variance of \gls{CASMOPOLITAN} with each kernel in Fig. \ref{fig:braninposteriorvar}: categorical kernel measures similarity via the Hamming distances only, and thus each observation $\mathbf{h}_i$ only reduces posterior variance on the points sharing at least one common dimension as $\mathbf{h}_i$. On the other hand, ordinal kernel further accounts for the similarity amongst different values an input may take, and thus each evaluation also reduces the variance in the vicinity of $\mathbf{h}_i$.

While we only consider a toy problem here, the fact that $k_o$ is a simple modification over $k_h$ means it is trivial to scale the approach to high dimensions with the local \gls{TR} approaches described in the main text and/or to the mixed inputs, such as ordinal-continuous or even ordinal-categorical-continuous search space. We defer a thorough investigation to this even richer class of problems to a future work, which we believe would be an exciting extension to the present work.

\subsection{Parallel \gls{CASMOPOLITAN} by number of objective function queries} 
\label{subsec:parallel_by_func_queries}
\begin{figure}[h]
    \centering
     \begin{subfigure}{0.44\linewidth}
    \includegraphics[trim=0cm 0cm 0cm  0cm, clip, width=1.0\linewidth]{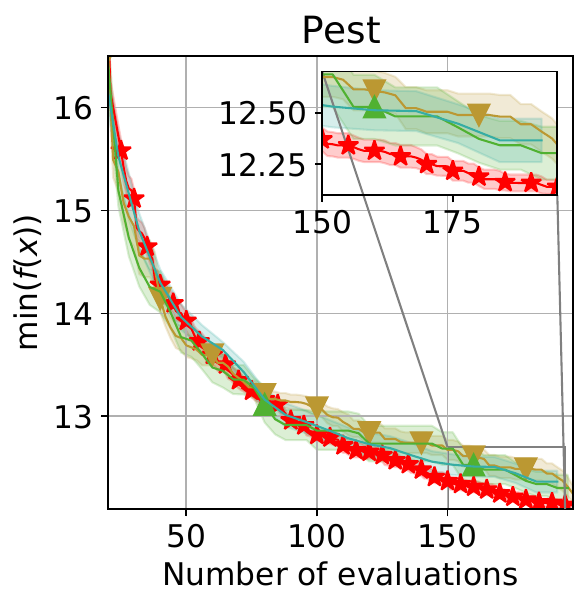}
    \end{subfigure}
     \begin{subfigure}{0.44\linewidth}
    \includegraphics[trim=0cm 0cm 0cm  0cm, clip, width=1.0\linewidth]{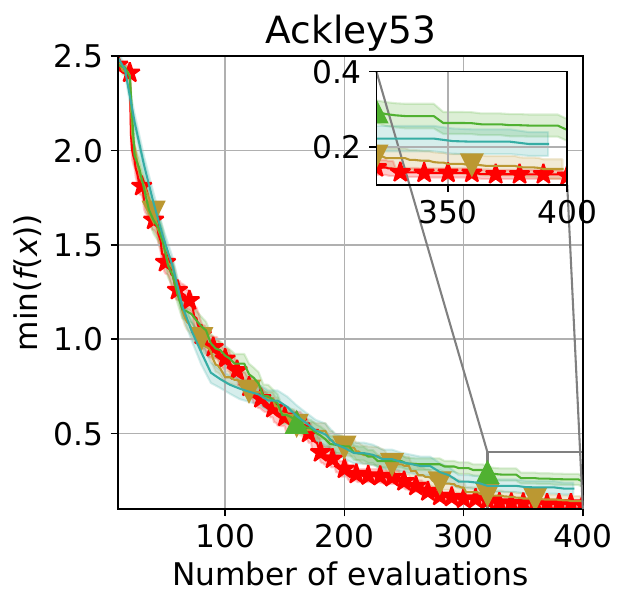}
    \end{subfigure}
    
    \begin{subfigure}{0.7\linewidth}
    \includegraphics[trim=0cm 0.5cm 0cm  5.5cm, clip, width=1.0\linewidth]{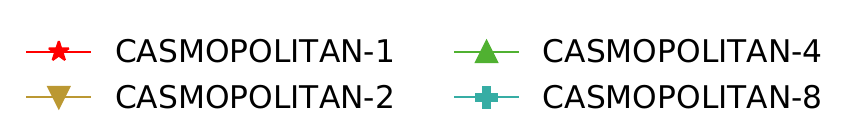}   
    \end{subfigure}
    \caption{Parallel \gls{CASMOPOLITAN} on representative categorical and mixed problems by \emph{number of function queries}.
    }    
    \label{fig:parallelbyquery}
\end{figure}

Supplementary to Fig. \ref{fig:parallel} which shows the comparison of performances of \gls{CASMOPOLITAN} of varying batch sizes by \emph{number of batches}, here we compare the performance by \emph{number of objective function queries} in Fig. \ref{fig:parallelbyquery}. It is evident that increasing the number of batches, at least in the experiments we consider, does not lead to significant performance deterioration even though we may achieve near-linear reduction in wall-clock time if we have sufficient parallel computing resources.

\subsection{Additional Results on the Black-Box Attack Task}
\label{subsec:additionalresults}

\begin{figure}[h]
    \centering
     \begin{subfigure}{1\linewidth}
    \includegraphics[trim=0cm 0cm 0cm  0cm, clip, width=1.0\linewidth]{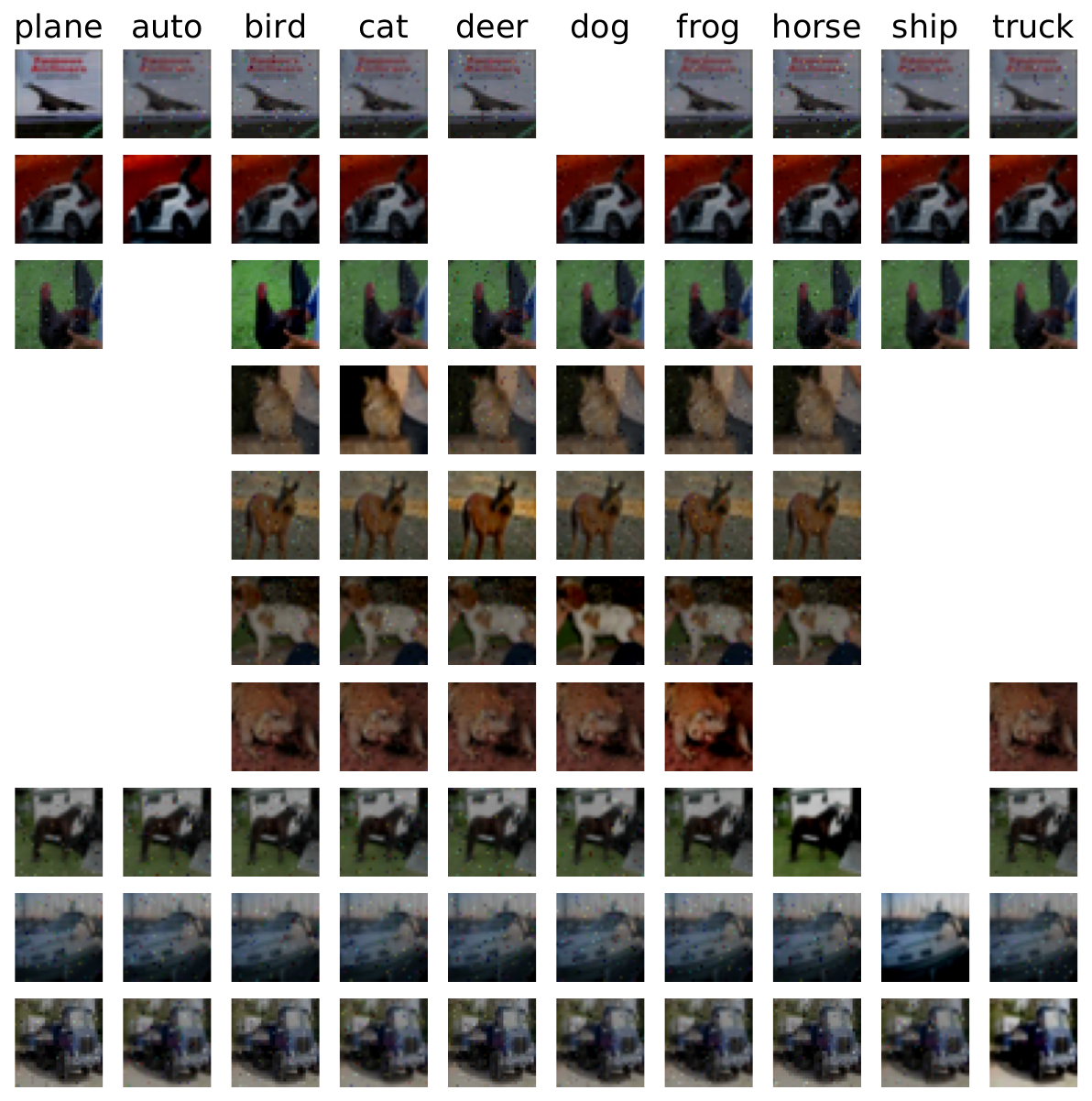}
    \end{subfigure}
        \caption{Some adversarial examples generated by our method.
    }  
    \label{fig:furteradvexamples}
\end{figure}

\begin{figure*}[t]
    \centering
     \begin{subfigure}{0.3\linewidth}
    \includegraphics[trim=0cm 0cm 0cm  0cm, clip, width=1.0\linewidth]{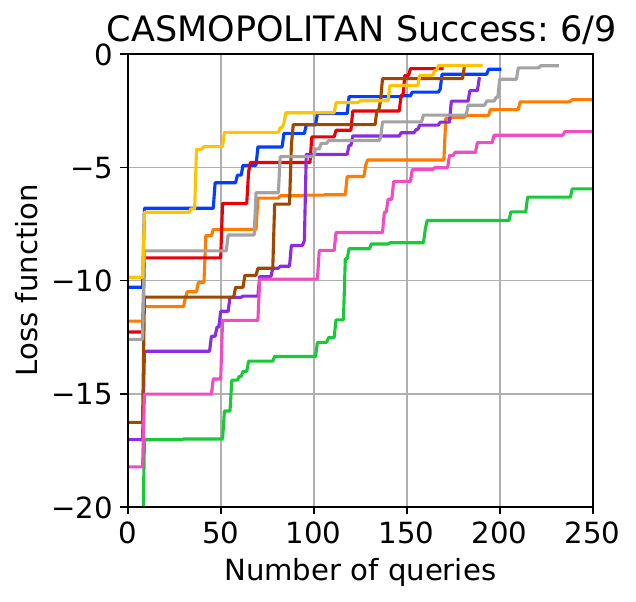}
   \caption{\gls{CASMOPOLITAN}}   
    \end{subfigure}
    \begin{subfigure}{0.3\linewidth}
        \includegraphics[trim=0cm 0cm 0cm  0cm, clip, width=1.0\linewidth]{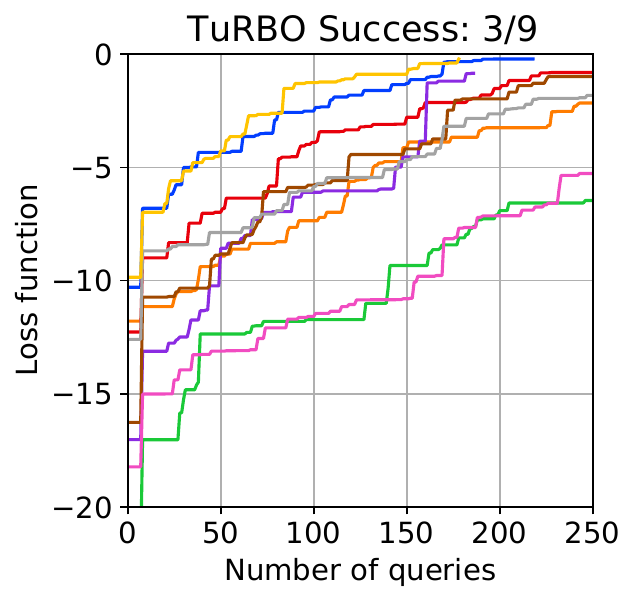}
     \caption{\gls{TuRBO}}
    \end{subfigure}
         \begin{subfigure}{0.3\linewidth}
    \includegraphics[trim=0cm 0cm 0cm  0cm, clip, width=1.0\linewidth]{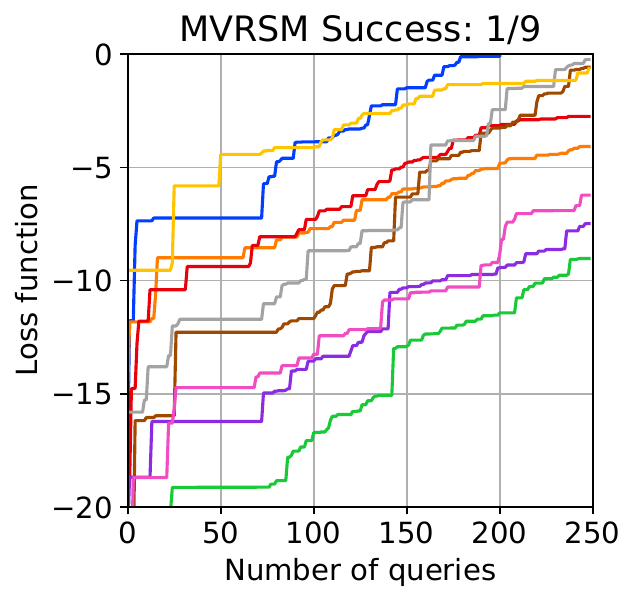}
      \caption{\gls{MVRSM}}   
    \end{subfigure}
    
    \caption{Attack loss function (described in Eq. (\ref{eq:attackloss}) against the number of queries of an attacked image in \gls{CASMOPOLITAN}, \gls{TuRBO} and \gls{MVRSM}, three most competitive methods of the black-box adversarial attack problem. Lines on each image denote the targeted attack to the 9 target classes. Some lines terminate earlier than the full budget because the attack is successful before using up all the query budget.}
    \label{fig:cwlossexamples}
\end{figure*}

Supplementary to the main text, in Fig. \ref{fig:furteradvexamples} we show more examples of the adversarial examples generated by our method, where the diagonal images are the original, unperturbed images in the \textsc{cifar-10} validation dataset that the \textsc{cnn} initially classifies correctly while the off-diagonal entries are the adversarial examples. From the $50$ images attacked by us, we select an image and we compare the objective function value against number of queries in the $9$ attack instances in Fig. \ref{fig:cwlossexamples}. It is clear that our method achieves higher success rate within the highly limited budget (successful in $6/9$ instances, as opposed to $3$ and $1$ in \gls{TuRBO} and \gls{MVRSM}), and even in cases where attack is unsuccessful within the budget, our method still increases the loss more and pushes it closer to the success boundary.

\subsection{Sensitivity Studies on the Additional Hyperparameters}
\label{subsec:sensitivity}
Similar to \gls{TuRBO}, our method introduces some additional hyperparameters related to the trust regions. In this section we examine the sensitivity of the performance of \gls{CASMOPOLITAN} towards these hyperparameters on Pest and Ackley53 problems. Specifically, we test the sensitivity towards:
\begin{itemize}
    \item Initial trust region length: unlike the hyperrectangular \gls{TR}s for the continuous space where $L^{x}_{\min}$ and $L^{x}_{\max}$ are additional hyperparemeters, the Hamming distance-based \gls{TR}s are constrained to be positive integers in $(0, d_h]$, relieving us from the need to tune $L^{h}_{\min}$ and $L^{h}_{\max}$. Nonetheless, the trust region length at the beginning of each restarts is still a free hyperparameter.
    \item Failure tolerance (\texttt{fail\_tol}): the number of successive failures to shrink the trust region. An aggressive \texttt{fail\_tol} setting (i.e. one that is very small) could lead to rapid trust region shrinking and possibly more frequent restarts. Note that it is generally rare to have a large number of consecutive successes in increasing the function value, and therefore we fix the success tolerance (\texttt{succ\_tol}) to be $2$.
    \item Shrinking rate of \gls{TR}s ($\alpha_s$): the multiplier when \gls{TR} shrinking is triggered $L \leftarrow \alpha_s L$; a more aggressive value of $\alpha_s$ leads to more rapid shrinking and restart upon stagnation in improving $f(\mathbf{z})$. Note that $\alpha_s$ is always coupled with the expansion rate $\alpha_e = \frac{1}{\alpha_s}$ when \gls{TR} expansion is triggered, and hence we do not further test the sensitivity to $\alpha_e$. Also, we use the same expansion and shrinking rates in both continuous and categorical \gls{TR}s.
\end{itemize}

\begin{figure*}[h]
    \centering
     \begin{subfigure}{0.24\linewidth}
    \includegraphics[trim=0cm 0cm 0cm  0cm, clip, width=1.0\linewidth]{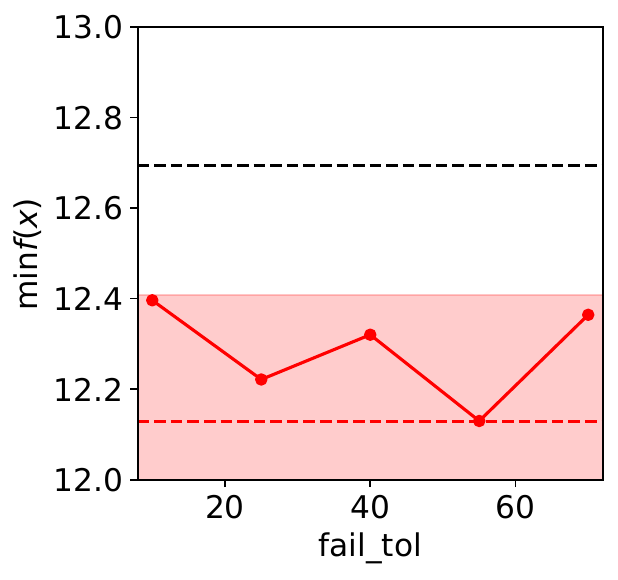}
    \caption{\texttt{fail\_tol} sensitivity. }
    \end{subfigure}
     \begin{subfigure}{0.24\linewidth}
    \includegraphics[trim=0cm 0cm 0cm  0cm, clip, width=1.0\linewidth]{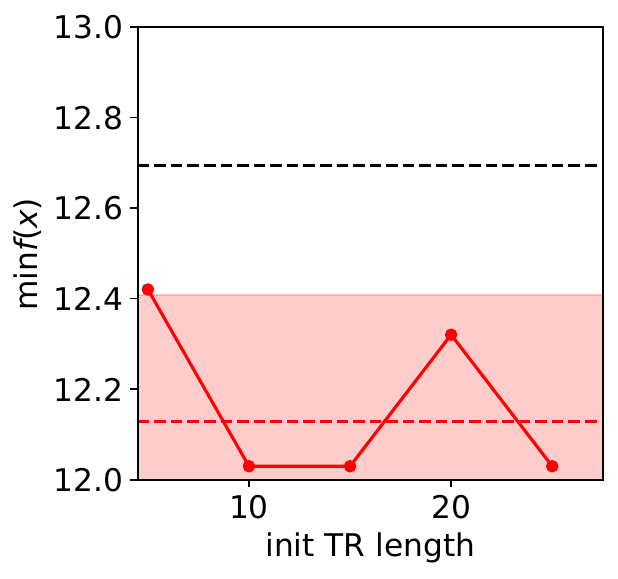}
    \caption{Initial \gls{TR} length sensitivity}
    \end{subfigure}
         \begin{subfigure}{0.24\linewidth}
    \includegraphics[trim=0cm 0cm 0cm  0cm, clip, width=1.0\linewidth]{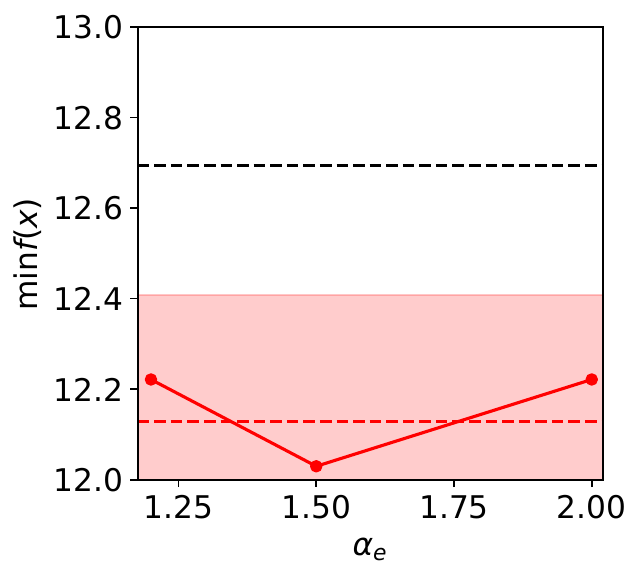}
    \caption{Expansion factor $\alpha_e$}
    \end{subfigure}
    
         \begin{subfigure}{0.24\linewidth}
    \includegraphics[trim=0cm 0cm 0cm  0cm, clip, width=1.0\linewidth]{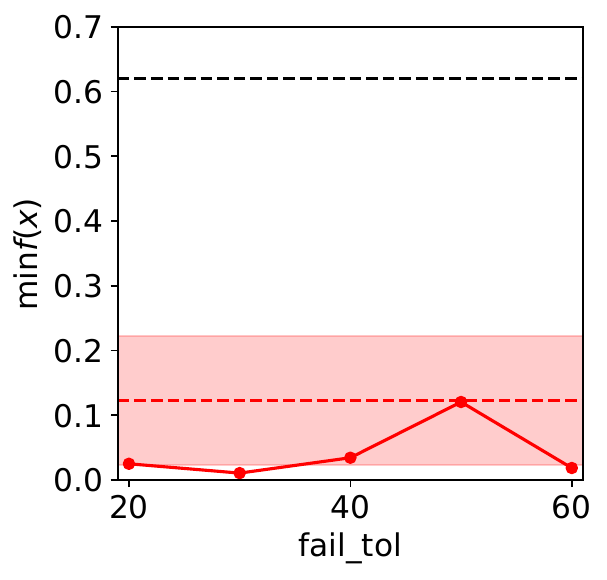}
    \caption{\texttt{fail\_tol} sensitivity. }
    \end{subfigure}
     \begin{subfigure}{0.24\linewidth}
    \includegraphics[trim=0cm 0cm 0cm  0cm, clip, width=1.0\linewidth]{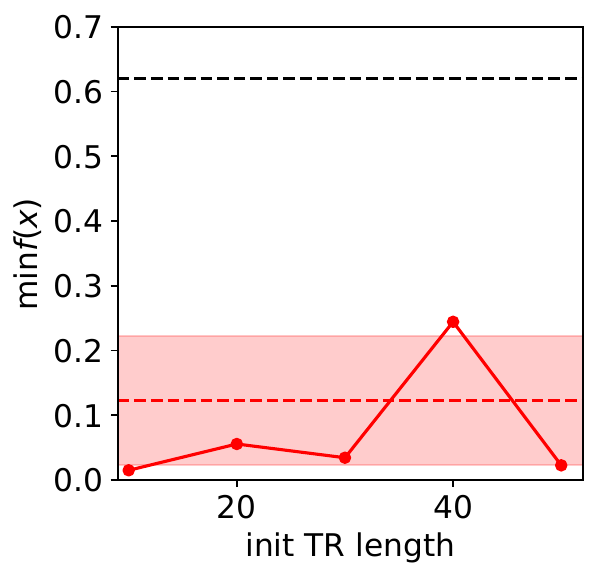}
    \caption{Initial \gls{TR} length sensitivity}
    \end{subfigure}
     \begin{subfigure}{0.24\linewidth}
    \includegraphics[trim=0cm 0cm 0cm  0cm, clip, width=1.0\linewidth]{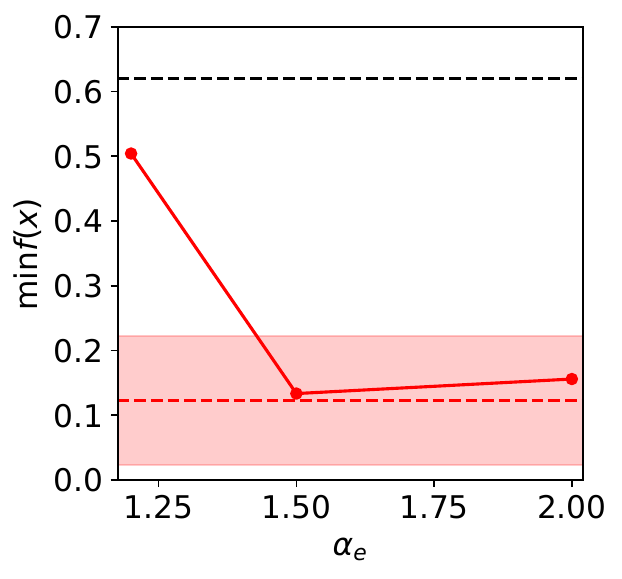}
    \caption{Expansion factor $\alpha_e$}
    \end{subfigure}
    
    \caption{Sensitivity of \gls{CASMOPOLITAN} performance towards various hyperparameters in Pest control (top row) and Ackley-53 (bottom row). The \textcolor{red}{red lines/shades} denote the mean $\pm$ 2 standard deviation of the baseline results in the main text, where as the dotted black line is the performance of the \emph{next best-performing baseline} (\gls{COMBO} and \gls{MVRSM} respectively).
    }    
    \label{fig:sensitivity}
\end{figure*}

We show the results in Fig. \ref{fig:sensitivity} where the default hyperparameter values are \texttt{fail\_tol} $=40$, initial trust region length $20$ (for Pest Control with $d_h = 25$) or $40$ (for Ackley-53 with $d_h = 50$) and $\alpha_s$ = $0.667$ (and thus $\alpha_e = 1.5$). In each of the experiments presented in Fig. \ref{fig:sensitivity}, we only tune the hyperparameter in interest, and leave all others at their default values. For the mixed problems, we do not tune the hyperparameters specific to the continuous \gls{TR}s (e.g. the initial, min and max continuous \gls{TR} lengths) and instead leave them at their default values in the official \gls{TuRBO} implementation. Furthermore, due to the large number of hyperparameter configurations, we only run each configuration once. The results show that the performance of \gls{CASMOPOLITAN} is generally insensitive to the hyperparameter choice, as the vast majority of the results fall within 2 standard deviations of results in the main text \emph{running the exactly the same configurations}, suggesting that, as a whole, the impact on performance due to different hyperparameter choices might not be more significant compared to the inherent randomness in initialisation in different trials. It is further worth noting that in all cases \gls{CASMOPOLITAN} still outperforms the corresponding next best baseline -- this suggests that the performance difference is mainly driven by the choice of \emph{different algorithms}, as opposed to \emph{different hyperparameters of the same algorithm}.

\subsection{Comparison against ALEBO and REMBO}
\begin{figure}[H]
    \centering
     \begin{subfigure}{0.54\linewidth}
    \includegraphics[trim=0cm 0cm 0cm  0cm, clip, width=1.0\linewidth]{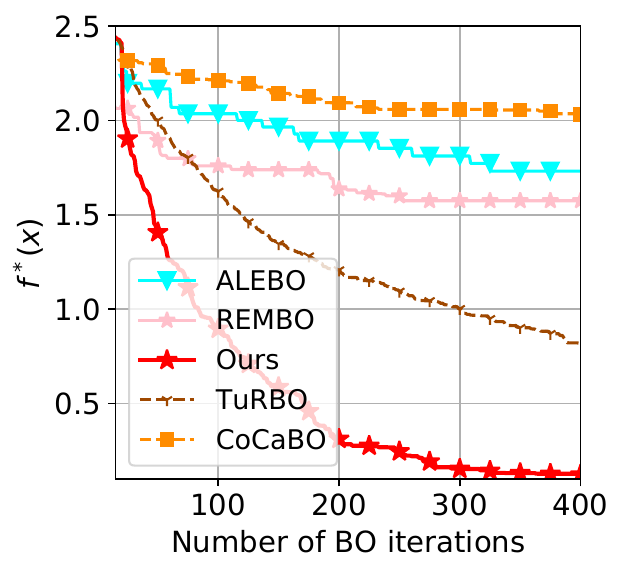}
    \end{subfigure}
        \caption{Comparison against \textsc{rembo} and \textsc{alebo} on the Ackley-53 problem. The lines denote the mean performance across 10 random trials.}  
    \label{fig:rembo_alebo}
\end{figure}

In this section we run a small comparison of our method against \textsc{rembo} \cite{Wang_2013Bayesian} and \textsc{alebo} \cite{letham2020re}, the representative methods of the class of high-dimensional \gls{BO} methods. We compare against them in the Ackley-53 problem with setups identical to the description in Sec. \ref{sec:experiments} in the main text, and we show the results in Fig. \ref{fig:rembo_alebo} where for both algorithms, we run under their respective default hyperparameter settings. We observe that while both outperform \textsc{cocabo}, they are outperformed by \gls{CASMOPOLITAN} and \textsc{turbo} by a large margin.

\subsection{Empirical Comparison of UCB-based and Random Restarts}

As discussed, the primary motivation of using \textsc{ucb}-based restarts of the trust regions is to theoretically driven, but in this section we investigate whether there exists any practical, finite-time benefits of using the \textsc{ucb}-based restarts. 

\begin{figure}[H]
    \centering
     \begin{subfigure}{0.54\linewidth}
    \includegraphics[trim=0cm 0cm 0cm  0cm, clip, width=1.0\linewidth]{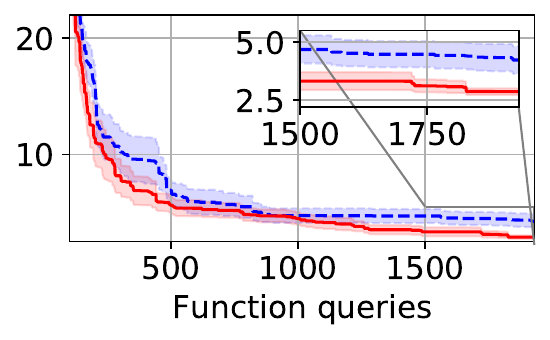}
    \end{subfigure}
        \caption{\textcolor{red}{\textsc{ucb}} vs \textcolor{blue}{random} restarts in 20d Levy over 10 trials. t-test p-value between the two method is 0.048 at the final iteration. Shades denote $\pm 1$ standard error.}  
    \label{fig:rand_vs_ucb}
\end{figure}

Practically, optimising the \textsc{ucb} on the auxiliary GP exactly is difficult. Instead, at each restart of the trust region, we simply sample a large number of points, compute their \textsc{ucb} scores based on the auxiliary \gls{GP}, and select the top ones as the initialising points for the next restart. We emphasise that the auxiliary \gls{GP} is not meant to fit well to the objective function, as otherwise we do not need trust regions to constrain the surrogate, but is to instead generate better initialising points than random selection. Based on this described procedure, we conduct an experiment comparing UCB vs random restarts on 20-dimensional Levy function, and we show the results in Fig. \ref{fig:rand_vs_ucb}. It can be seen that using UCB-based restarts leads to small but statistically significant improvements over the randomly initialising baseline. Furthermore, in terms of running time, since the auxiliary GPs scale with the number of \emph{restarts} instead of number of observations, we find the UCB variant to be only $1.9\%$ slower in terms of running time. With these results, we expect that the proposed UCB criterion to lead to practical benefits even in modestly higher-dimensional problems given an extended query budget (such that we would typically observe a number of \gls{TR} restarts for the effect of initialisation at the start of each restart to be significant).

\section{Implementation Details}
\label{appendix:details}

\begin{table*}[t]
    \centering
        \caption{Configurations of the test problems.}
        \begin{footnotesize}
    \begin{tabular}{p{0.15\linewidth} p{0.15\linewidth} p{0.5\linewidth}}
    \toprule
    Objective $f$ & Type & Inputs\\
    \midrule
    Contamination \cite{hu2010contamination} & real, cat, min, 25-dim & Choices on whether to use control at each stage $\mathbf{h} \in \{\text{True}, \text{False}\}^{25}$\\
    \midrule
    Pest \cite{oh2019combinatorial} & real, cat, min, 25-dim & Pesticide choice at each stage (or use no pesticide) $\mathbf{h} \in \{\text{No pesticide}, 1, 2, 3, 4\}^{25}$\\
    \midrule
        DifficultPest & real, cat, min, 80-dim & Pesticide choice at each stage (or use no pesticide) $\mathbf{h} \in \{\text{No pesticide}, 1, 2, 3, 4\}^{80}$\\
    \midrule
    \textsc{maxsat} & real, cat, min, 60-dim & $\mathbf{h} \in \{0, 1\}^{60}$\\
    \midrule
    Func2C & synthetic, mixed, max, 4-dim & $\mathbf{x} \in [-1, 1]^2$ \\
    \cite{cocabo}& & $h_1 = \{\text{ros}(\bx), \text{cam}(\bx), \text{bea}(\bx)\}$\\
    & & $h_2 = \{+\text{ros}(\bx), +\text{cam}(\bx), +\text{bea}(\bx), +\text{bea}(\bx), +\text{bea}(\bx)\}$\\
    \midrule
    Func3C& synthetic, mixed, max, 5-dim & $\mathbf{x} \in [-1, 1]^2$ \\
     \cite{cocabo} & & $h_1 = \{\text{ros}(\bx), \text{cam}(\bx), \text{bea}(\bx)\}$\\
    & & $h_2 = \{+\text{ros}(\bx), +\text{cam}(\bx), +\text{bea}(\bx), +\text{bea}(\bx), +\text{bea}(\bx)\}$\\
    & & $h_3 = \{+5\times\text{ros}(\bx), +2\times\text{cam}(\bx), +2\times\text{bea}(\bx), +3\times\text{bea}(\bx)\}$\\
    \midrule
    \textsc{xg-mnist} & real, mixed, max, 8-dim & $h_1$ (booster type type) $\in$ \{gbtree, dart\} \\
    & & $h_2$ (grow policies) $\in$ \{depthwise, loss\} \\
    & & $h_3$ (training objective) $\in$ \{softmax, softprob\} \\
    & & $x_1$ (learning rate) $\in [0, 1]$ \\
    & & $x_2$ (max depth) $\in [1, 10]$ \\
    & & $x_3$ (minimum split loss) $\in [0, 10$ \\
    & & $x_4$ (subsample) $\in [0.001, 1]$ \\ 
    & & $x_5$ (amount of regularisation) $\in [0, 5]$ \\ 
    \midrule
    Ackley-53 & synthetic, mixed, min, 53-dim & $\mathbf{h} \in \{0, 1\}^{50}$ \\
    \cite{bliek2020black} & & $\mathbf{x} \in [-1, 1]^3$ \\
    \midrule
    Rosen-200 & synthetic, mixed, min, 200-dim & $\mathbf{h} \in \{0, 1\}^{100}$ \\
     & & $\mathbf{x} \in [-2, 2]^{100}$ \\
    \midrule
    Black-box adversarial attack & real, mixed, max, 85-dim & Choice on the location of the pixel $h_i \in \{0, 1, ..., 13\} \text{ } \forall i \in [1, 42], i \in \mathbb{Z}$ \\
    & & Upsampling technique $h_{43} \in $ \{bilinear, nearest, bicubic\} \\
    & & Amount of perturbation $\mathbf{x} \in [-1, 1]^{42}$ \\
    
\bottomrule 
    \multicolumn{3}{l}{Note: real/synthetic: whether the problem is/simulates a real-life task or whether it is a standard benchmark function. } \\
    \multicolumn{3}{l}{cat/mixed: categorical or mixed categorical-continuous problem.} \\
    \multicolumn{3}{l}{max/min: maximisation or minimisation problem. We flip the sign of the objective function values where appropriate.} \\
    \end{tabular}
    \label{tab:config}
    \end{footnotesize}
\end{table*}

\subsection{Description of the categorical problems}
\label{subsec:descp_cat}
A table containing the details and other characteristic details of all the test problems are shown in Table \ref{tab:config}.

\paragraph{Contamination Control} Contamination Control is a binary optimisation problem in food supply chain \cite{hu2010contamination}: at each stage, we have the choice of whether to introduce contamination control, but early use of contamination control could inevitably lead to increase in cost and as such our objective is to minimise food contamination with the smallest monetary cost (hence a minimisation problem). It is worth noting that in this problem and the Pest Control problem described below, the actions taken by the previous stage have implications on the following stages, thus leading to highly complicated interactions amongst the different variables. In this problem, we use the implementation used in \citet{oh2019combinatorial}. However, it is worth noting that while \citet{oh2019combinatorial} consider a 21-stage (with a total of $2^{21} \approx 2.1 \times 10^{6}$ configurations) problem, we increase the total number of stages to $25$ (with a total of $2^{25} \approx 3.4 \times 10^{7}$ configurations). In this experiment we limit the maximum number of evaluations of $150$, as the running time of \textsc{bocs} quickly increases beyond our computing budget if we set the it to a significantly higher value.
    
\paragraph{Pest Control} We use the problem proposed in \citet{oh2019combinatorial} which expands the contamination control problem into a multi-categorical optimisation problem: at each stage, we now need not only to determine whether to take an action (to use pesticide or not), but also the type of the pesticide ($4$ choices in total). This thus gives rise to $5$ potential choices for each stage. Similar to Contamination control, we again increase the total number of stages to $25$ (as opposed to $21$ in \citet{oh2019combinatorial}) to give an expanded and more complicated search space. In Ablation Studies of Sec. \ref{sec:experiments}, we also include a variant named DifficultPest, where the total number of stages is further increased to $80$.
    
\paragraph{Weighted Maximum Satisfiability}
Maximum satisfiability problem is a classical combinatorial optimisation problem that aims to determine the maximum number of clauses of a given Boolean formula in conjunctive normal form (\textsc{cnf}) that can be made true by an assignment of truth values to the variables. Similar to \citet{oh2019combinatorial}, we take the same $60$-variable benchmark from Maximum Satisfiability Competition 2018\footnote{\url{http://sat2018.azurewebsites.net/competitions/}} (\texttt{frb-frb10-6-4.wcnf } problem from \url{https://maxsat-evaluations.github.io/2018/benchmarks.html})

\subsection{Description of the mixed problems}
\label{subsec:descp_mixed}
\paragraph{Func2C and Func3C} These synthetic problems were first proposed in \citet{cocabo}. In Func2C ($d_x = 2, d_h = 2$), the value of $\mathbf{h}$ determines the objective function value that is a linear combination of three benchmark functions, namely Beale, Six-Hump Camel and Rosenbrook (abbreviated as bea, cam and ros in Table \ref{tab:config}); the function form of these 3 functions are:
\begin{align*}
	\text{bea}(\mathbf{x}) &= (1.5 - x_1 + x_1x_2)^2 + (2.25 - x_1 + x_1x_2^2)^2 + \\
	& (2.625-x_1+x_1x_2^3)^2.
\end{align*}
\begin{align*}
	\text{cam}(\mathbf{x}) &= (5 - 2.1x_1^2 + \frac{x_1^4}{3})x_1^2 + x_1x_2 + (-4 + 4x_2^2)x_2^2.
\end{align*}
\begin{align}
	\text{ros}(\mathbf{x}) &= (1 - x_1)^2 + 100(x_2 - x_1^2)^2.
\end{align}

Func3C ($d_x=2, d_h=3$) is similar but has one extra categorical dimension to enable more complicated interactions.

\paragraph{\textsc{xg-mnist}} ($d_x = 5, d_h=3$) This is a real hyperparameter tuning task of a machine learning model (XGBoost). The tunable continuous hyperparameters are maximum depth, minimum split loss, subsample, learning rate of the optimiser and the amount of regularisation. The categorical variables are the booster type, grow policies and training objective. We use the \texttt{xgboost} python package and adopt a train-test split of $7:3$ on the \textsc{mnist} data. Note that this setup is identical to that used in \citet{cocabo}.

\paragraph{Ackley-53} is a stylised version of the original 53-dimensional Ackley function, whose original form is given by:
\begin{align}
    f(\mathbf{z}) &= -a \exp\Big(-b\sqrt{\frac{1}{d}\sum_{i=1}^d z_i^2)}\Big) - \nonumber \\ 
    & \exp\Big(\frac{1}{d}\sum_{i=1}^d \cos(cz_i)\Big) + a + \exp({1}),
\end{align}
where in this case $a = 20, b = 0.2, c = 2\pi$ and $d = 53$ and we define $\mathbf{z} \in [-1, 1]^{53}$. From this continuous form, the first 50 dimensions are modified to be \textit{binary} variables that take the value of either $0$ and $1$, and the final $3$ variables are continuous and limited in the range of $[-1, 1]^3$. This adaptation is first proposed in \citet{bliek2020black}. This function has a known global minimiser of $\mathbf{h}^* = [0, ..., 0]$ and $\mathbf{x}^* = [0, 0, 0]$ with $f^*(\mathbf{z}) = 0$.

\paragraph{Rosenbrock-200} is a stylised and scaled version of the classical Rosenbrock function. The Rosenbrock function is given by:
\begin{equation}
    f(\mathbf{z}) = \frac{1}{50000} \Big( \sum_{i=1}^{d-1} \big( 100(z_{i+1} - z_i^2)^2 + (z_i - 1)^2 \big) \Big),
\end{equation}
where in this case $d=200$. The first 100 dimensions are then converted to binary variables, while the final 100 dimensions are continuous in the range of $[-2, 2]^{100}$.

\paragraph{Black-box adversarial attack} We adapt black-box setup from \citet{ru2019bayesopt}, one of the first works that introduce \gls{BO} in the image adversarial attack setting. Specifically, denoting $\mathcal{M}$ as the target model (or the victim model, in this case a \textsc{cnn} image classifier) from which we may query an image input $I$, the \gls{BO} agent can only observe the prediction scores on all $C$ classes (for \textsc{cifar-10}, $C=10$): $\mathcal{M}(I): \mathbb{R}^d_+ \rightarrow [0, 1]^C$ (thus a ``black-box'', since gradients, architecture and other information of the classifier itself are never revealed to the attack agent). Therefore, denoting $I$ as the original, unperturbed image that $\mathcal{M}$ correctly gives its prediction as $c$, the targeted adversarial attack objective is to find some perturbation $\boldsymbol{\delta} \in \mathbb{R}^d$ to be superposed on the original image such that $\mathcal{M}$ now mis-classify the perturbed image to another target class $t$. In this work, we use the identical \textsc{cnn} models to the previous works \cite{ru2019bayesopt,tu2019autozoom, alzantot2019genattack}, which approximately gives 80\% validation accuracy on the \textsc{cifar-10} dataset. \citet{ru2019bayesopt} further claim that the query efficiency of the BayesOpt attack strategy can be enhanced by searching the perturbation over a latent space $\tilde{\boldsymbol{\delta}} \in \mathbb{R}^{d_r}$ with reduced dimension $d_r \ll d$ and upsampling it back to the original high-resolution image space $\mathbb{R}^{d}$. This leads to a categorical variable which is the downsampling/upsampling technique, and in this work we have 3 options: bilinear, nearest and bicubic interpolations. In our attack on CIFAR10 images, we set $d=32 \times 32 \times 3$ and $d_r=14 \times 14 \times 3$ following \citet{ru2019bayesopt}. 

In our work, we adopt a \emph{sparse} setup where instead of perturbing all the pixels, we only perturb one pixel per row per colour in the latent space, allowing a total of $s=14 \times 3 = 42$ pixels in the reduced space to take non-zero values. Such setup corresponds to add perturbation to some pixels of the original image only, which is more actionable in real life (for e.g., to evade real-life image classifiers this only requires one to carefully manipulate some parts of a printed image; this is contrasted to $L_2$ attack, another often studied setup where we perturb a small amount on \emph{every} pixel of the image which is less feasible in real life). We additionally impose a constraint on the pixels $\epsilon$ to limit the maximum amount of perturbations. Mathematically, the goal is formulated as:
\begin{equation*}
    \arg \max_{j \in \{1, ..., C\}} \mathcal{M}\left(I + \text{Upsample}(\tilde{\boldsymbol{\delta}} ) \right)_j = t.
\end{equation*}
\begin{equation}
    \text{s.t. } ||\{\tilde{\delta}_i \mid \tilde{\delta}_i \neq 0\}|| \leq s \text{ and } ||\tilde{\boldsymbol{\delta}}||_{\infty} \leq \epsilon,
\end{equation}
where the first $||\cdot||$ denote the cardinality of the set of non-zero elements of $\tilde{\boldsymbol{\delta}}$ and the second $||\cdot||$ is the $L_{\infty}$ norm.

In summary, the variables $\mathbf{z} \in \mathbb{R}^{85}$ that we need to search over include $42$ categorical variables deciding the positions of the pixels to be perturbed at each row (thus $14$ choices for each variable), $1$ categorical variable on the type of upsampling technique chosen ($3$ choices) and $42$ continuous variables defining the amount of perturbation to be added to each chosen pixel. These setups conveniently cast the problem of finding adversarial perturbation as a mixed continuous-categorical optimisation problem for which \gls{CASMOPOLITAN} is suitable. In this case, we follow \citet{ru2019bayesopt} and select the following as the objective function $f$ we aim to maximise:
\begin{equation}
    f(\mathbf{z}) = \Big [\log \mathcal{M}\big(I + \boldsymbol{\delta}(\mathbf{z})\big)_t - \log \mathcal{M}\big(I + \boldsymbol{\delta}(\mathbf{z})\big)_c \Big],
    \label{eq:attackloss}
\end{equation}
where $\boldsymbol{\delta}(\mathbf{z})$ is the image perturbation $\boldsymbol{\delta}$ induced by our combined choices of the pixel locations and the corresponding amount of perturbations. Essentially, in Eq. (\ref{eq:attackloss}), we aim to maximise the difference between the logit value of the target class $t$ and the true class $c$, and trivially the attack succeeds if and when $f(\mathbf{z}) > 0$. Thus, we terminate each experiment either the attack succeeds or the maximum budget ($250$) is reached. It is finally worth noting that the combined dimension $\mathbf{z}$ is $85$-dimensional whose one-hot transformed dimension amounts to $633$, which is clearly beyond the common scope of usage of vanilla \gls{GP}-\gls{BO} that neither gives special treatments to the categorical dimensions nor is tailored for high-dimensional optimisation.

\subsection{Experimental setup}
\label{subsec:setup}
We run all experiments on a shared Intel Xeon server with 256GB of \textsc{ram}. For all categorical problems, we run $20$ random trials with the exception of \textsc{bocs} on Contamination Control, where we only run $5$ trials due to the very long running time of \textsc{bocs} and our computing constraints (reported in App. \ref{subsec:runtime}). For the mixed problems, we follow \citet{cocabo}, where we run $20$ trials for the synthetic problems and $10$ trials for the real-life problems. For black-box attack, we run attack once on all $450$ attack instances on $50$ images. We report mean and standard error in all cases.

\paragraph{\gls{CASMOPOLITAN}} Our algorithm introduces a number of additional hyperparameters relating to the initialisation, adjustment and restarting of the trust regions. In the categorical space, the distances (Hamming distance) are always integers, and the minimum ($L^h_{\min}$) and maximum ($L^h_{\max}$) trust region sizes are always set to $0$ and the dimensionality of the problem (i.e. the diameter of the combinatorial graph), respectively. The failure tolerance, which is the number of successive failures in increasing the best objective function value before shrinking the trust region size (\texttt{fail\_tol}), is set to $40$ unless otherwise specified; the success tolerance (\texttt{succ\_tol}), which is the number of successive successes in increasing the best function value before expanding the trust region, is set to $2$. We investigate the sensitivity in performance of our algorithm to these hyperparameters in App. B3. The only other \gls{GP} hyperparameter is the amount of noise (or jitter), where we constrain the noise variance in the interval of $[10^{-5}, 0.1]$ and this value is learnt as a hyperparameter during the log-marginal likelihood optimisation. We always start the experiments with $20$ initial randomly sampled points.

In the mixed setting that involves continuous variables, unless specified otherwise we always use the Matern $5/2$ kernel. In these continuous problems, we bound the lengthscale in the range of $[0.01, 0.5]$ and outputscale in $[0.5, 5]$ and in all cases, we normalise the continuous inputs $\mathbf{x}$ into hypercubes $[0, 1]^{d_x}$ and standardise the targets by their mean and standard deviation from the initially randomly sampled locations $\mathbf{y}$: $\tilde{\mathbf{y}} = \frac{y - \bar{y}}{\sigma{(y)}}$. When we compute the mixed kernel in Eq. (\ref{eq:mixkernel}), we set $\lambda=0.5$ as it is empirically shown to perform the best in \citet{cocabo} that initially propose this kernel. On the hyperparameters specific to the continuous trust regions, since they are identical to those introduced in \gls{TuRBO} \cite{eriksson2019scalable}, we do not change these settings from their default values ($L^x_{\min} = 0.5^7, L^x_{\max} = 1.6, L^x_{0} = 0.8$) with the exceptions of $\alpha_s$ (and hence $\alpha_e$), \texttt{succ\_tol} and \texttt{fail\_tol} which all follow the settings of the categorical trust regions described above, instead of being independent hyperparameters. For the trade-off parameter $\beta_i$ at each restart, we follow the common practice of setting $\beta_i$ to a constant value \cite{Berkenkamp2019}. In our case, we set $\sqrt{\beta_i}  = 1.96$ as it has been shown this value of $\beta_i$ performs well for a variety of \gls{BO} tasks \cite{Berkenkamp2019}.



During optimisation of the acquisition function, we use the local search strategy (for categorical optimisation; in Sec. \ref{subsec:combinatorialmethod}) or the interleaved strategy (for mixed optimisation; in Sec. \ref{subsec:mixedmethod}). In all cases, we initialise the search at the best location found so far, and we set the maximum number of local/interleaved search to be $100$; for interleaved search in mixed space, one local search move + one gradient-based optimisation step count as one interleaved search step, we use Adam \cite{kingma2014adam} as the default optimiser for the log-marginal likelihood with learning rate $0.1$ and maximum step $100$, although we find the performance to be relatively stable at least for maximum step in the range of $[100, 300]$ and learning rate in the range of $[0.03, 0.3]$. By default, we optimise the log-marginal likelihood $3$ times and select the point(s) with the largest acquisition function across the 3 runs, although we do not find optimising with just $1$ restart to be significantly worse. In this work we use expected improvement (\textsc{ei}) as the acquisition function, although our work is compatible with any other common choice such as \textsc{gp-ucb} or Thompson sampling. When \gls{CASMOPOLITAN} is run in the batch setting, we use the Kriging Believer strategy \cite{ginsbourger2010kriging} to select $b$ points simultaneously: specifically, given observation data $D_t = \{\mathbf{z}_i, y_i\}_{i=0}^t$ and a \gls{GP} model, we first optimise the acquisition function as usual to propose the first out of the $b$ points required:
\begin{equation}
    \mathbf{z}_t^{(1)} = \arg \max \alpha(\mathbf{z}\mid D_t)
\end{equation}
We then fully trust $\mu(\mathbf{z}_t^{(1)})$, the predictive mean at $\mathbf{z}_t^{(1)}$, as a perfect proxy of the true objective function value $f(\mathbf{z}_t^{(1)})$, and use this ``hallucinated'' input-output tuple $(\mathbf{z}_t^{(1)}, \mu(\mathbf{z}_t^{(1)}))$ to update the \gls{GP} $D_{t-1} \leftarrow D_{t-1} \cup (\mathbf{z}_t^{(1)}, \mu(\mathbf({z}_t^{(1)}))$. Conditioned on this \gls{GP} with ``hallucinated'' inputs and observations, we then optimise the acquisition function value again to propose the second point $ \mathbf{z}_t^{(2)}$ and this process is repeated until all $b$ proposals are selected.

\paragraph{Other baselines} Where possible and open-sourced, for the other baselines we use the implementation from their respective original authors:
\begin{itemize}[leftmargin=0.1in]
 \itemsep0.1em
    \item \textbf{\gls{TuRBO}} We use the official implementations made available by \citet{eriksson2019scalable} at \url{https://github.com/uber-research/TuRBO}. For the implementation of \gls{TuRBO} in categorical space or where categorical variables are involved, we use the modified implementation supplied by the organisers of the 2020 NeurIPS Black-box Optimisation Challenge which includes \gls{TuRBO} as a baseline that additionally supports one-hot transformation on the categorical variables (\url{https://github.com/rdturnermtl/bbo_challenge_starter_kit}). Note that for the vanilla one-hot \gls{GP}-\gls{BO}, we also adapt from the \gls{TuRBO} implementation but remove the additional features such as trust regions and restarting.
    \item \textbf{\gls{CoCaBO}} We use the official implementation by \citet{cocabo} at \url{https://github.com/rubinxin/CoCaBO_code}. Note that in original \gls{CoCaBO}, there is an option for the value of $\lambda$ in Eq. (\ref{eq:mixkernel}) to be optimised as a hyperparameter within bounds of $[0, 1]$; in our work for fairness of comparison, we fix $\lambda$ to $0.5$ since it is the value used in our method. It is worth noting that $\lambda = 0.5$ is also shown to be performing overall the best in \gls{CoCaBO} from the results reported in \citet{cocabo}.
    \item \textbf{\gls{MVRSM}} We use the official implementation by \citet{bliek2020black} at \url{https://github.com/lbliek/MVRSM}.
    \item \textbf{\gls{COMBO}} We use the official implementation by \citet{oh2019combinatorial} at \url{https://github.com/QUVA-Lab/COMBO}.
    \item \textbf{\textsc{bocs}} We use the official implementation by \citet{baptista2018bayesian} at \url{https://github.com/baptistar/BOCS}.
    \item \textbf{\textsc{tpe}} \textsc{tpe} (Tree Parzans Estimator) is implemented by the Hyperopt python package, available at \url{http://hyperopt.github.io/hyperopt/}.
    \item \textbf{\textsc{smac}} We use the implementation at \url{https://github.com/automl/SMAC3}.
\end{itemize}

\section{Proofs and Further Theoretical Analysis}
\label{appendix:proofs}

\subsection{Lemma \ref{lemma:kernel}}
\label{sec:proof-psd}

\begin{lemma} \label{lemma:kernel}
The proposed categorical kernel in Eq. (\ref{eq:catkernel}) and mixed kernel in Eq. (\ref{eq:mixkernel}) are valid kernels (i.e. positive semi-definite kernels).
\end{lemma}

\textbf{Proof.}
For the categorical kernel in Eq. (\ref{eq:catkernel}), we have that exponential of a kernel is also a kernel, and since the categorical overlap kernel is a valid kernel \cite{cocabo}, its exponentiated version is also a valid kernel. For the mixed kernel in Eq. (\ref{eq:mixkernel}), since addition and multiplication between kernels result in valid kernels, and since both $k_x(.,.)$ and $k_h(.,.)$ are valid kernels, therefore, the mixed kernel in Eq. (\ref{eq:mixkernel}) is also a valid kernel. $\quad \square$

\subsection{Proof of Theorem \ref{thr:kernelmig}} \label{sec:proof-kernelmig}
In this section, we derive the maximum information gain of the categorical kernel $k_h$ (Section \ref{sec:proof-kernelcatmig}) and the mixed kernel $k$ (Section \ref{sec:proof-kernelmixmig}).

\subsubsection{Maximum information gain of the categorical kernel}
\label{sec:proof-kernelcatmig}
We derive the maximum information gain of the categorical kernel $k_h$ proposed in Eq. (\ref{eq:catkernel}) by bounding $\gamma(T; k_h; \mathcal{H})$ directly. Let us first consider the case when the objective function $f$ has only one categorical variable $h$ with $n$ distinct values (i.e. $h \in \lbrace A_1, A_2, \dots, A_n \rbrace$ where $A_i$ is a categorical value and $A_i \neq A_j$ when $i \neq j$). Let us consider $T$ data points $h_1, h_2, \dots, h_T$, then its corresponding covariance matrix $K_T$ is $\lbrack k_h(h_i, h_j) \rbrack_{i,j=1}^{T}$. As the maximum information gain $\gamma(T; k_h; \mathcal{H})$ is equal to $\log \vert I_T + \sigma^{-2}K_T \vert$ where $I_T$ is the identity matrix of size $T$,\footnote{$\vert S \vert$ denotes the determinant of matrix $S$.} thus, we will bound $\gamma(T; k_h; \mathcal{H})$ by bounding $\log \vert I_T + \sigma^{-2}K_T \vert$. Our general idea is to perform a decomposition of $K_T$, i.e. expressing $K_T = \Phi E \Psi^T$ where $\Phi \in \mathbb{R}^{T \times n}$, $\Psi \in \mathbb{R}^{T \times n}$, and $E \in \mathbb{R}^{n \times n}$, and then apply the Sylvester's determinant theory and the Hadamard's inequality to derive an upper bound for $\log \vert I_T + \sigma^{-2}K_T \vert$.

In the sequel, for ease of notation, we define the function $q$ as a mapping from $A_i$ to $i$. In particular, $q(A_i)=i, \ \forall i=1,\dots,n$. With the categorical kernel $k_h(h,h') = \exp{(l \delta(h, h')})$, in the following, we will prove that $K_T$ can be decomposed as,\footnote{When $T=n$, this decomposition is  equivalent to the eigendecomposition. That is, the diagonal of matrix $E$ consists of the eigenvalues of $K_T$ and each column of $\Phi$ is an eigenvector of $K_T$.}
\begin{equation} \label{eq:kernel-eig}
\begin{aligned}
    \quad \quad \ K_T = \Phi E \Psi^T,
\end{aligned}
\end{equation}
where $\Phi, \Psi \in \mathbb{R}^{T \times n}, E \in \mathbb{R}^{n \times n}$, and
\begin{equation} \nonumber
\begin{aligned}
\Phi &= \begin{bmatrix} \phi(h_1) \\ \phi(h_2) \\\dots \\ \phi(h_T) \end{bmatrix},
\Psi =  \begin{bmatrix} \psi(h_1) \\ \psi(h_2) \\ \dots \\ \psi(h_T) \end{bmatrix}, \\
E &= \diag{(\exp{(l)} + n - 1, \exp{(l)}-1, ..., \exp{(l)}-1)},
\end{aligned}
\end{equation}
with $\phi(h_i)$ being an $n$-dimensional row vector with $1$ at the 1st column, $(-1)$ at the $q(h_i)$-th column, and $1$ at the $(q(h_i)+1)$-th column, i.e.,
\begin{equation} \nonumber
\phi(h_i) = \begin{cases}
 \lbrack 1 \ 1 \ 0 \ ... \ 0 \ 0 \rbrack, & \text{if } q(h_i)=1\\
 \lbrack 1 \ 0 \ 0 \ ... \ 0 \ (-1) \ 1 \ 0 \dots 0 \rbrack, & \text{if } 1<q(h_i)<n \\
 \lbrack 1 \ 0 \ 0 \ ... \ 0  \ (-1) \rbrack, & \text{if } q(h_i)=n
\end{cases},
\end{equation}
and $\psi(h_i)$ being an $n$-dimensional row vector with the following formula,
\begin{equation} \nonumber
\psi(h_i) = \begin{cases}
 \dfrac{1}{n} \lbrack 1 \ (n-1) \ (n-2) \ ... \ 1 \rbrack, \quad \text{if } q(h_i)=1\\
 \dfrac{1}{n} \lbrack 1 \ (-1) \ ... \ -(q(h_i)-1) \ (n-q(h_i)) \ ... \ 1 \rbrack, \\ \quad \quad \quad \quad \quad \quad \quad \quad \quad \quad \quad \quad \text{if } 1<q(h_i)<n \\
 \dfrac{1}{n} \lbrack 1 \ (-1) \ ... \ -(n-1) \rbrack, \ \quad \ \ \text{if } q(h_i)=n.
\end{cases}
\end{equation}
To prove the decomposition in Eq. (\ref{eq:kernel-eig}), we compute the element at the $i$-th row and $j$-th column of $\Phi E \Psi^T$, i.e. $[\Phi E \Psi^T]_{ij}$, and then prove that $[\Phi E \Psi^T]_{ij}$ is equal to $[K_T]_{ij}$. To compute $[\Phi E \Psi^T]_{ij}$, it can be directly seen that,
\begin{equation} \nonumber
    [\Phi E \Psi^T]_{ij} = \sum_{r=1}^n \phi_r(h_i) E_r \psi_r(h_j),
\end{equation}
where $\phi_r(h_i)$ denotes the $r$-th element of $\phi(h_i)$, $\psi_r(h_j)$ denotes the $r$-th element of $\psi(h_j)$ and $E_r$ denotes the $r$-th element on the diagonal of matrix $E$. We then consider the following three cases:

\textit{Case 1}: $q(h_j) = q(h_i)$. First, let us consider $1<q(h_i)<n$, then we have,
\begin{equation} \nonumber
\begin{aligned}
    [\Phi E \Psi^T]_{ij} = & 1 \times (\exp(l)+n-1)\times \dfrac{1}{n} \\
                           & + (-1) \times (\exp(l)-1) \times \dfrac{(-q(h_i)+1)}{n} \\
                           & + 1 \times (\exp(l)-1) \times \dfrac{(n-q(h_i))}{n} \\
                         = & \exp(l).
\end{aligned}
\end{equation}
Note that when $q(h_j) = q(h_i)$, we will have $h_j=h_i$, thus, the element $[K_T]_{ij}$ is equal to $\exp(l)$. Similar arguments can be made when $q(h_i)=1$ or $q(h_i)=n$, that is, $[K_T]_{ij}$ is equal to $\exp(l)$. Therefore, $[\Phi E \Psi^T]_{ij} = [K_T]_{ij} = \exp(l)$.

\textit{Case 2}: $q(h_j)\geq q(h_i)+1$. Let us first consider $1<q(h_i)$, then,
\begin{equation} \nonumber
\begin{aligned}
    [\Phi E \Psi^T]_{ij} = & 1 \times (\exp(l)+n-1)\times \dfrac{1}{n} \\
                           & + (-1) \times (\exp(l)-1) \times \dfrac{(-q(h_i)+1)}{n} \\
                           & + 1 \times (\exp(l)-1) \times \dfrac{(-q(h_i))}{n} \\
                         = & 1.
\end{aligned}
\end{equation}
In this case, with $q(h_j)\geq q(h_i)+1$, we will have $h_i \neq h_j$, hence, the element $[K_T]_{ij}$ is equal to $1$. Similar arguments can be made when $q(h_i)=1$, that is, in this case, $[K_T]_{ij}$ is also equal to $1$. Therefore, $[\Phi E \Psi^T]_{ij} = [K_T]_{ij} =1 $.

\textit{Case 3}: $q(h_j) \leq q(h_i)-1$. Let us first consider $q(h_i)<n$, then,
\begin{equation} \nonumber
\begin{aligned}
    [\Phi E \Psi^T]_{ij} = & 1 \times (\exp(l)+n-1)\times \dfrac{1}{n} \\
                           & + (-1) \times (\exp(l)-1) \times \dfrac{(n-q(h_i))}{n} \\
                           & + 1 \times (\exp(l)-1) \times \dfrac{(n-q(h_i)-1)}{n} \\
                         = & 1.
\end{aligned}
\end{equation}
Similar to \textit{Case} 2, we also have $h_i \neq h_j$. Similar arguments can be made when $q(h_i)=n$,  $[K_T]_{ij}$ is also equal to $1$. Hence, $[\Phi E \Psi^T]_{ij} = [K_T]_{ij} = 1$. 

Combining \textit{Cases} 1, 2, 3, we proved the decomposition in Eq. (\ref{eq:kernel-eig}). Now using this decomposition, we have,
\begin{equation} \nonumber
    \gamma(T; k_h; \mathcal{H}) = \log \vert I_T + \sigma^{-2}K_T \vert = \log \vert I_T + \sigma^{-2}\Phi E \Psi^T  \vert.
\end{equation}
By Sylvester's determinant theorem \cite{sylvester1851det},
\begin{equation} \label{eq:kernelcat-sylvester}
    \gamma(T; k_h; \mathcal{H}) = \log \vert I_n + \sigma^{-2}\Psi^T \Phi E \vert.
\end{equation}

Next, we prove the matrix $\Psi^T \Phi E$ is a positive semi-definite (p.s.d.) matrix, and the maximum element on the diagonal of $\Psi^T \Phi E$ is equal or less than $ T(\exp(l)+n-1)$. Let us denote $m_i$ as the number of times the categorical value $A_i$ appears in $T$ data points.
It can be directly seen that,
\begin{equation} \nonumber
\begin{aligned}
    [\Psi^T \Phi]_{ij} &= \sum_{r=1}^T \psi_i(h_r) \phi_j(h_r)                     = \sum_{r=1}^n m_r \psi_i(A_r) \phi_j(A_r).
\end{aligned}
\end{equation}
Hence, the matrix $\Psi^T \Phi$ can be written as,
\begin{equation} \label{eq:psiphi-decomp}
    \quad \Psi^T \Phi = \Psi_A^T F \Phi_A, \quad \quad \quad \Phi_A, \Psi_A, E \in \mathbb{R}^{n \times n},
\end{equation}
where
\begin{equation} \nonumber
\begin{aligned}
    &\Phi_A = \begin{bmatrix} \phi(A_1) \\ \phi(A_2) \\\dots \\ \phi(A_n) \end{bmatrix}
    = 
    \begin{bmatrix} 1 & 1 & 0 & 0 & ... & 0 & 0 \\
                             1 & -1 & 1 & 0 & \dots & 0 & 0 \\
                             1 & 0 & -1 & 1 & \dots & 0 & 0 \\
                              &  &  &  & \dots &  &  \\
                             1 & 0 & 0 & 0 & \dots & -1 & 1 \\
                             1 & 0 & 0 & 0 & \dots & 0 & -1
             \end{bmatrix}, \\
\end{aligned}
\end{equation}
\begin{equation} \nonumber
\begin{aligned}
    &\Psi_A = \begin{bmatrix} \psi(A_1) \\ \psi(A_2) \\ \dots \\ \psi(A_n) \end{bmatrix}, \\
    &= 
    \dfrac{1}{n} \begin{bmatrix} 1 & n-1 & n-2 & ... & 2 & 1 \\
                             1 & -1 & n-2  & ... & 2 & 1 \\
                             1 & -1 & -2  & ... & 2 & 1 \\
                              &  &  &  & ... &  &  \\
                             1 & -1 & -2 & ... & -(n-2) & 1 \\
                             1 & -1 & -2 & ... & -(n-2) & -(n-1)
             \end{bmatrix}, \\
    & \ F \ = \diag(m_1, m_2, \dots, m_n).
\end{aligned}
\end{equation}
It is straightforward that $\Psi_A^T \Phi_A = I_n$, thus, from Eq. (\ref{eq:psiphi-decomp}), we can see that $\Psi_A^T F \Phi_A$ is an eigendecomposition of $\Psi^T \Phi$, and hence, the eigenvalues of $\Psi^T \Phi$ are $m_1, m_2, \dots, m_n$. As $m_i \geq 0 , \forall i=1,...,n$, so $\Psi^T \Phi$ is a p.s.d. matrix, and therefore, $\Psi^T \Phi E$ is also a p.s.d. matrix. Besides, note that the $r$-th element on the diagonal of $\Psi^T \Phi E$ can be computed as $[\Psi^T \Phi]_{rr} E_r$ where $[\Psi^T \Phi]_{rr}$ and $E_r$ are the $r$-th elements on the diagonal of $\Psi^T \Phi$ and $E$, respectively. Since $[\Psi^T \Phi]_{rr} \leq \sum_{i=1}^T (n-1)/n \times 1 \leq T$, and $E_r \leq (\exp(l)+n-1)$, hence, $[\Psi^T \Phi]_{rr} E_r \leq T(\exp(l)+n-1)$. This results that the maximum element on the diagonal of $\Psi^T \Phi E$ is equal or smaller than $T(\exp(l)+n-1)$.

Combining Eq. (\ref{eq:kernelcat-sylvester}) and the Hadamard's inequality \cite{Vladimir1999Hadamard} on the positive semi-definite matrix $\Psi^T \Phi E$, we have,
\begin{equation} \nonumber
    \gamma(T; k_h; \mathcal{H}) \leq \log \vert I_n + \sigma^{-2}W \vert,
\end{equation}
where $W=\diag (\diag^{-1} (\Psi^T \Phi E))$. Since the maximum element on the diagonal of $\Psi^T \Phi E$ is equal or smaller than $T(\exp(l)+n-1)$. Therefore, $\gamma(T; k_h; \mathcal{H}) = \mathcal{O}(n\log(1+\sigma^{-2}T(\exp(l)+n-1))) = \mathcal{O}(n\log T)$. $\quad \square$

Now let consider the case when the objective function $f$ has $d_h$ categorical variables where each variable has $n_j$ distinct values. This can be considered to be equivalent to the case when $f$ has one variable with $\prod_{j=1}^{d_h} n_j$ distinct values. Thus, the same proof can be used, and we have $\gamma(T; k_h; \mathcal{H}) = \mathcal{O}\big((\prod_{j=1}^{d_h} n_j) \log T \big)$. $\quad \square$

\subsubsection{Maximum information gain of the mixed kernel}
\label{sec:proof-kernelmixmig}

We make use of Theorems 2 and 3 in \citet{Krause_2011Contextual} to bound the maximum information gain of the mixed kernel $k$. In particular, Theorem 2 states that given two kernels: $k_h$ on $\mathcal{H}$ and $k_x$ on $\mathcal{X}$, and if $k_h$ is a kernel on $\mathcal{H}$ with rank at most $m$, then $\gamma(T; k_h k_x; [\mathcal{H}, \mathcal{X}]) \leq m \gamma(T; k_x; \mathcal{X}) + m\log T$. On the other hand, Theorem 3 states that for any two kernels $k_h$ on $\mathcal{H}$ and $k_x$ on $\mathcal{X}$, then $\gamma(T; k_h + k_x; [\mathcal{H}, \mathcal{X}]) \leq \gamma(T; k_x; \mathcal{X}) + \gamma(T; k_h; \mathcal{X}) + 2\log T$.

As proven in Section \ref{sec:proof-kernelcatmig}, the kernel $k_h$ has at most rank $\tilde{N} = \prod_{j=1}^{d_h} n_j$ (based on the eigendecomposition). Thus, using Theorem 2 in \citet{Krause_2011Contextual}, we have
\begin{equation} \label{eq:kernel-mul}
    \gamma(T; k_h k_x; [\mathcal{H}, \mathcal{X}]) \leq \tilde{N} \gamma(T; k_x; X) + \tilde{N} \log T.
\end{equation}
Similarly, using Theorem 3 in \citet{Krause_2011Contextual}, we obtain
\begin{equation} \label{eq:kernel-sum}
    \gamma(T; k_h + k_x; [\mathcal{H}, \mathcal{X}]) \leq \mathcal{O}\big(\gamma(T; k_x; X) + (\tilde{N} + 2) \log T \big).
\end{equation}

We have the mixed kernel $k$ defined as $\lambda (k_x k_h) + (1 - \lambda) (k_h + k_x ) $ where $\lambda \in [0, 1]$ is a trade-off parameter. By combining Eqs. (\ref{eq:kernel-mul}) and (\ref{eq:kernel-sum}), we have, 
\begin{equation}
\begin{aligned} \nonumber
    \gamma(T; k; [\mathcal{H}, \mathcal{X}] & \leq \lambda \mathcal{O} \big(\tilde{N} \gamma(T; k_x; X) + \tilde{N} \log T \big) \\
    & \ + (1-\lambda)(\gamma(T; k_x; X) + (\tilde{N} + 2) \log T) \\
    & \leq \mathcal{O} \big( (\tilde{N}\lambda + 1 - \lambda)\gamma(T; k_x; X) \\
    & \ + (\tilde{N} + 2 - 2\lambda) \log T \big). \quad \square
\end{aligned}
\end{equation}

\subsection{Proof of Theorem \ref{thr:local-converge}}
\label{sec:proof-localconverge}

We prove that under Assumptions \ref{assu:f-bounded} \& \ref{assu:gp-approx}, after a restart, (1) if \gls{CASMOPOLITAN} terminates after a finite number of iterations, then it converges to a local maxima of $f$, or, (2) if \gls{CASMOPOLITAN} does not terminate after a finite number of iterations, then it converges to the global maximum of $f$. We prove this property by contradiction. 

First, let us assume after a restart, case (2) occurs, i.e. \gls{CASMOPOLITAN} does not terminate after a finite number of iterations. This means when the iteration $t$ goes to infinity, the \gls{TR} length $L^h$ is not shrunk below $L^h_{\min}$ in the categorical setting, or, both $L^h$ and $L^x$ are not shrunk below $L^h_{\min}$ and $L^x_{\min}$ in the mixed space setting. From the algorithm description, the \gls{TR} is shrunk after \texttt{fail\_tol} consecutive failures. Thus, if after $N_{\min} = \texttt{fail\_tol} \times m$ iterations where $m = \lceil \log_{\alpha_e}(L^h_0/L^h_{\min}) \rceil$\footnote{The operator $\lceil . \rceil$ denotes the ceiling function} in the categorical setting and $m = \max(\lceil \log_{\alpha_e}(L^h_0/L^h_{\min}) \rceil, \lceil \log_{\alpha_e}(L^x_0/L^x_{\min}) \rceil)$ in the mixed space setting, there is no success, \gls{CASMOPOLITAN} terminates. This means, in order for case (2) to occur, \gls{CASMOPOLITAN} needs to have at least one improvement per $N_{\min}$ iterations. Let consider the series $\lbrace f(\mathbf{z}^k) \rbrace_{k=1}^{\infty}$ where $f(\mathbf{z}^k)= \max_{i=(k-1)N_{\min}+1,\dots, kN_{\min}} \lbrace f(\mathbf{z}_i) \rbrace$ and $f(\mathbf{z}_i)$ is the function value at iteration $i$. This series is strictly increasing and the objective function $f(\mathbf{z})$ is bounded (Assumption \ref{assu:f-bounded}). Thus, using the monotone convergence theorem \cite{bibby1974axiomatisations}, this series converges to the global maximum of the objective function $f$.

Second, let consider case (1) occurs, i.e. \gls{CASMOPOLITAN} terminates after a finite number of iterations. We will prove that in this case, \gls{CASMOPOLITAN} converges to a local maxima of  $f(\mathbf{z})$ given Assumption \ref{assu:gp-approx}. For simplicity, let us consider the categorical setting first. Let us denote $L_s$ as the largest \gls{TR} length that after being shrunk, the algorithm terminates. By the definition of $L_s$, we have $\lfloor \alpha_s L_s \rfloor \leq L^h_{\min}$.\footnote{The operator $\lfloor . \rfloor$ denotes the floor function} Due to $\lfloor \alpha_s L_s \rfloor \leq \alpha_s L_s < \lfloor \alpha_s L_s \rfloor +1$, we have $ L_s < (L^h_{\min}+1)/\alpha_s$. And because $L_s$ is an integer, we finally have $L_s \leq \lceil (L^h_{\min}+1)/\alpha_s \rceil - 1$. By choosing $L_s = \lceil (L^h_{\min}+1)/\alpha_s \rceil - 1$, we have that $\forall L > L_s, \alpha_s L \geq \alpha_s \lceil (L^h_{\min}+1)/\alpha_s \rceil > L^h_{\min} $. This says that for all \gls{TR} with length $L > L_s$, after being shrunk one time, the algorithm doesn't terminate yet. Therefore, $L_s= \lceil (L^h_{\min}+1)/\alpha_s \rceil - 1$ is the largest \gls{TR} length that after being shrunk, the algorithm terminates. This tells us when the \gls{TR} length first becomes smaller or equal than $L_s$, \gls{CASMOPOLITAN} does not terminate yet (\textit{Conclusion} 1). In addition, since \gls{GP} can fit $f$ accurately within a \gls{TR} with length $L_s$ (Assumption \ref{assu:gp-approx}), for any \gls{TR} with length $L \le L_s$, the solution of BO is a success (\textit{Conclusion} 2). Combining \textit{Conclusions} 1 \& 2, we have that when \gls{TR} length first becomes smaller or equal than $L_s$, if the current \gls{TR} center is not a local maxima, \gls{CASMOPOLITAN} can find a new data point whose function value larger than the function value of current \gls{TR} center. Thus, in the next iteration, the \gls{TR} still keeps the same length whilst having center as the new found data point. This process occurs iteratively until a local maxima is reached (i.e. when \gls{CASMOPOLITAN} fails to improve from the current center), and \gls{CASMOPOLITAN} terminates.

Similar arguments can be made for the mixed space setting. Let us remind that for the mixed space setting, \gls{CASMOPOLITAN} terminates when either the continuous \gls{TR} length $\leq L^x_{\min}$ or the categorical \gls{TR} length $\leq L^h_{\min}$. Now let us consider two cases. Case (i): when the continuous \gls{TR} reaches $L^x_{\min}/\alpha_s$, the corresponding length of the categorical \gls{TR} is $\lceil L_0^h L^x_{\min}/ (\alpha_s L_0^x) \rceil$. Case (ii): when the categorical \gls{TR} length reaches $\lceil (L^h_{\min}+1)/\alpha_s \rceil -1$, the corresponding length of the continuous \gls{TR} is $L_0^x (\lceil (L^h_{\min}+1)/\alpha_s \rceil -1)/L_0^h$. Based on Assumption \ref{assu:gp-approx}, \gls{GP} can fit accurately a \gls{TR} with continuous length $L^x \leq \max \big(L^x_{\min}/\alpha_s, L_0^x (\lceil (L^h_{\min}+1)/\alpha_s \rceil -1)/L_0^h \big)$ and $L^h \leq \max \big(\lceil (L^h_{\min}+1)/\alpha_s \rceil - 1, \lceil L_0^h L^x_{\min}/ (\alpha_s L_0^x) \rceil \big)$, then when Case (i) or Case (ii) occurs, the \gls{GP} approximates accurately the objective function $f$ within the corresponding \gls{TR}, and thus similar argument as in the categorical setting can be made. That is, if the current \gls{TR} center is not a local maxima, then \gls{CASMOPOLITAN} can find a new data point whose function value larger than the function value of current \gls{TR} center. And this process occurs iteratively until a local maxima is reached, and  \gls{CASMOPOLITAN} terminates. $\quad \square$

\vspace{-0.1cm}
\subsection{Proof of Theorem \ref{thr:gconverge-cat}} \label{sec:proof-catconverge}

Let us first remind our restart strategy in the categorical setting. At the $i$-th restart, we first fit an auxiliary global \gls{GP} model $GP(0, k_h)$ on a subset of data $D^*_{i-1}= \lbrace \mathbf{h}^*_j, f(\mathbf{h}^*_j) \rbrace_{j=1}^{i-1}$, where $\mathbf{h}^*_j$ is the local maxima found after the $j$-th restart, or, a random data point, if the found local maxima after the $j$-th restart is same as one of previous restart. Let us also denote $\mu_{gl}(\mathbf{h}; D^*_{i-1})$ and $\sigma^2_{gl}(\mathbf{h}; D^*_{i-1})$ as the posterior mean and variance of the global \gls{GP} learned from $D^*_{i-1}$. Then, at the $i$-th restart, we select the following location $\mathbf{h}^{(0)}_i$ as the initial centre of the new \gls{TR}:
\begin{align} \nonumber
    \mathbf{h}^{(0)}_i= \arg \max_{\mathbf{h} \in \mathcal{H}} \mu_{gl}(\mathbf{h}; D^*_{i-1}) + \sqrt{\beta_i} \sigma_{gl}(\mathbf{h}; D^*_{i-1}),
\end{align}
where $\beta_i$ is the trade-off parameter in \gls{GP}-\textsc{ucb} \cite{Srinivas_2010Gaussian}. 

To prove the convergence property of \gls{CASMOPOLITAN}, apart from Assumptions \ref{assu:f-bounded} \& \ref{assu:gp-approx}, let us also assume that at the $i$-th restart, there exists a function $g_i(\mathbf{h})$ that: (a) is a sample from the global  $GP(0, k_h)$, (b) shares the same global maximum $\mathbf{h}^*$ with $f$, and, (c) passes through the all the local maxima of $f$ and any data point $\mathbf{h'}$ in $\mathcal{D}^*_{i-1} \cup \lbrace \mathbf{h}^{(0)}_i \rbrace$ that are not local maxima (i.e. $g_i(\mathbf{h'}) = f(\mathbf{h'}) \ \forall \mathbf{h'} \in D^*_{i-1} \cup \lbrace \mathbf{h}^{(0)}_i \rbrace$). In layman's terms, the function $g_i(\mathbf{h})$ is a function that passes through all the maxima of $f$ and is a sample from the auxiliary global $GP(0, k_h)$. It is worth noting that our assumption is more relaxed than the assumption in \citet{Srinivas_2010Gaussian} where it is assumed that the objective function $f$ must be sampled from the global $\gls{GP}(0, k_h)$. Specifically, it can be seen that if the assumption in \citet{Srinivas_2010Gaussian} holds, our assumption also holds because if $f(\mathbf{h})$ is a sample from $\gls{GP}(0, k_h)$, then a choice for $g_i(\mathbf{h})$ is $f(\mathbf{h})$, thus, our assumption holds.

Using Lemmas 5.1 and 5.2 in \citet{Srinivas_2010Gaussian} for the function $g_i$, when $\beta_i=2\log(\vert \mathcal{H} \vert i^2 \pi^2 / 6 \zeta)$, for all $i$, with probability $1-\zeta$, we have,
\begin{equation} \nonumber
\begin{aligned}
& \mu_{gl}(\mathbf{h}^{(0)}_i; D^*_{i-1}) + \sqrt{\beta_i} \sigma_{gl}(\mathbf{h}^{(0)}_i; D^*_{i-1}) \\
& \quad \geq \mu_{gl}(\mathbf{h}^{*}; D^*_{i-1}) + \sqrt{\beta_i} \sigma_{gl}(\mathbf{h}^{*}; D^*_{i-1}) \\
& \quad \geq g_i(\mathbf{h}^{*}).
\end{aligned}
\end{equation}
Thus, with probability $1-\zeta$,
\begin{equation} \nonumber
\begin{aligned}
& g_i(\mathbf{h}^{*}) - g_i(\mathbf{h}^{(0)}_i) \\ & \leq \mu_{gl}(\mathbf{h}^{(0)}_i; D^*_{i-1}) + \sqrt{\beta_i} \sigma_{gl}(\mathbf{h}^{(0)}_i; D^*_{i-1}) - g_i(\mathbf{h}^{(0)}_i) \\
& \leq 2\sqrt{\beta_i} \sigma_{gl}(\mathbf{h}^{(0)}_i; D^*_{i-1}).
\end{aligned}
\end{equation}

Combining this inequality with the fact that $g_i(\mathbf{h}^{(0)}_i) = f(\mathbf{h}^{(0)}_i)$, and $g_i(\mathbf{h}_i^*)=f(\mathbf{h}_i^*)$, we have, with probability $1-\zeta$,
\begin{equation} \nonumber
    f(\mathbf{h}^{*}) - f(\mathbf{h}^{(0)}_i) \leq 2\sqrt{\beta_i} \sigma_{gl}(\mathbf{h}^{(0)}_i; D^*_{i-1}).
\end{equation}

Let us denote $\mathbf{h}_i^*$ as the local maxima found by \gls{CASMOPOLITAN} at the $i$-th restart. As $f(\mathbf{h}^{(0)}_i) \leq f(\mathbf{h}_i^*)$, therefore,
\begin{equation} \nonumber
    f(\mathbf{h}^{*}) - f(\mathbf{h}_i^*) \leq 2\sqrt{\beta_i} \sigma_{gl}(\mathbf{h}^{(0)}_i; D^*_{i-1}).
\end{equation}
This results that, with probability $1-\zeta$,
\begin{equation} \nonumber
    R_I = \sum_{i=1}^I (f(\mathbf{h}^{*}) - f(\mathbf{h}_i^*)) \leq \sum_{i=1}^I 2\sqrt{\beta_i} \sigma_{gl}(\mathbf{h}^{(0)}_i; D^*_{i-1}).
\end{equation}
Finally, using Lemmas 5.3 and 5.4 in \citet{Srinivas_2010Gaussian}, we can bound $R_I$ as $R_I \leq \sqrt{I C_1 \beta_I \gamma(I; k_h, \mathcal{H})}$ with $C_1=8/\log(1+\sigma^{-2})$ and $\gamma(I; k_h, \mathcal{H})$ being the maximum information gain for the categorical kernel derived in Theorem \ref{thr:kernelmig}. $\quad \square$

\subsection{Proof of Theorem \ref{thr:gconverge-mix}}
\label{sec:proof-mixconverge}

Similar to the proof for categorical setting in Section \ref{sec:proof-catconverge}, let us first remind our restart strategy in the mixed space setting. Suppose we are restarting the $i$-th time, we first fit the global \gls{GP} model on a subset of data $D^*_{i-1}= \lbrace \mathbf{z}^*_j, f(\mathbf{z}^*_j) \rbrace_{j=1}^{i-1}$, where $\mathbf{z}^*_j$ is the local maxima found after the $j$-th restart, or, a random data point, if the found local maxima after the $j$-th restart is same as one of previous restart. Let us also denote $\mu_{gl}(\mathbf{z}; D^*_{i-1})$ and $\sigma^2_{gl}(\mathbf{z}; D^*_{i-1})$ as the posterior mean and variance of the global \gls{GP} learned from $D^*_{i-1}$. Then, at the $i$-th restart, we select the following location $\mathbf{z}^{(0)}_i$ as the initial centre of the new \gls{TR}:
\begin{align} \nonumber
    \mathbf{z}^{(0)}_i= \arg \max_{\mathbf{z} \in [\mathcal{H}, \mathcal{X}]} \mu_{gl}(\mathbf{z}; D^*_{i-1}) + \sqrt{\beta_i} \sigma_{gl}(\mathbf{z}; D^*_{i-1}),
\end{align}
where $\beta_i$ is the trade-off parameter in \gls{GP}-\textsc{ucb} \cite{Srinivas_2010Gaussian}. 

To prove the convergent property of \gls{CASMOPOLITAN} in the mixed space setting, apart from Assumptions \ref{assu:f-bounded} \& \ref{assu:gp-approx}, let us also assume that at the $i$-th restart, there exists a function $g_i(\mathbf{z})$: (a) lies in the RKHS $\mathcal{G}_k(\lbrack \mathcal{H}, \mathcal{X} \rbrack)$ and $\Vert g_i \Vert_k^2 \leq B$, (b) shares the same global maximum $\mathbf{z}^*$ with $f$, and, (c) passes through all the local maxima of $f$ and any data point $\mathbf{z'}$ in $\mathcal{D}^*_{i-1} \cup \lbrace \mathbf{z}^{(0)}_i \rbrace$ which are not local maxima (i.e. $g_i(\mathbf{z'}) = f(\mathbf{z'}) \ \forall \mathbf{z'} \in D^*_{i-1} \cup \lbrace \mathbf{z}^{(0)}_i \rbrace$). In layman's terms, the function $g_i(\mathbf{z})$ is a function that passes through the maxima of $f$ whilst lying in the RKHS $\mathcal{G}_k(\lbrack \mathcal{H}, \mathcal{X} \rbrack)$ and satisfying $\Vert g_i \Vert_k^2 \leq B$. Our assumption is more relaxed than \citet{Srinivas_2010Gaussian} which assumed that the objective function $f$ lies in the RKHS $\mathcal{G}_k(\lbrack \mathcal{H}, \mathcal{X} \rbrack)$. Specifically, it can be seen that if the assumption in \citet{Srinivas_2010Gaussian} holds, our assumption also holds because if $f(\mathbf{z})$ lies in the RKHS $\mathcal{G}_k(\lbrack \mathcal{H}, \mathcal{X} \rbrack)$, then a choice for $g_i(\mathbf{z})$ is $f(\mathbf{z})$, thus, our assumption holds.

Using Theorem 6 in \citet{Srinivas_2010Gaussian} for function $g_i$, when $\beta_i= 2\Vert g_i \Vert_k^2 + 300\gamma_i \log(i/\zeta)^3$, $\forall i$, $\forall z \in \lbrack \mathcal{H}, \mathcal{X} \rbrack$, we have,
\begin{equation} \label{eq:sriniva-tr6}
\begin{aligned}
    & \text{Pr} \lbrace \vert \mu_{gl}(\mathbf{z}; D_{i-1}^*) - g_i(\mathbf{z}) \vert \leq \sqrt{\beta_i} \sigma_{gl}(\mathbf{z}; D_{i-1}^*) \vert  \rbrace \geq 1-\zeta.
\end{aligned}
\end{equation}

Note since $\Vert g_i \Vert_k^2 \leq B$, Eq. (\ref{eq:sriniva-tr6}) is also correct using $\beta_i = 2B + 300\gamma_i \log(i/\zeta)^3$. By using the inequality in Eq. (\ref{eq:sriniva-tr6}), the proof technique is similar to that in Section \ref{sec:proof-catconverge}. In particular, with probability $1-\zeta$, we have that,
\begin{equation}
\begin{aligned}
& \mu_{gl}(\mathbf{z}^{(0)}_i; D^*_{i-1}) + \sqrt{\beta_i} \sigma_{gl}(\mathbf{z}^{(0)}_i; D^*_{i-1}) \\
& \quad \geq \mu_{gl}(\mathbf{h}^{*}; D^*_{i-1}) + \sqrt{\beta_i} \sigma_{gl}(\mathbf{z}^{*}; D^*_{i-1}) \geq g_i(\mathbf{z}^{*}).
\end{aligned}
\end{equation}

Thus, with probability $1-\zeta$, we have
\begin{equation} \nonumber
\begin{aligned}
& g_i(\mathbf{z}^{*}) - g_i(\mathbf{z}^{(0)}_i) \\ & \leq \mu_{gl}(\mathbf{z}^{(0)}_i; D^*_{i-1}) + \sqrt{\beta_i} \sigma_{gl}(\mathbf{z}^{(0)}_i; D^*_{i-1}) - g_i(\mathbf{z}^{(0)}_i) \\
& \leq 2\sqrt{\beta_i} \sigma_{gl}(\mathbf{z}^{(0)}_i; D^*_{i-1}).
\end{aligned}
\end{equation}
Since $g_i(\mathbf{z}^{(0)}_i) = f(\mathbf{z}^{(0)}_i)$, and $g_i(\mathbf{z}_i^*)=f(\mathbf{z}_i^*)$, hence, $f(\mathbf{z}^{*}) - f(\mathbf{z}^{(0)}_i) \leq 2\sqrt{\beta_i} \sigma_{gl}(\mathbf{z}^{(0)}_i; D^*_{i-1})$ with probability $1-\zeta$. With $\mathbf{z}_i^*$ as the local maxima found by \gls{CASMOPOLITAN} at the $i$-th restart. As $f(\mathbf{z}^{(0)}_i) \leq f(\mathbf{z}_i^*)$, therefore,
\begin{equation} \nonumber
    f(\mathbf{z}^{*}) - f(\mathbf{z}_i^*) \leq 2\sqrt{\beta_i} \sigma_{gl}(\mathbf{z}^{(0)}_i; D^*_{i-1}).
\end{equation}
This results, with probability $1-\zeta$,
\begin{equation} \nonumber
    R_I = \sum_{i=1}^I (f(\mathbf{z}^{*}) - f(\mathbf{z}_i^*)) \leq \sum_{i=1}^I 2\sqrt{\beta_i} \sigma_{gl}(\mathbf{z}^{(0)}_i; D^*_{i-1}).
\end{equation}

Finally, using Lemmas 5.3 and 5.4 in \citet{Srinivas_2010Gaussian}, we can bound $R_I$ as $R_I \leq \sqrt{I C_1 \beta_I \gamma(I; k, \lbrack \mathcal{H}, \mathcal{X} \rbrack)}$ with $C_1=8/\log(1+\sigma^{-2})$ and $\gamma(I; k, \lbrack \mathcal{H}, \mathcal{X} \rbrack)$ is the maximum information gain for the mixed kernel derived in Theorem \ref{thr:kernelmig}. $\quad \square$

\paragraph{Discussion} It is worth emphasizing that the assumption of the existence of such a function $g_i(\mathbf{z})$ (at the $i$-th restart) that satisfies our requirements generally does not need to hold when $i \rightarrow \infty$. In fact, if this assumption needs to satisfy $\forall i$ when $i \rightarrow \infty$ then it will be same as the assumption in \citet{Srinivas_2010Gaussian}. We will show that generally this assumption only needs to hold for a finite number of restarts. In particular, it is common that for the objective function $f$, there exists a local maxima $\mathbf{\tilde{z}}^*$ which is larger than all other local maxima and only smaller than the global maximum. Then as $\lim_{I \rightarrow \infty} R_I/I = 0$, there exists a finite number $I_0$ that the function value of the \gls{TR} center at the $I_0$-th restart will be larger than $f(\mathbf{\tilde{z}}^*)$, and thus the `local maxima' found after the $I_0$-th restart is actually the global maximum, and \gls{CASMOPOLITAN} converges. Therefore, our assumption regarding the existence of $g_i(\mathbf{z})$ only needs to hold until the $I_0$-th restart. This discussion is applicable for both categorical and mixed space settings.

\end{document}